\documentclass[10pt,twocolumn,letterpaper]{article}

\usepackage{iccv}              

\usepackage{graphicx}
\usepackage{booktabs}

\usepackage[utf8]{inputenc} 
\usepackage[T1]{fontenc}    
\usepackage{url}            
\usepackage{amsfonts}       
\usepackage{nicefrac}       
\usepackage{microtype}      
\usepackage{xcolor}         
\usepackage{graphicx}
\usepackage{multirow}
\usepackage{graphicx}
\usepackage{amsmath}
\usepackage{bbm}
\usepackage{placeins}
\usepackage{subcaption}
\usepackage{placeins}
\usepackage{overpic}
\usepackage{float}

\newif\ifdraft
\draftfalse
\drafttrue 

\definecolor{orange}{rgb}{1,0.5,0}
\definecolor{violet}{RGB}{70,0,170}
\definecolor{magenta}{RGB}{170,0,170}
\definecolor{dgreen}{RGB}{0,150,0}

\ifdraft
 \newcommand{\PF}[1]{{\color{red}{\bf PF: #1}}}

 \newcommand{\BGU}[1]{{\color{olive}{\bf BG: #1}}}
 
  \newcommand{\ME}[1]{{\color{dgreen}{\bf ME: #1}}}

 \newcommand{\TODO}[1]{\textbf{\color{red}[TODO: #1]}}
\else
 \newcommand{\PF}[1]{}
 
 \newcommand{\WL}[1]{}
 
 \newcommand{\BGU}[1]{}
 
 \newcommand{\ME}[1]{}
  
  \newcommand{\TODO}[1]{}
  
\fi

\newcommand{\comment}[1]{}
\newcommand{\parag}[1]{\vspace{-3mm}\paragraph{#1}}

\newcommand{\bC}{\mathbf{C}}

\newcommand{\bg}{\mathbf{g}}

\newcommand{\bu}{\mathbf{u}}
\newcommand{\bv}{\mathbf{v}}

\newcommand{\bx}{\mathbf{x}}

\newcommand{\bgrad}{\mathbf{g}}

\usepackage[export]{adjustbox}
\usepackage{multirow}
\definecolor{iccvblue}{rgb}{0.21,0.49,0.74}
\usepackage[pagebackref,breaklinks,colorlinks,hidelinks,allcolors=iccvblue]{hyperref}

\title{Gradient Distance Function}

\author{
Hieu Le$^{1}$ \quad
Federico Stella$^{2}$ \quad
Benoit Guillard$^{2}$ \quad 
Pascal Fua$^{2}$ \\
$^{1}$UNC-Charlotte \quad 
$^{2}$EPFL\\[0.25em]
{\tt\small hle40@charlotte.edu \quad {first-name.last-name}@epfl.ch}
}

\begin{document}
\maketitle

\begin{abstract}

Unsigned Distance Functions (UDFs) can be used to represent non-watertight surfaces in a deep learning framework. However, UDFs tend to be brittle and difficult to learn, in part because the surface is located exactly where the UDF is non-differentiable. In this work, we show that Gradient Distance Functions (GDFs) can remedy this by being differentiable at the surface while still being able to represent open surfaces. This is done by associating to each 3D point a 3D vector whose norm is taken to be the unsigned distance to the surface and whose orientation is taken to be the direction towards the closest surface point. We demonstrate the effectiveness of GDFs on ShapeNet Car, Multi-Garment, and 3D-Scene datasets with both single-shape reconstruction networks or categorical auto-decoders.

\end{abstract}    

\section{Introduction}


\begin{figure}[th!]
\begin{center}
\includegraphics[width=0.95\linewidth]{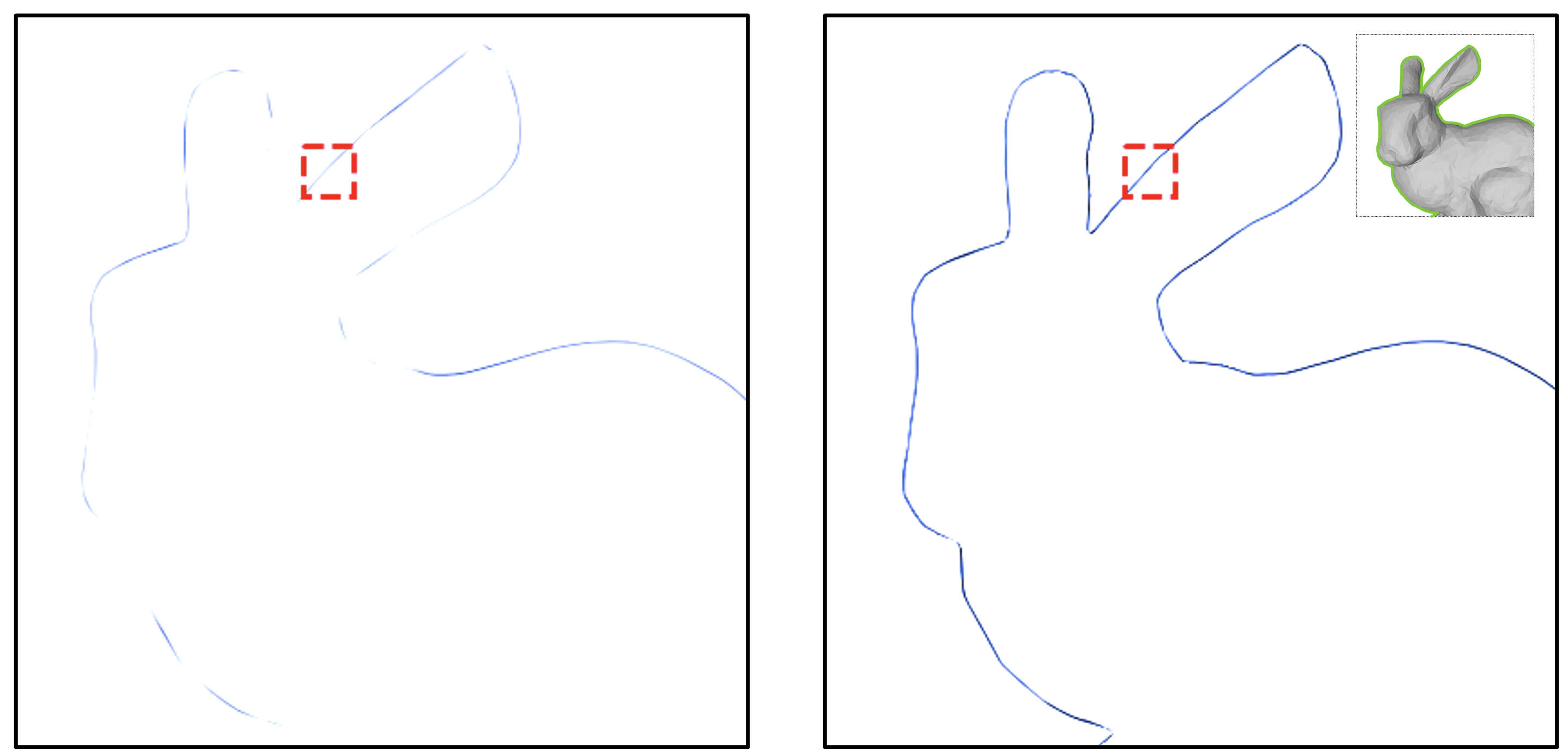}
\makebox[0.45\linewidth]{(a) UDF}
\makebox[0.45\linewidth]{(b) GDF}
\includegraphics[width=0.95\linewidth]{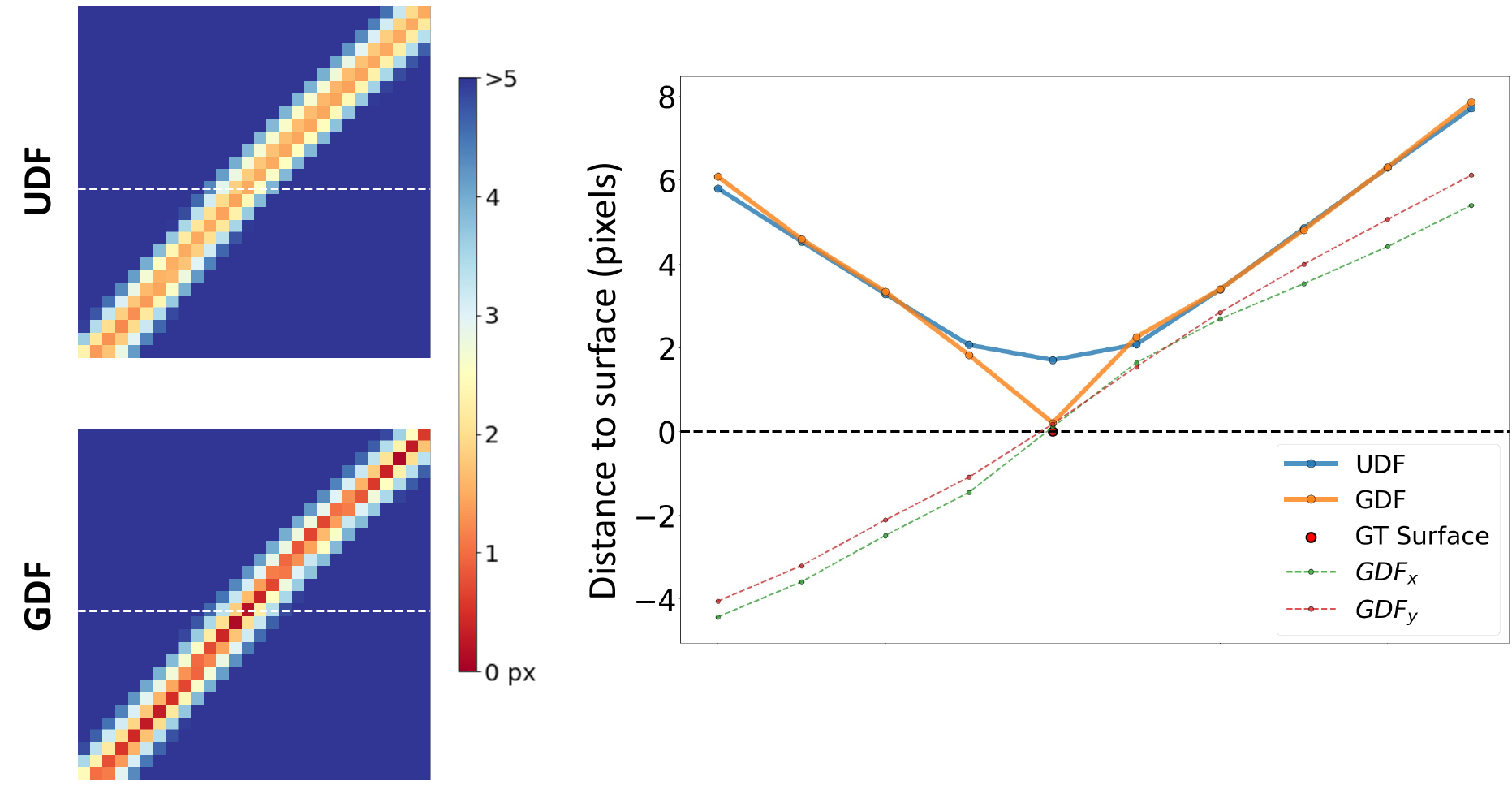}

\makebox[0.3\linewidth]{(c)}
\makebox[0.65\linewidth]{(d)}
\end{center} 
\vspace{-5mm}
\caption{
\textbf{UDF vs GDF.}  (a) We trained a network to output a UDF representing an open 2D contour for a portion of the Stanford Bunny. 
The pixels whose distances to the resulting contours is smaller than 2 pixels are shown in blue. Note that many parts of the contour are missing. (b) We repeated the experiment but using our GDF instead and there are no more holes. (c)  Estimated distances to the contour in the small patch denotes by the red box in (a,b). (d) Value of the UDF and GDF along the dotted line in (c). Note that the UDF never reaches 0, resulting in a hole in the contour. This problem goes away when using the GDF, hence the much better defined contour in (b). 
}  
\label{fig:teaser} 
\end{figure}

Signed Distance Functions (SDFs) excel at representing deep implicit surfaces~\cite{Park19c,Chen19c}. However, they are best suited for representing watertight surfaces. They can also be used to represent non-watertight ones but that usually involves inflating the surface by fitting an implicit surface around it, which incurs a loss of accuracy. Unsigned Distance Functions (UDFs) can remedy this~\cite{Long22,Chibane20b,Guillard22b} but pose their own challenges: they are strictly positive everywhere except for zero-values at surface points, which means that the surface is located exactly at the non-differentiable part of the distance field. Consequently, optimizing a UDF is numerically unstable, resulting the function never reaching the zero value and creating holes in the resulting surface, or reaching zero in a volume and creating double surfaces. Hence, UDF-based representations are more difficult to learn than their SDF-based counterparts. 

In this work, we show that instead of learning the distance directly, learning the gradient of the distance function can overcome these limitations. Given a surface we wish to represent, we associate to each point in a volume containing it the vector pointing to the closest surface point. Its norm is the unsigned distance to the surface, while its direction is the negative gradient of the unsigned distance. When crossing the surface, each vector component switches signs while its magnitude goes to zero and increases again afterwards. In essence, we represent a non-watertight surface by three surface differentiable sub functions, 
as depicted in Fig.~\ref{fig:teaser} for a 2D case. 



We show that this gradient distance function, namely GDF, is easy to learn through deep networks. The reconstructed surface is smooth and without staircase artefacts that often exhibit in the UDF counterparts. Additionally, the GDF representation inherently includes gradient direction, a critical factor for meshing non-watertight surfaces \cite{Guillard22b}. Like UDFs, GDFs can also be parameterized by latent vectors, allowing a pre-trained network to reconstruct 3D meshes from partial observations, such as 3D points on the surface of a target object. We conduct various experiments on the Multi-Garment, ShapeNet-Car, and 3DScenes datasets to demonstrate the effectiveness of GDFs, where they outperform UDFs with both single-shape reconstruction networks or categorical auto-decoders.

\section{Related Work}

Deep implicit surfaces~\cite{Park19c,Mescheder19,Chen19c} excel at modeling watertight surfaces using occupancy grids and SDFs. They can also be used to represent open surfaces by inflating an SDF around them~\cite{Corona21,Guillard22b,Venkatesh21}. This involves wrapping a watertight surface around them.  If needed, a triangulated surface can then be obtained using the Marching Cubes algorithm~\cite{Lorensen87}. However, this method is resolution-dependent and the resulting surface will be some distance away from the target surface, resulting in a loss of accuracy. 

 
\parag{Unsigned Distance Functions (UDFs).} 

Thus, an attractive alternative is to use UDFs~\cite{Chibane20b,Corona21,Venkatesh20,Chen22e,Wang22b,Liu23a,Meng23} because they do not suffer from this inaccuracy issue. Furthermore, triangulations can be obtained from them by reasoning on their gradients~\cite{Guillard22b,Zhang23b}. However, a UDF is non-differentiable at its zero level set, which is precisely where the surface is supposed to be. This is not easy to learn for a deep network and can result in the UDF values never reaching zero or remaining zero in a whole volume, leading to the surface disappearing or appearing as a being doubled. 

Several approaches to mitigating these drawbacks have been proposed. In~\cite{Zhou22}, the level-sets close to the surface are constrained to be parallel to it to learn UDFs from sparse point clouds. In~\cite{Zhao21a}, 3D position features extracted via a set of anchor points on the surface are used as additional inputs for predicting UDF values. This approach improves single-view garment reconstruction.  Some methods involve learning additional information.  In~\cite{Wang22b}, an additional network is used to predict a pseudo-sign for each spatial point, defined via the surface gradient of the closest surface point while the algorithm of~\cite{Venkatesh20} predicts surface orientation together with the UDF for improved modeling of local geometric details. However, none of these methods addresses the challenge inherent to training deep networks to predict  UDF values accurate enough to prevent the creation of holes and non-smooth surfaces.

\parag{Open Surface Representations.} 

Instead of improving UDF-based schemes, one can also propose different representation schemes for open surfaces. In 3PSDF~\cite{Chen22e}, they are modeled by using occupancy grids, by masking out empty regions, and by classifying unmasked grid cells as either positive or negative. These pseudo signs provide a notion of "inside" or "outside" the local surface based on the surface gradient of the nearest surface point, as in \cite{Wang22b}. However, this method requires pre-orienting the training data, and does not provide the continuous distances required for precise meshing.  Instead of directly learning the UDF, the CSP algorithm of~\cite{Venkatesh21} predicts the closest surface point. This representation is differentiable at the surface, but the scalar field does not exhibit distance-like continuity and is challenging to learn. In contrast, our proposed GDF predicts the vector pointing to the nearest surface point. When crossing the surface: like SDFs, GDFs scalar field changes from positive to negative or vice versa, while CSP yields constant values. As a result, CSPs are harder to train and yield less precise reconstructions, as we will demonstrate. One approach that is related to our method and CSP is NVF \cite{Yang_2023_CVPR}, which predicts the gradient vector for each query point using codebooks to memorize UDF patterns that match the input point cloud. Our method, whereas, shows that directly learning gradient distance functions yields a general-purpose, surface-differentiable representation—well beyond codebook-style point-cloud modeling—while remaining simple to train with a standard auto-decoder.

\section{Method}

In this section we describe how the Gradient Distance Functions (GDF) is differentiable at the surface and more robust than the original Unsigned Distance Function (UDF), which can effectively represent non-watertight shapes.

\subsection{Definition}


\begin{figure}[t]
\begin{center}
\includegraphics[width=1\linewidth]{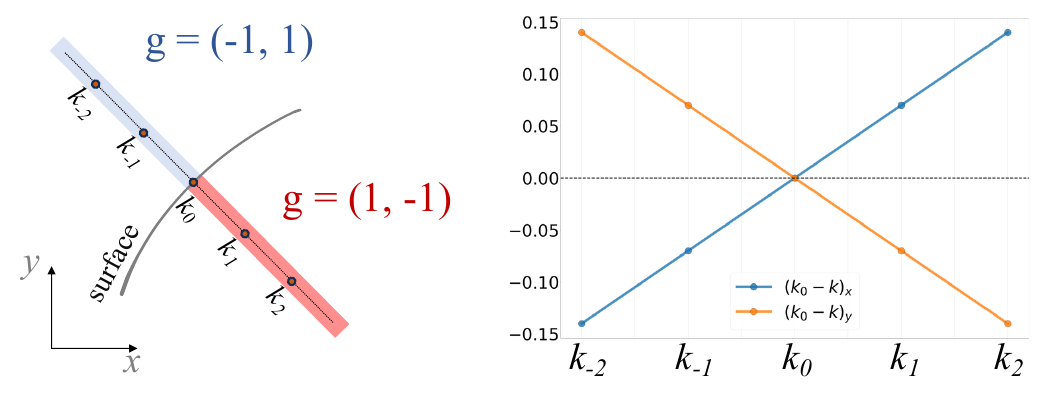}
\makebox[0.48\linewidth]{(a) }
\makebox[0.48\linewidth]{(b) }
\end{center} 

\caption{
\textbf{GDF's surface differentiability.}  (a) Points on different sides of the surface have gradients with opposite signs. (b) GDF is the product of normalized gradient and unsigned distance. It monotonically changes from one sign to the other when crossing the surface, making it differentiable. 
} 
\label{fig:adf} 
\end{figure}

We define a GDF as the function
\begin{align} 
   \textrm{GDF  : }   \mathbb{R}^3 &\rightarrow \mathbb{R}^3, \nonumber \\
    \bx \rightarrow \bv &= \hat{\bx} - \bx   \; , \label{eq:gdf} 
\end{align}
which associates to each 3D point $\bx$ the vector $\bv$ pointing to the closest surface point $\hat{\bx}$. The unsigned distance $u$ and the unit-norm gradient $\bg$ can be estimated from $\bv$  as 
\begin{align} 
     u & =\| \bv  \| \; , \label{eq:esdf}  \\ 
     \bg & = \frac{1}{u} \bv  \;  . \label{eq:esgrad} 
 \end{align}
As with UDFs, the surface is defined implicitly as the location where $\bv = [0,0,0]$  and, consequently, $u$ is zero and $\bg$ is taken to be the null vector. While a UDF is non-differentiable at a surface point, our GDF is. Each component of $\bv$ switches sign when crossing the surface while its magnitude goes to zero, which makes the behavior of GDF similar to that of SDF in this respect. Fig. \ref{fig:adf} illustrates this for a 2D case:  Let us consider a set of points $\{k_i, i\in [-N,N]\}$ sampled along a line perpendicular to the surface and crossing it at $k_0$, which makes it the surface point closest to all of them. When crossing the surface, one element of $\bv$ goes from positive to negative while the other goes from negative to positive. 




\subsection{GDFs as Output of a Deep Network}

Deep networks can be trained to regress GDFs, instead of SDFs or UDFs, in essentially the same manner.  

In a fully supervised setup, we would be given a set  $S$ of $N$ training pairs 
\begin{align}
    S     &= \{ S_i=(\bx^i,\bv^i) , 1 \leq i \leq N \} \; ,   \nonumber \\
    \bx^i &= \{\bx_1^i \ldots \bx_k^{i}\}  \; , \label{eq:pairs}  \\
    \bv^i &= \{\bv_1^i \ldots \bv_k^{i}\} \; , \nonumber 
\end{align}
where the $\bx^i_{1\ldots k}$ are sampled points for a particular shape $S_i$ and the $\bv^i$ represent the corresponding GDF vectors. Similar to deepSDF \cite{Park19a}, we sample more aggressively near the surface and a small fraction of points (5\%) is uniformly sampled throughout the entire volume.

Let  $f_\Phi(\cdot, \bC)$ be a function implemented by a network with weights $\Phi$ that takes as input a 3D point and a latent code $\bC_i$ associated to each $S_i$. We want to guarantee that 
\begin{align}
  \forall i , j  , \quad f_\Phi(\bx_j^i,\bC_i) \approx \bv_j^i \; . \label{eq:minimize}
\end{align}
To this end, we minimize with respect to the weights $\Phi$ and the latent vectors $\bC_i$ the loss function
\begin{align}
\mathcal{L}(\Phi,\bC_1,\ldots,\bC_N) &=  \sum_{i,j}   |\hat{\bv}_j^i  - \bv_j^i| \; , \label{eq:loss}\\
\hat{\bv}_j^i &=  f_\Phi(\bx_j^i,\bC_i)  \; . \nonumber
\end{align}

\section{Experiments}

We now demonstrate the representational abilities of GDFs in two different scenarios. In the first, we train a network to represent a single 3D scene. In the second. we use an auto-decoding approach to learning a latent vector representation for a class of shapes. We show that, in both cases, GDFs are easy to learn and capture fine geometric details. In the second scenario, they also generalize well to unseen shapes of the same category. 

\subsection{Datasets, Metrics, Baselines, and Settings}

We use three main datasets for our experiments: MGN \cite{Bhatnagar19}, ShapeNet-Car \cite{Chang15}, and 3D-Scene \cite{Zhou13}. MGN comprises 320 garments. They are mainly made of single-layer with a smooth 3D shape. We use two versions of the ShapeNet-Car dataset, which we denote by {\it ShapeNet-Car(R)}, {\it ShapeNet-Car(P)}. The first comprises 3091 cars with open surfaces and inner structures. The second is a cleaned-up version with 3D car models pre-processed to yield closed surfaces without interiors~\cite{Xu19b}. Last, the 3D-Scene dataset comprises seven samples of 3D scenes with highly complex geometry and many details. 

\parag{Metrics.} 
To evaluate performance, we use the Chamfer distance (CD) and the Normal consistency (NC). The former measures the distance between 3D points sampled from the surfaces, the lower the better. The latter quantifies the surface normal similarity between the predicted mesh and ground-truth mesh, the higher the better. 

\parag{Baselines.} 
We compare our method against three other implicit-surface methods, NDF~\cite{Chibane20b} that predicts UDF values directly, CSP~\cite{Wang22b} that regresses the closest surface point, and 3PSDF~\cite{Chen22e}  that labels points as either ``$-1$'',``$1$'', or ``\textit{null}''. To ensure fair comparison in each experiment, we train different networks with the same architecture, training data, and optimization settings. The only difference is how the output is representated. For each training instance, we extract 400000 points near the surface and 20000 points uniformly sampled throughout the entire volume. We use Adam optimizer \cite{Kingma14a} with a starting learning rate of $10^{-4}$ that decreases by 25\% after each quater of the training. We use MeshUDF~\cite{Guillard22b} as the meshing algorithm for all methods except for the 3PSDF representation that can be meshed using Marching Cubes. MeshUDF reasons on the gradient of the UDF for meshing. For NDF, we compute this gradient by running a back-propagating pass through the network, as in deepSDF \cite{Park19c}. For CSP and our method, the gradient can be computed directly from the output representation. In our case, it is given by Eq.~\ref{eq:esgrad}, without the need to back-propagate. We never perform any post-processing step. 


\subsection{Representing a Single Complex Shape}

\def\subfigsize{0.192\linewidth}
\begin{figure*}[th!]
\begin{center}
\includegraphics[width=\subfigsize]{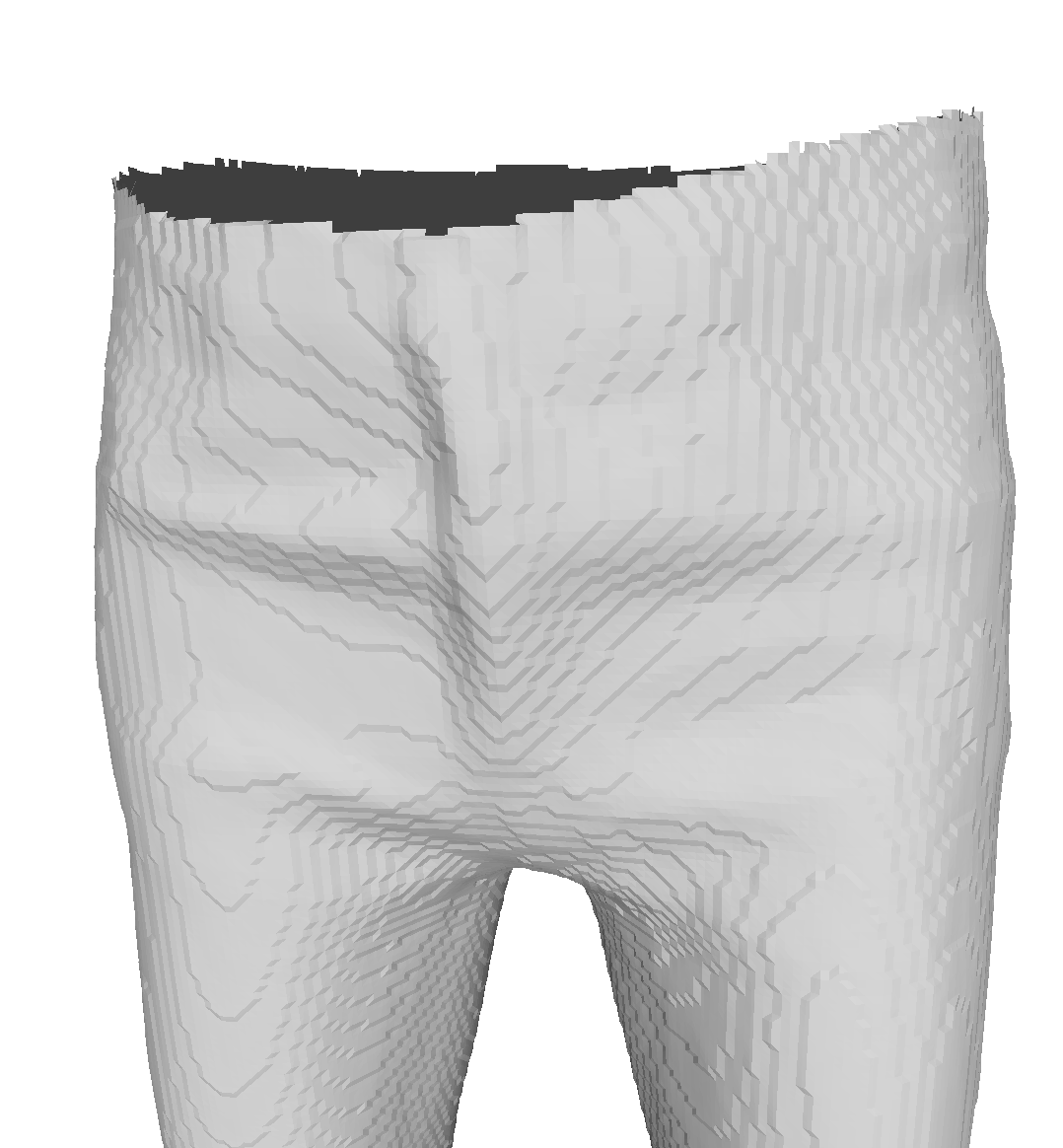}
\includegraphics[width=\subfigsize]{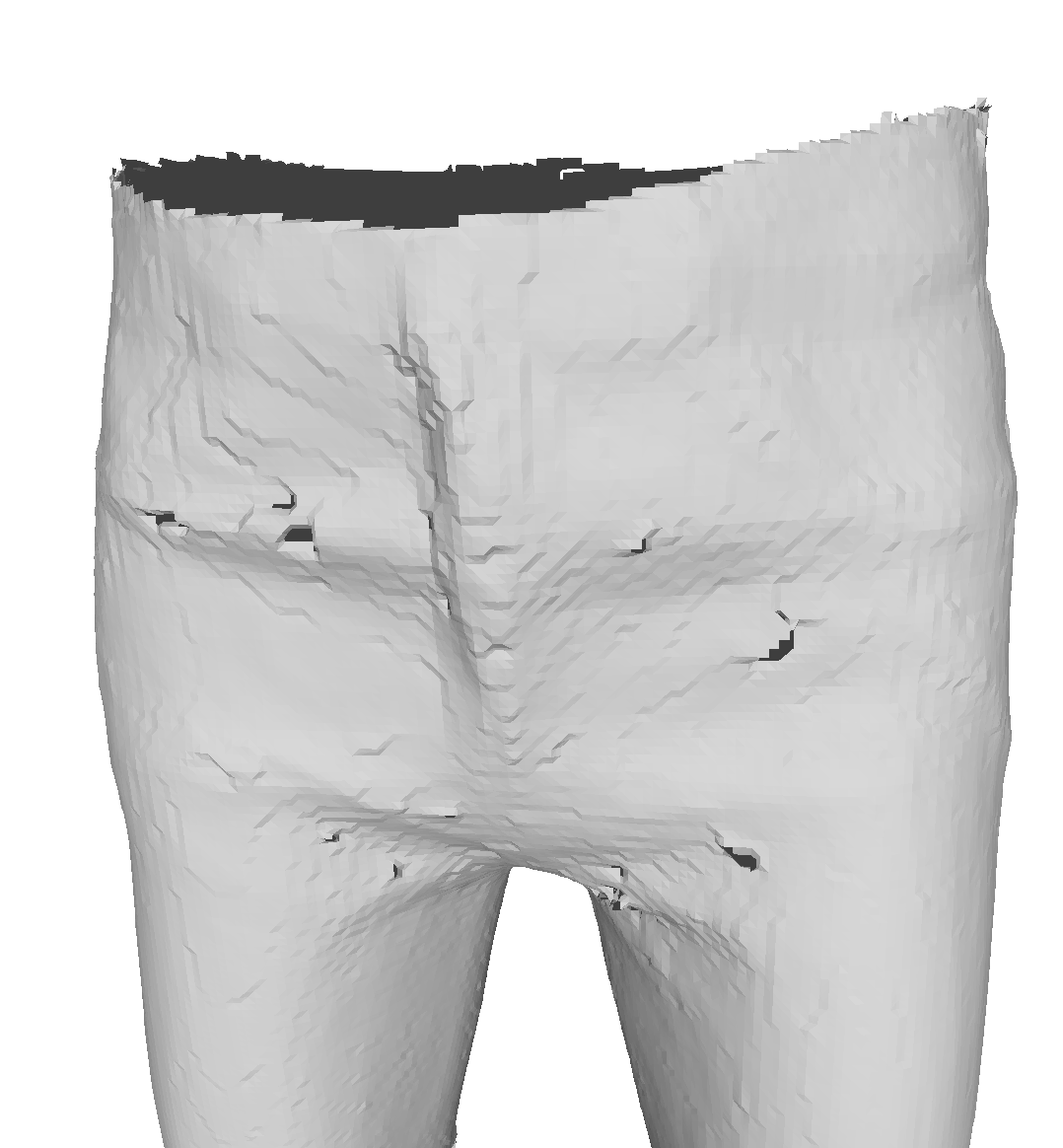}
\includegraphics[width=\subfigsize]{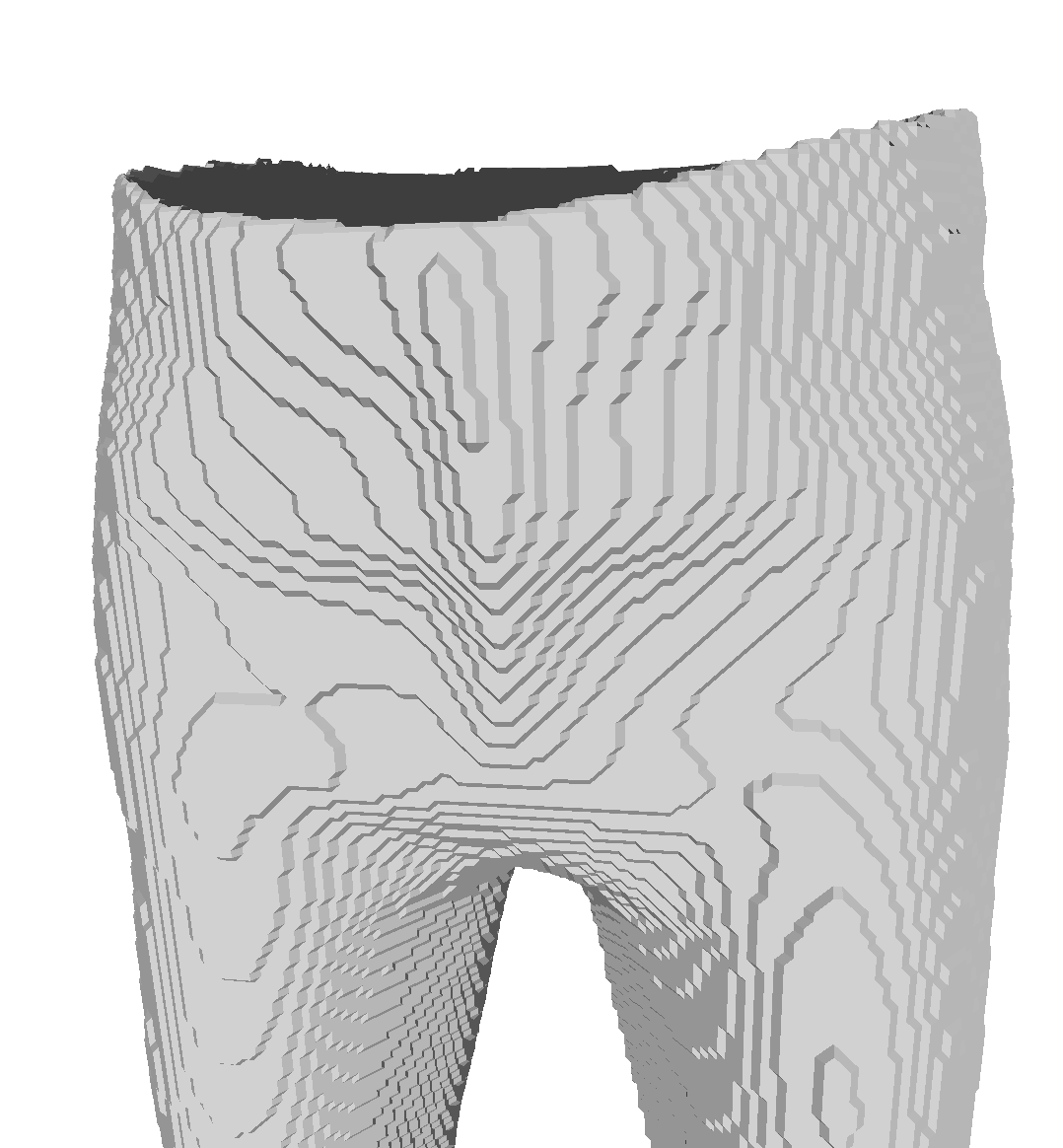}
\includegraphics[width=\subfigsize]{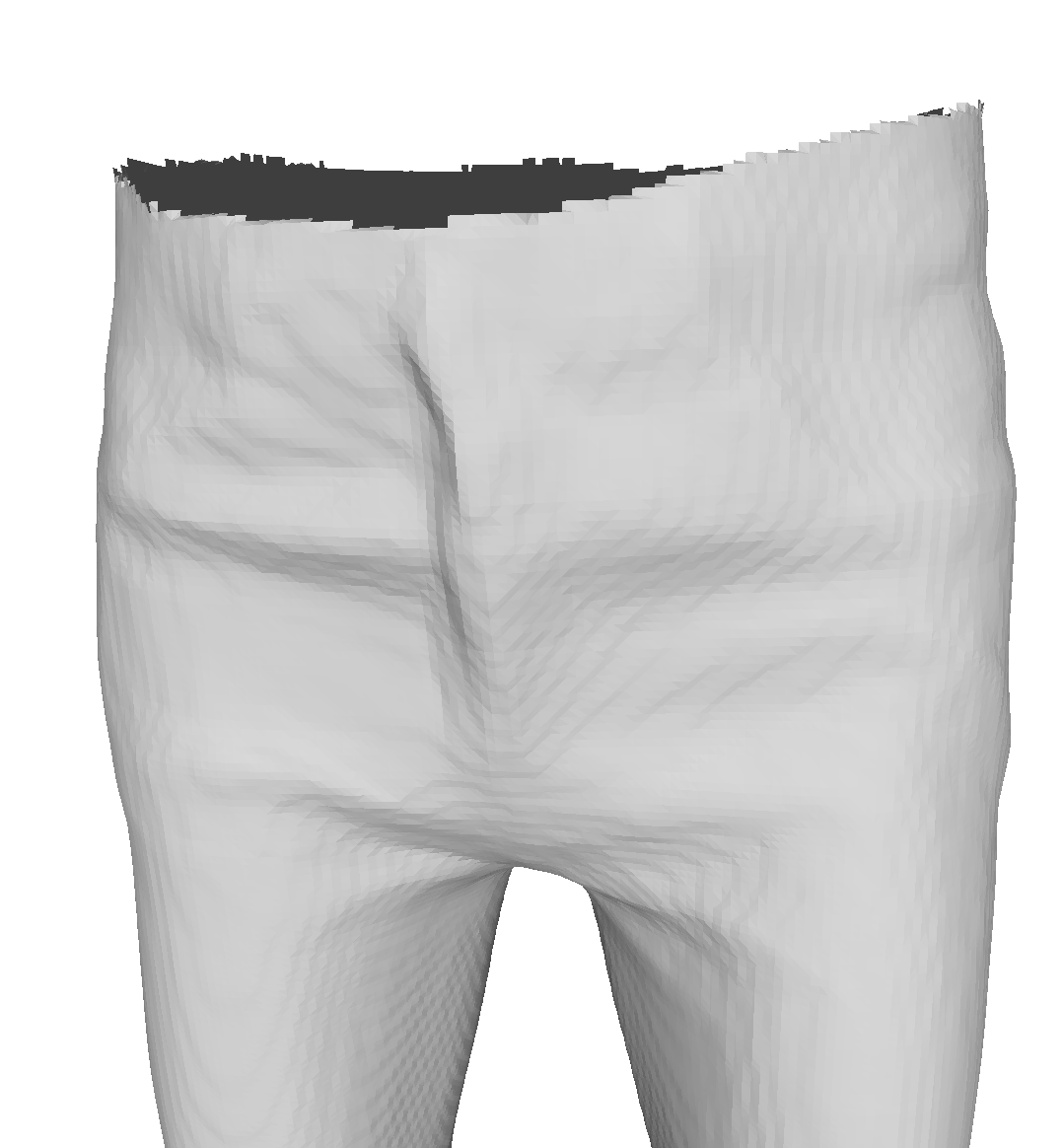}
\includegraphics[width=\subfigsize]{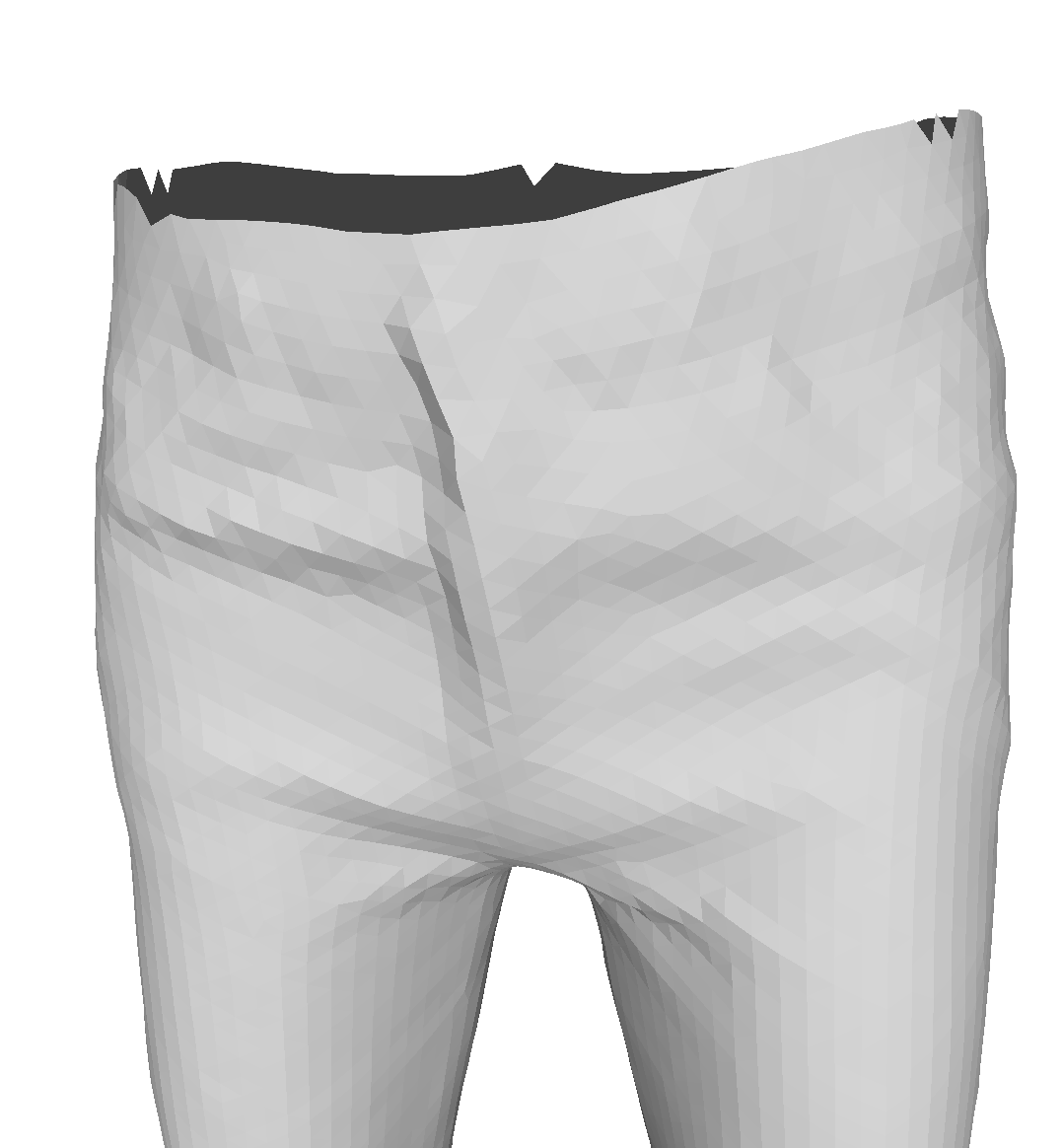}\\
\includegraphics[width=\subfigsize]{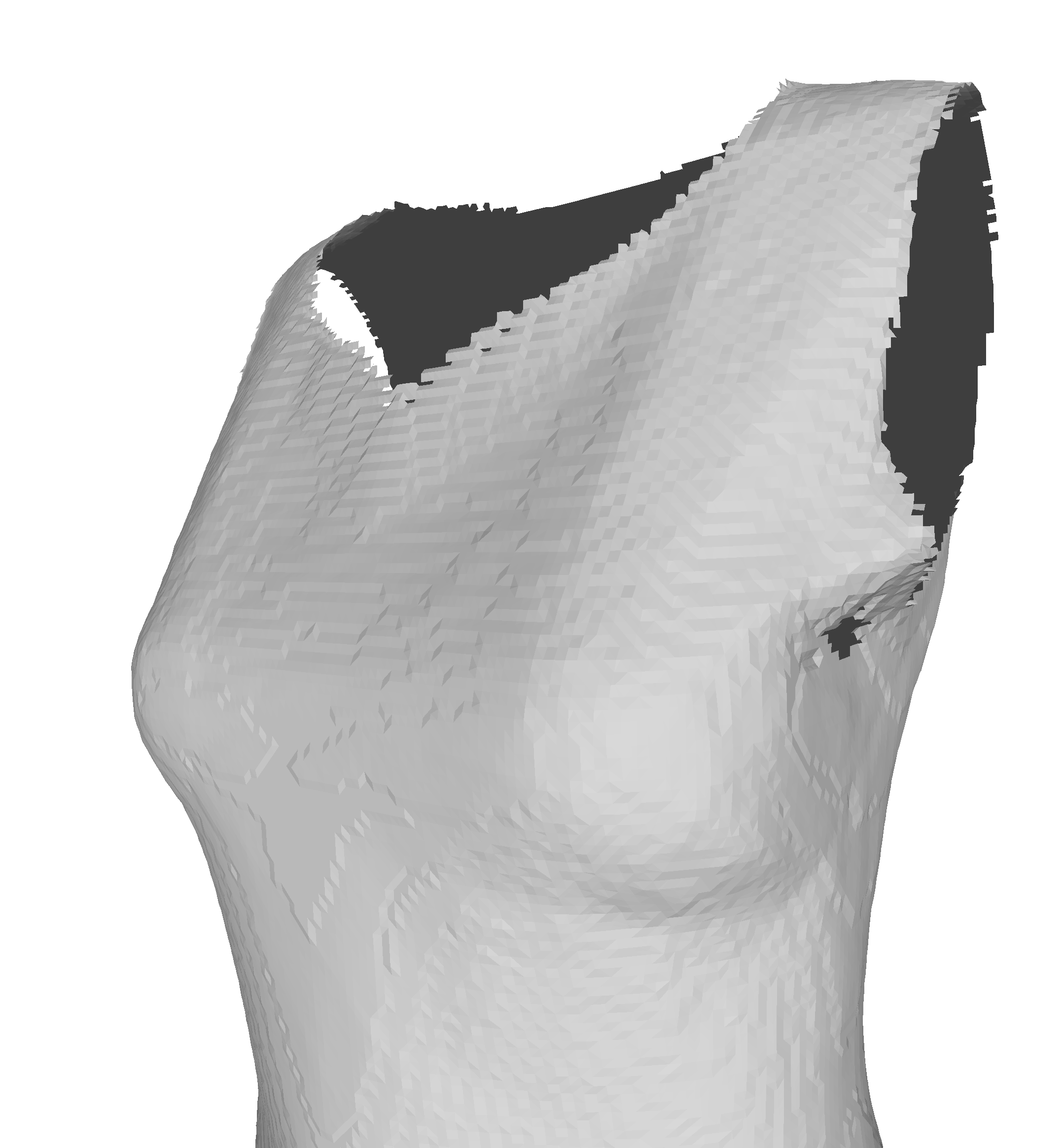}
\includegraphics[width=\subfigsize]{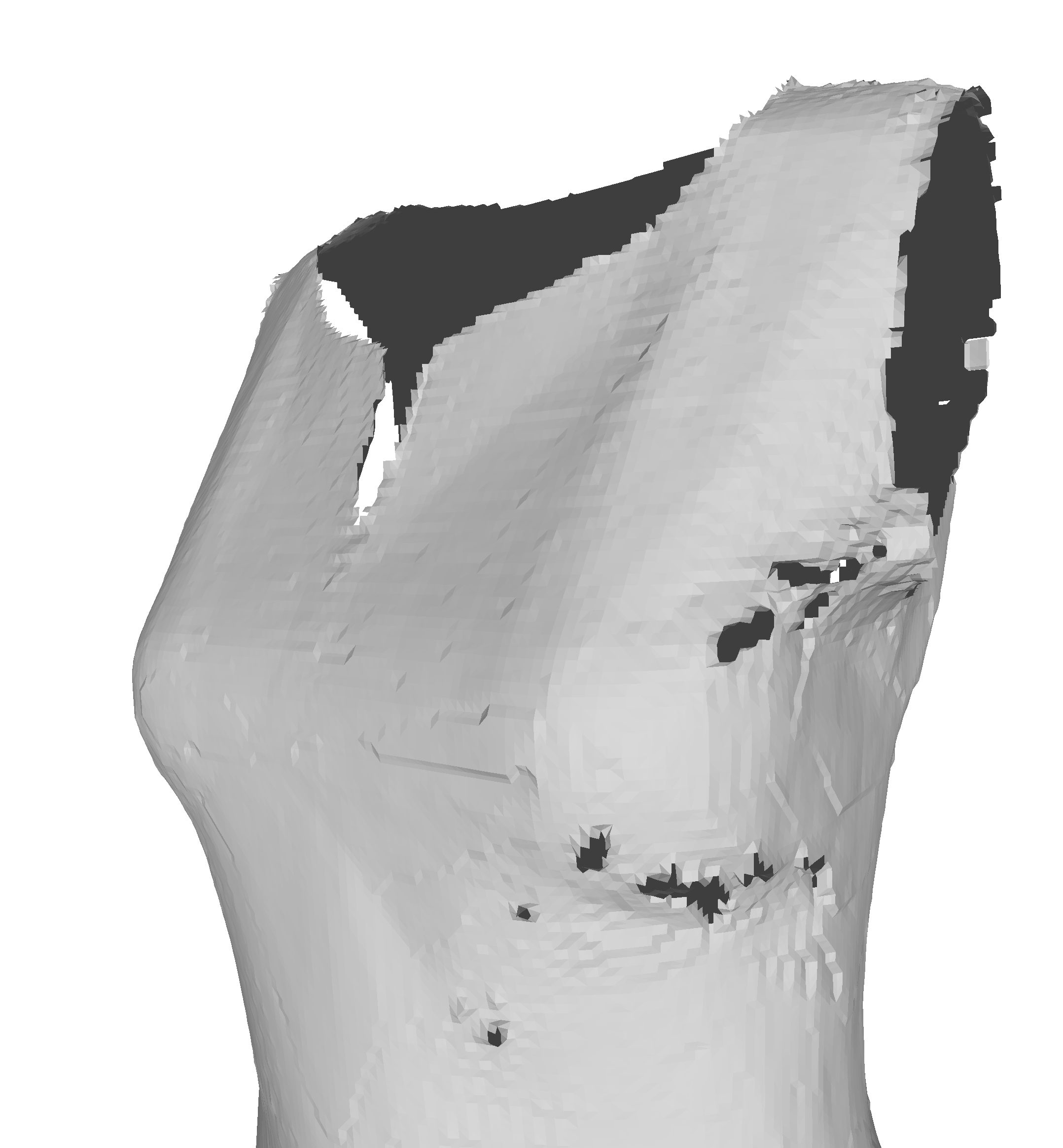}
\includegraphics[width=\subfigsize]{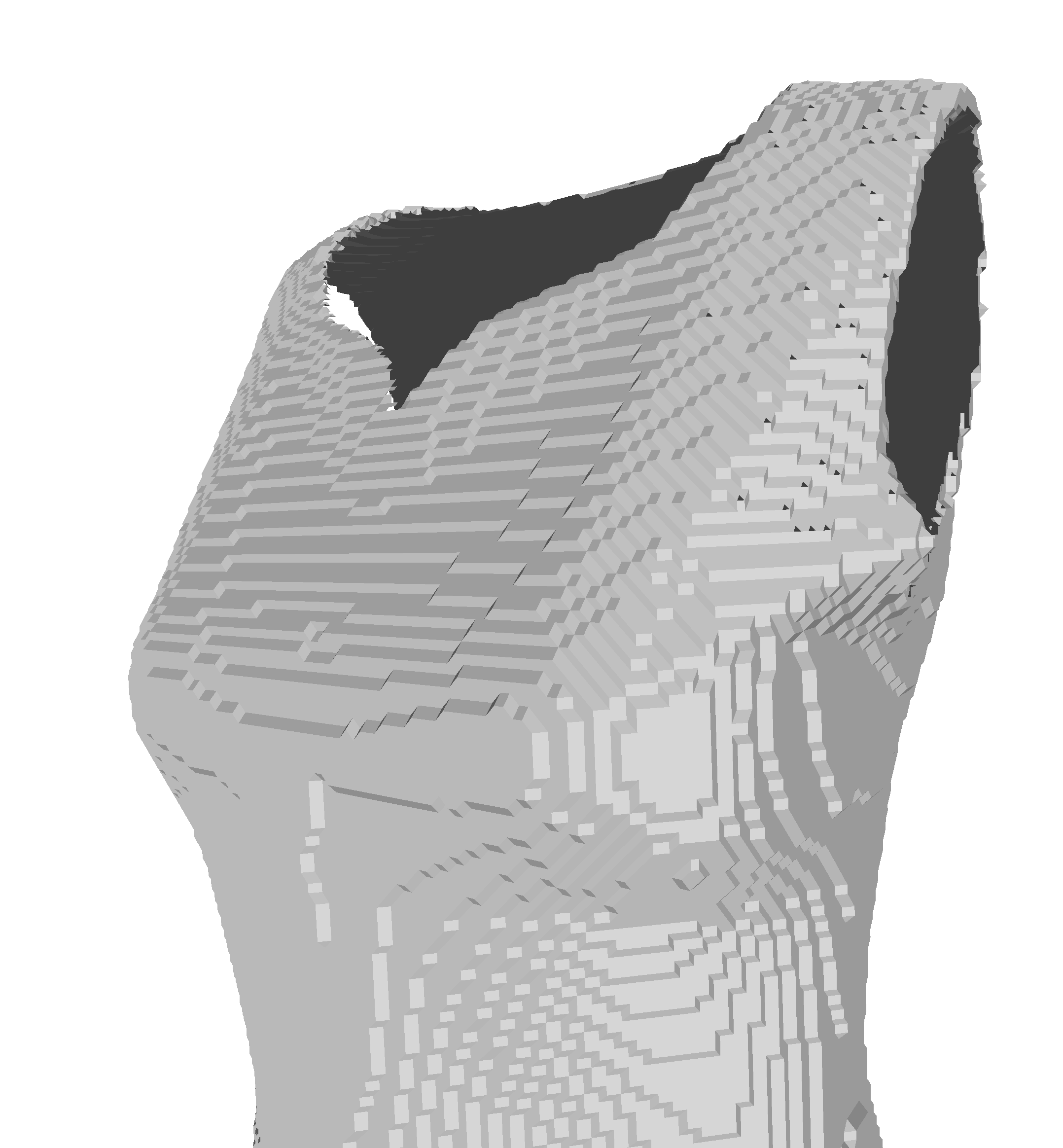}
\includegraphics[width=\subfigsize]{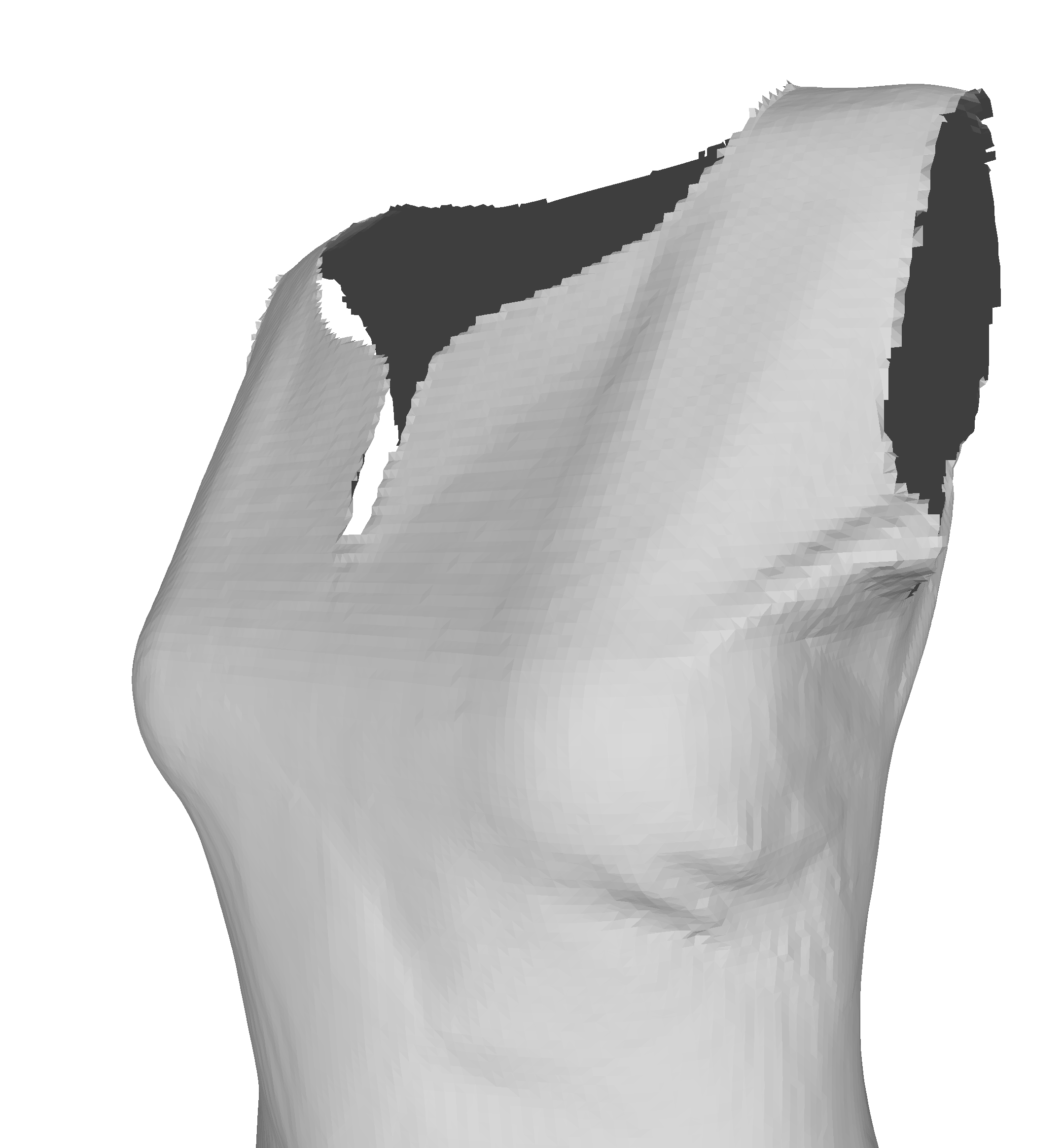}
\includegraphics[width=\subfigsize]{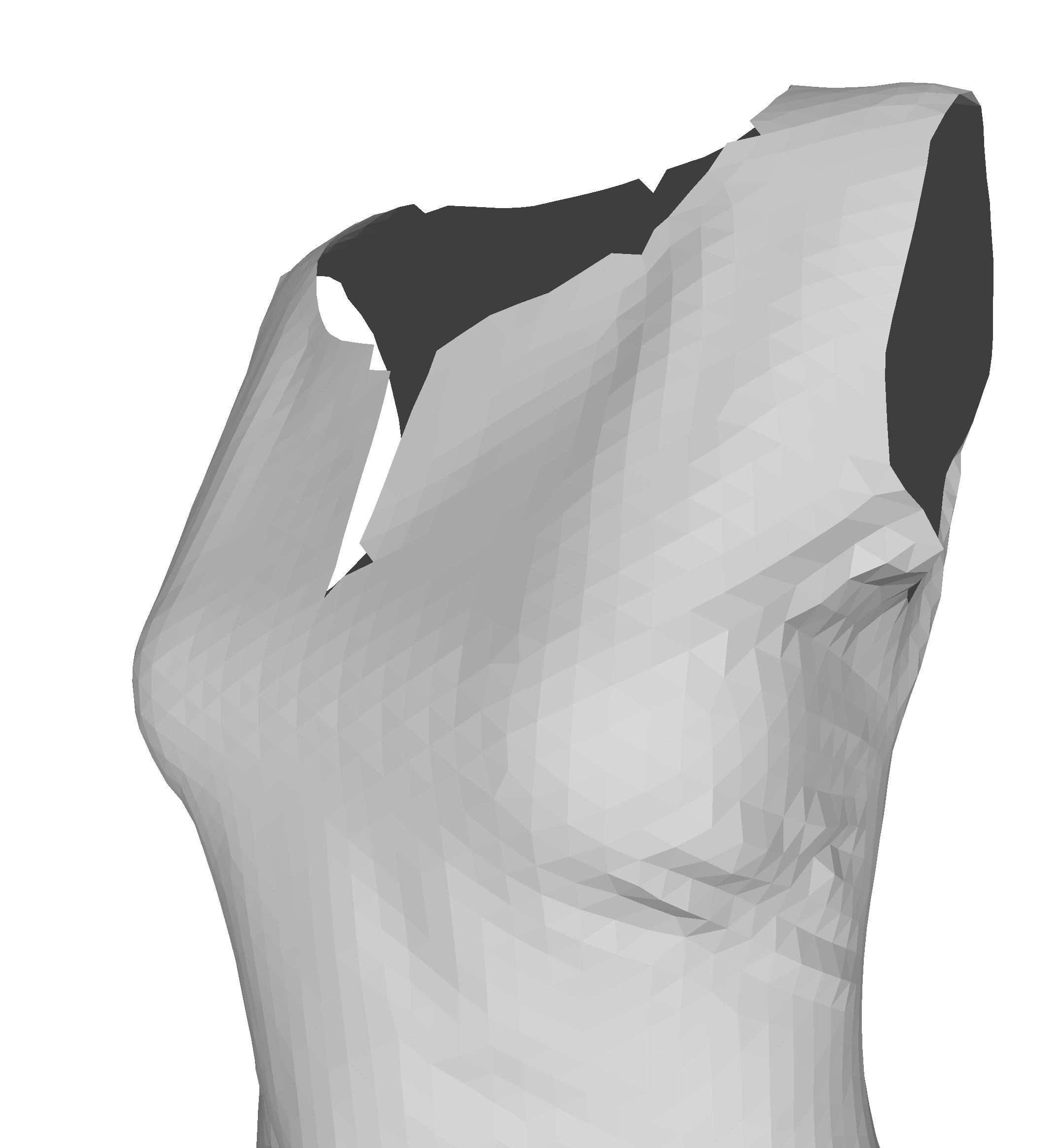}\\
\includegraphics[width=\subfigsize,height=3cm]{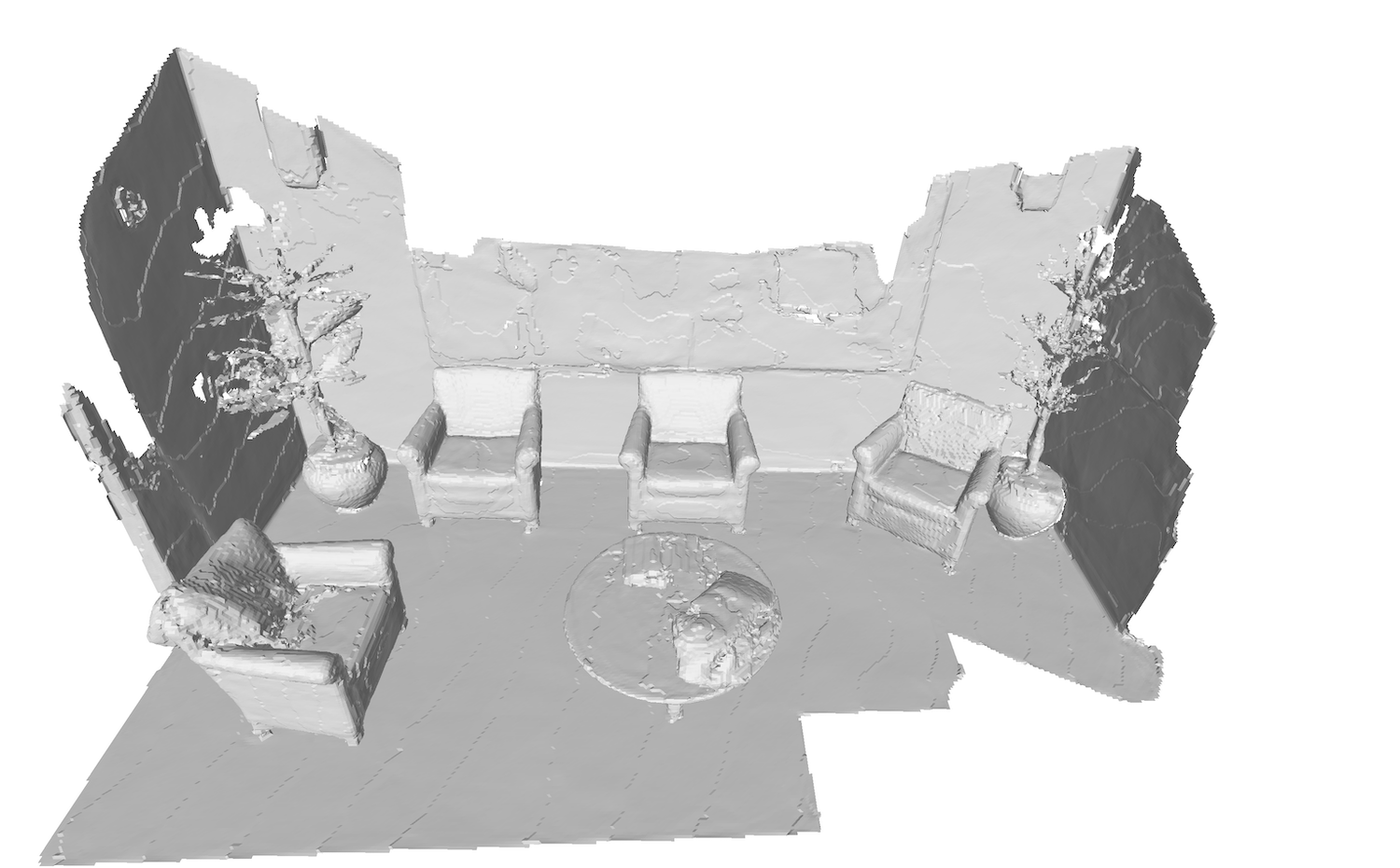}
\includegraphics[width=\subfigsize,height=3cm]{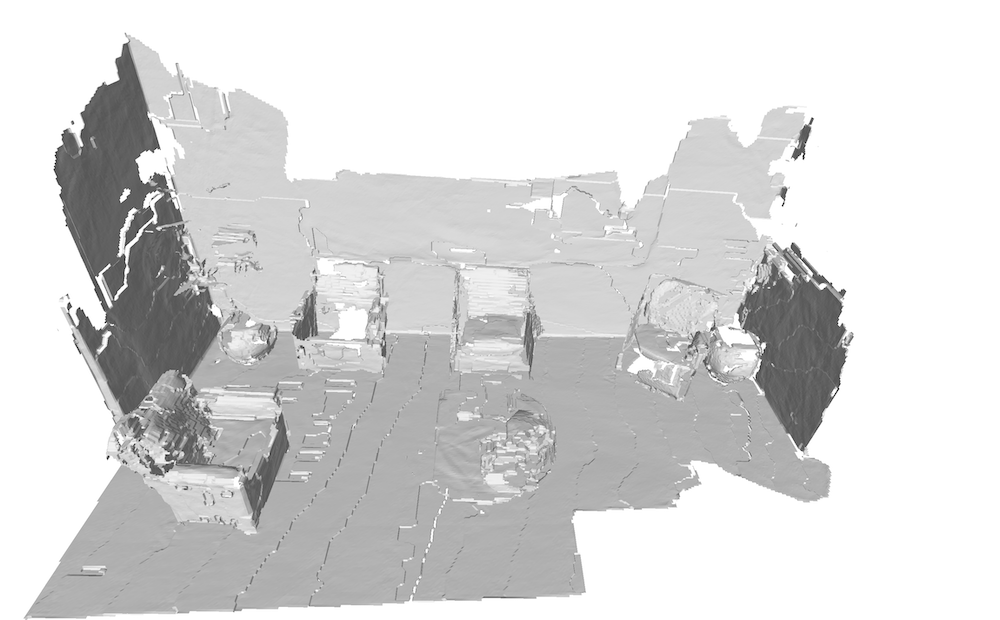}
\includegraphics[width=\subfigsize,height=3cm]{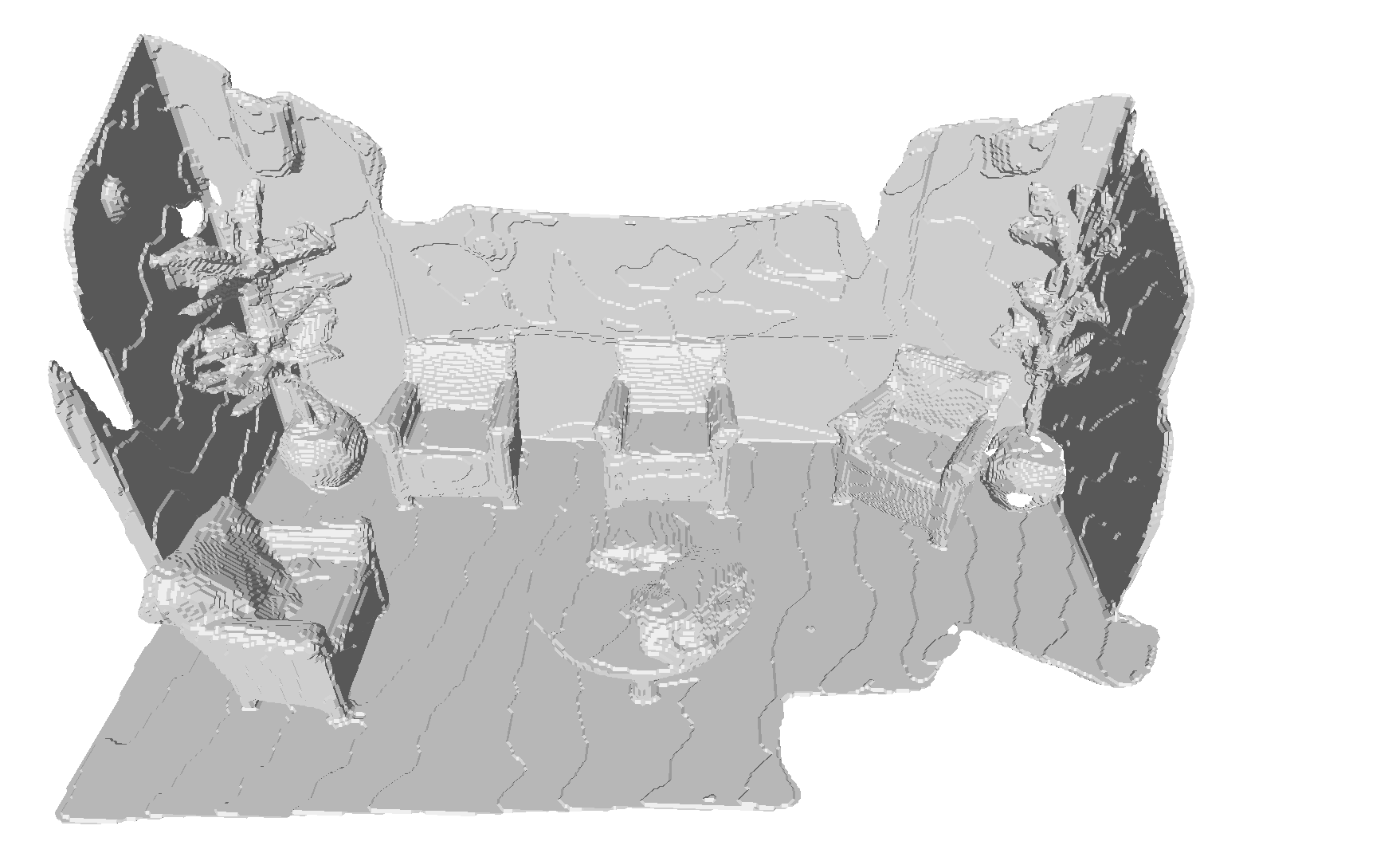}
\includegraphics[width=\subfigsize,height=3cm]{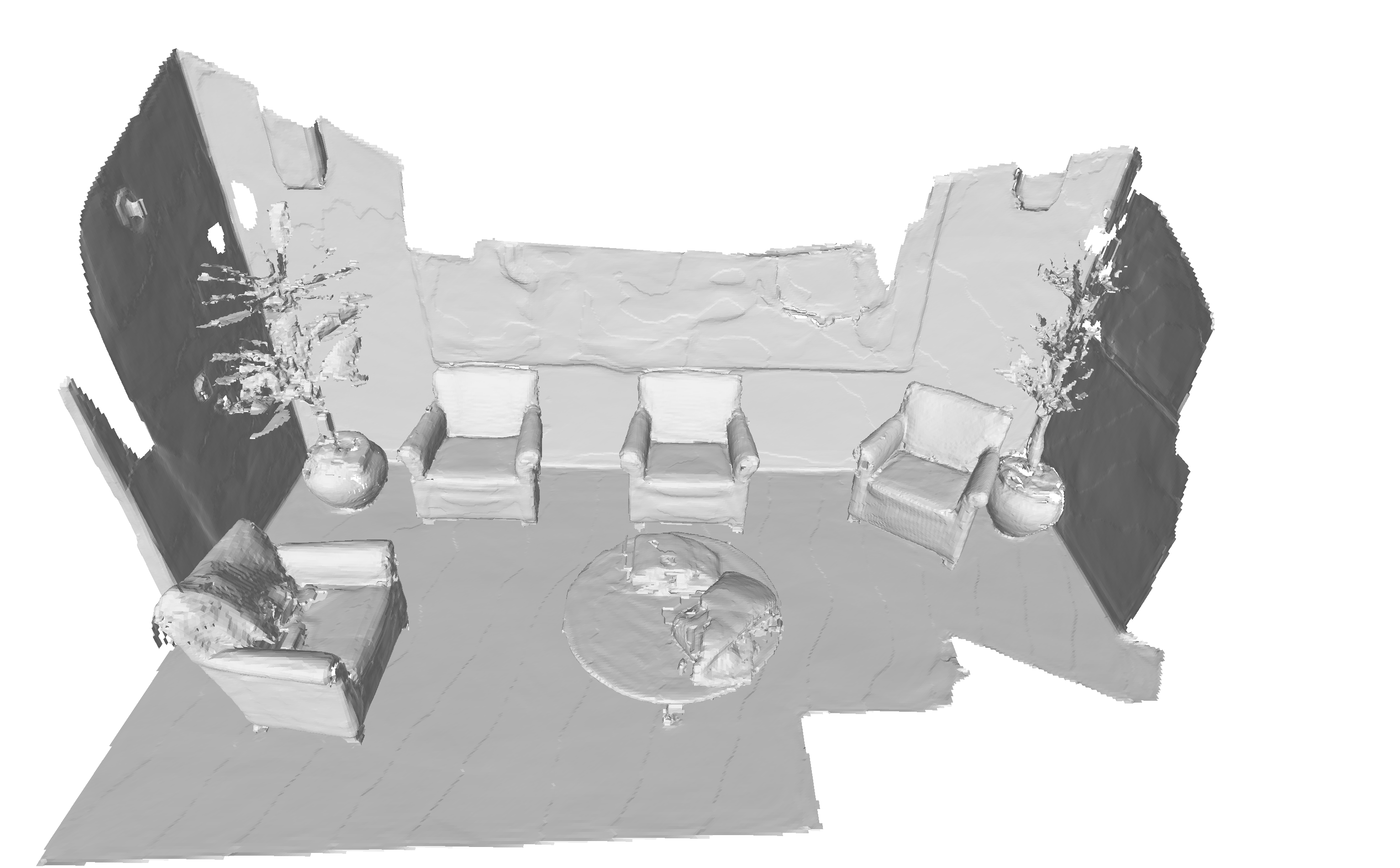}
\includegraphics[width=\subfigsize,height=3cm]{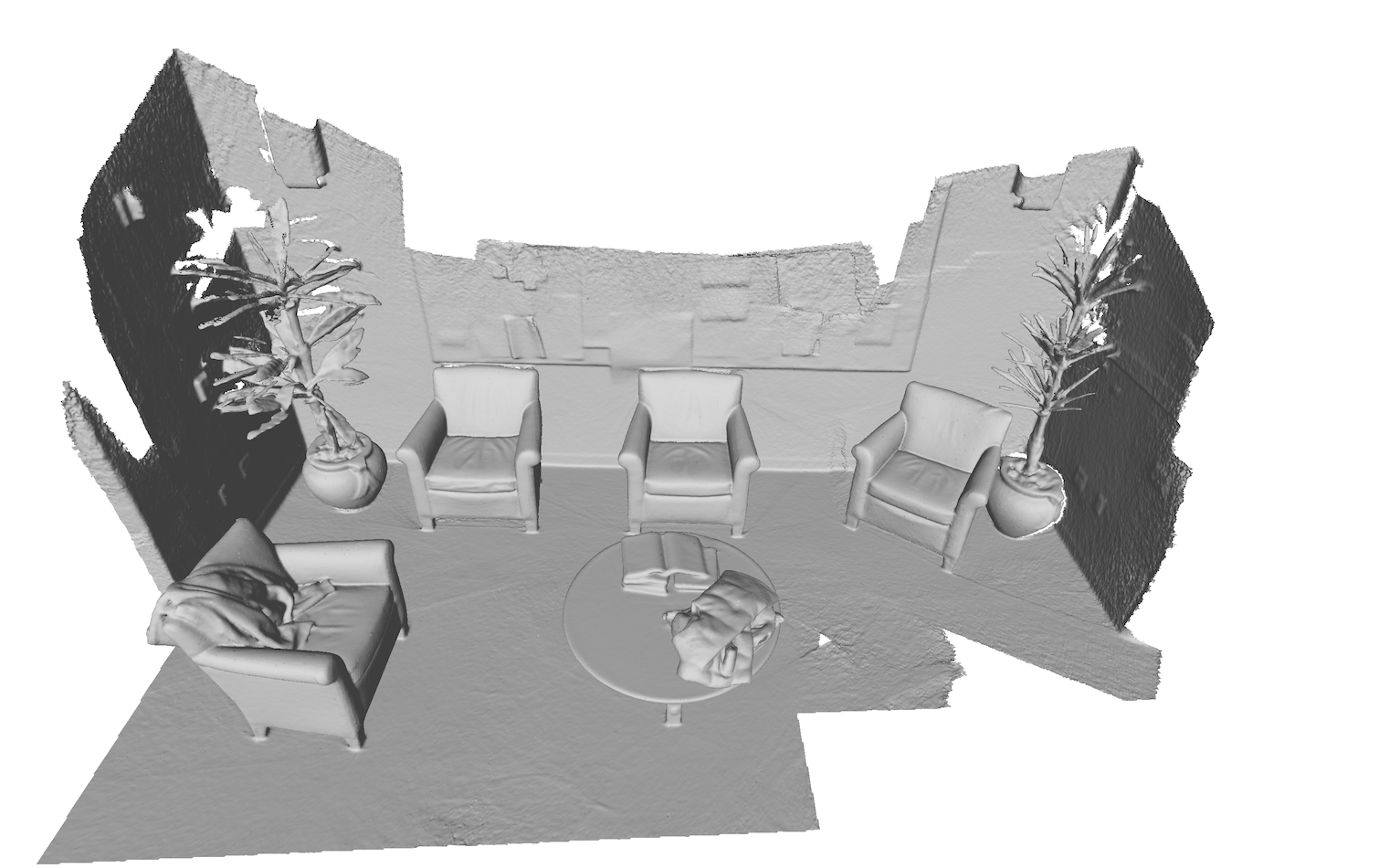}\\
\includegraphics[width=\subfigsize]{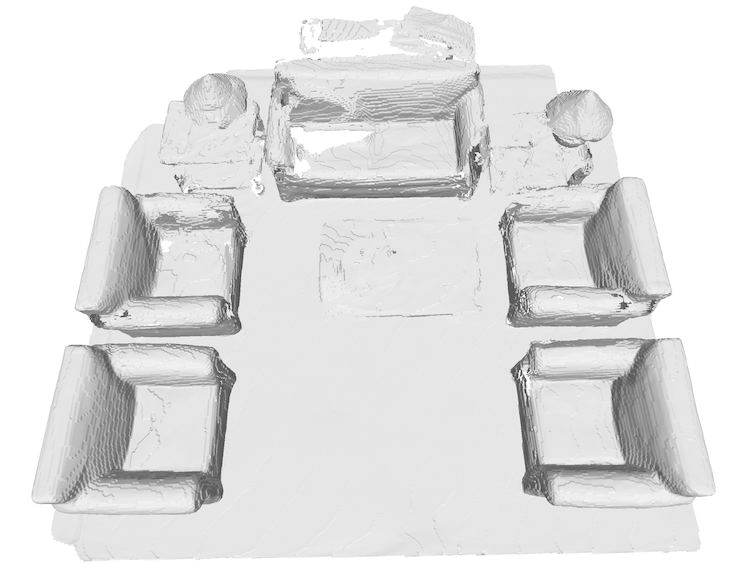}
\includegraphics[width=\subfigsize]{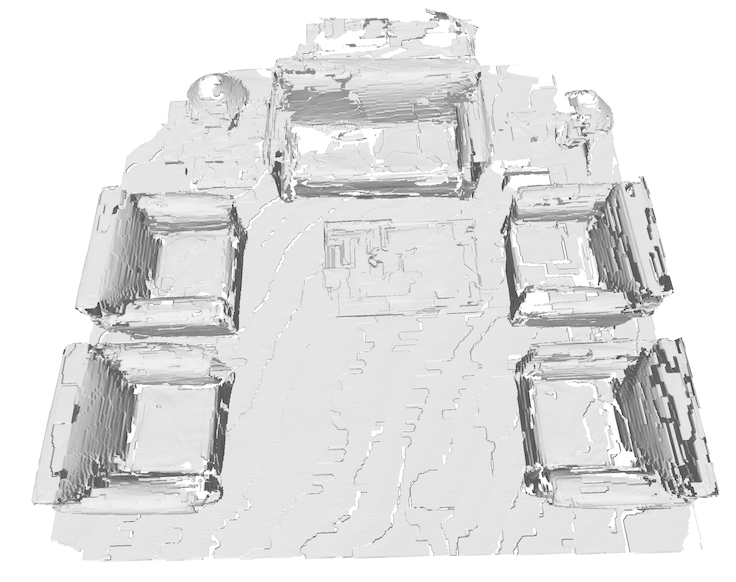}
\includegraphics[width=\subfigsize]{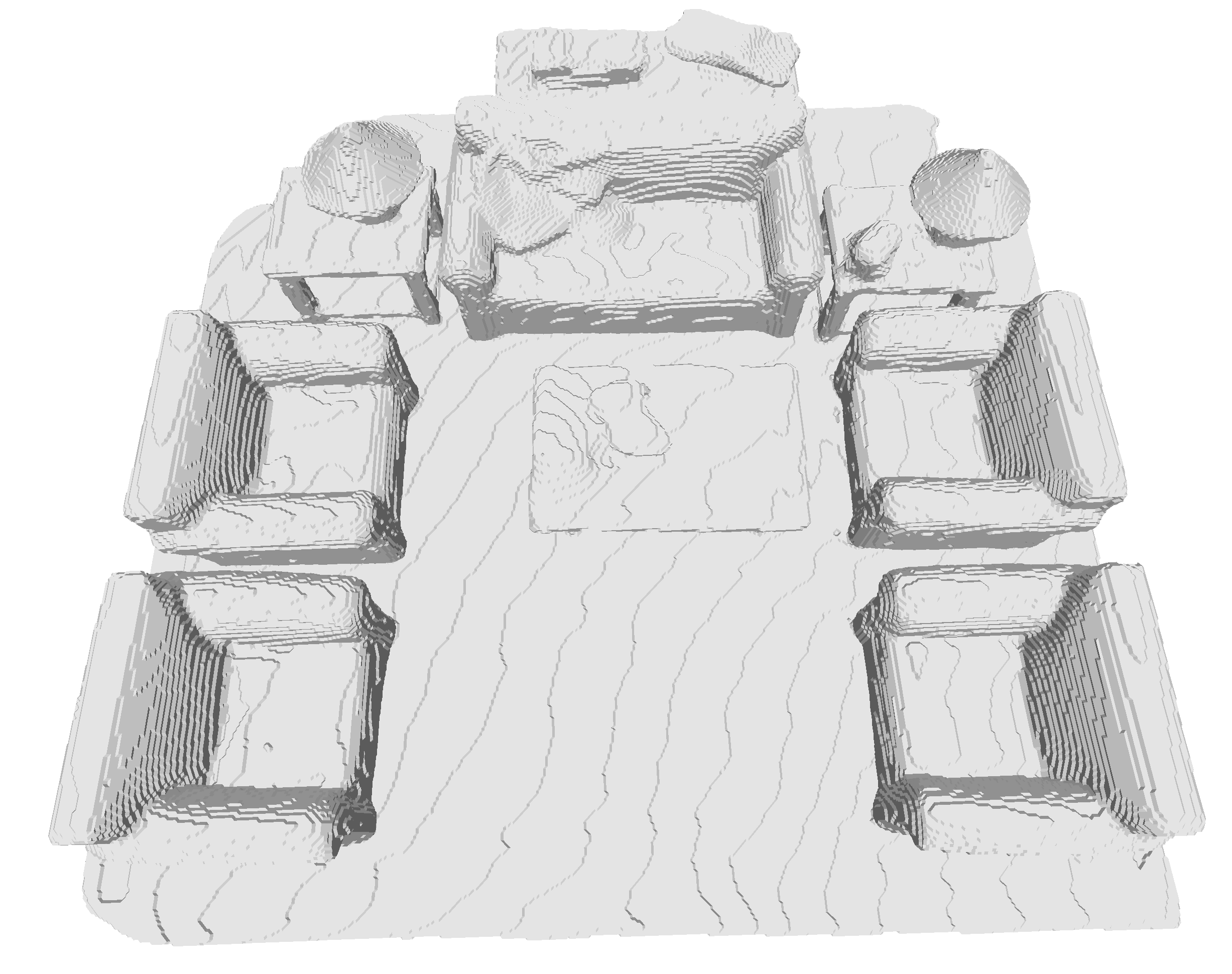}
\includegraphics[width=\subfigsize]{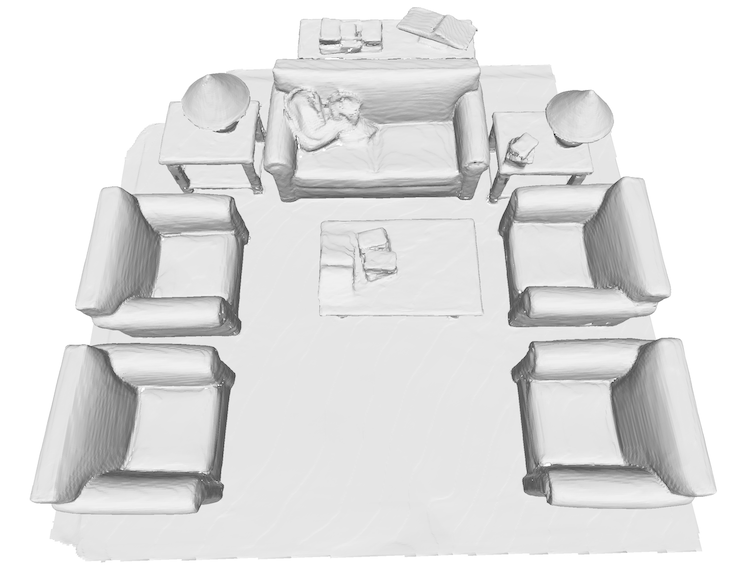}
\includegraphics[width=\subfigsize]{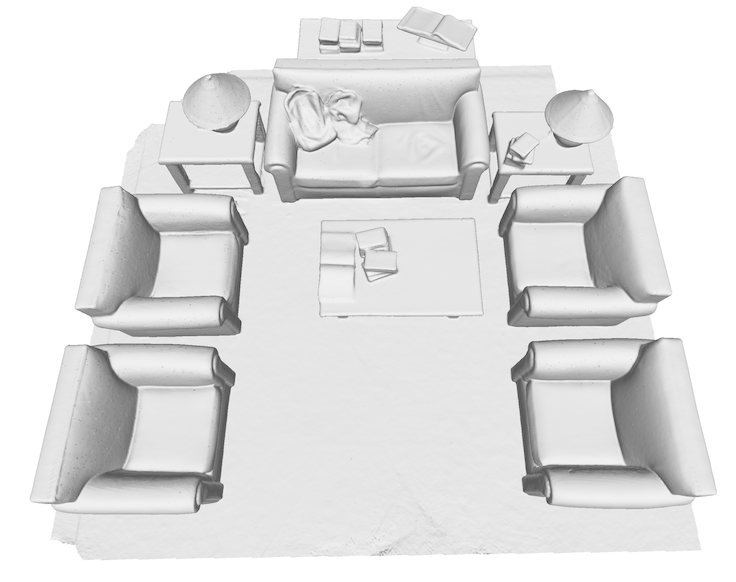}\\
\makebox[\subfigsize]{(a) NDF\cite{Chibane20b} }
\makebox[\subfigsize]{(b) CSP \cite{Venkatesh21}}
\makebox[\subfigsize]{(c) 3PSDF \cite{Chen22e} }
\makebox[\subfigsize]{(d) GDF (Ours) }
\makebox[\subfigsize]{(e) GT }
\end{center} 
\vspace{-5mm}
\caption{
\textbf{Single 3D Scene Reconstruction.} We train identical MLPs to reconstruct a single 3D surface using different implicit representations. The top two rows depict garments from the MGN dataset \cite{Bhatnagar19}. The bottom two  rows are examples from the 3D-Scene dataset \cite{Zhou13}. 
} 
\label{fig:single_mesh} 
\end{figure*}


We first evaluate all methods on a basic test aimed at reconstructing a single 3D shape via a deep network. In this test, we use an 8-layer multi-layer perceptron (MLP) with 512 feature maps per layer and train the network to regress the target representations for sampled query points (identical in all cases). For the MGN dataset, we train all networks for 30000 iterations. For 3D-Scene, which is significantly more complex, we train them for 100000 iterations. All outputs are meshed at $512^3$ resolutions for evaluation.

In Table \ref{tab:single-mesh}, we report the Chamfer Distances $\times 10^4$ (CD) and the Normal Consistency of the resulting surfaces from different methods. Our proposed GDF produces smooth meshes that closely match the ground-truth ones in all cases, as shown in Fig. \ref{fig:single_mesh}. This is because surfaces in our GDF representation locate in continuous, differentiable parts of the scalar field, enabling the deep network to easily memorize them. On the contrary, the results of NDF exhibit staircase artifacts and surface holes, typical issues associated with the direct learning of UDFs. CSP creates smoother surfaces. However, they have many holes, despite being trained for an identical number of iterations as other methods. Compared to ours, this representation is more difficult to learn as it does not exhibit distance-like scalar field. The 3PSDF representation makes network 
training easier with fast convergence and no holes in the reconstructed surfaces, even for challenging cases such as the "reading room" of the 3D-Scene dataset shown in bottom row of the figure. However, this efficiency comes with a cost of discretizing distance values into either ``-1'' or ``1,'' resulting in less accurate meshes with noticeable staircase artifacts.

\begin{table}[t]
    \renewcommand{\arraystretch}{1.0}
    \caption{\small \textbf{Fitting to a  single complex mesh.} Average L2 Chamfer Distance $\times 10^{-4}$ (CD) and Normal Consistency (NC) for the MGN and 3D-Scenes datasets. We use the same MLP for each scene and only change the output representation. 
    }
    \vspace{-5mm}
    \label{tab:single-mesh}
    \begin{small}
        \begin{center}
            \setlength{\tabcolsep}{3pt}
            \begin{tabular}{ccc|cc} 
                \multicolumn{1}{l}{} & \multicolumn{2}{c}{\textbf{Garments}}  & \multicolumn{2}{c}{\textbf{3D Scenes}} \\ \midrule
                \multicolumn{1}{c|}{} & \textit{CD} ($\downarrow$) & \textit{NC}($\uparrow$) & \textit{CD} ($\downarrow$) & \textit{NC}($\uparrow$) \\ \midrule
                \multicolumn{1}{c|}{NDF\cite{Chibane20b}}        & 0.22 & 99.10 & 0.27 & 85.7 \\
				\multicolumn{1}{c|}{CSP\cite{Venkatesh21}}        & 0.22 & 98.38 & 3.01 & 71.5 \\
                \multicolumn{1}{c|}{3PSDF\cite{Chen22e}}        & 0.32 & 96.73 & 0.60 & 83.2 \\
				\midrule
                \multicolumn{1}{c|}{GDF (Ours)} & \textbf{0.21} & \textbf{99.16} & \textbf{0.20} & \textbf{87.1} \\ \midrule
            \end{tabular}
        \end{center}
    \end{small}
\end{table}

.

\subsection{Auto-Decoders to Model a Category of Shapes}

Auto-decoders were introduced in~\cite{Park19c} and are now an important 3D shape modeling tool. They make it possible to represent multiple 3D shapes from a given category using a single network, each one being characterized by a latent vector. The auto-decoder learns a shape prior for the category, enabling the reconstruction of unseen instances from partial input observations. We evaluate two aspects of using auto-decoder in conjunction with our proposed implicit representation: the capacity of the network to memorize training instances and the modeling of previously unseen data given 3D point clouds.

\parag{Representing Known Shapes.}
We train auto-decoders on three datasets of increasing complexity: MGN, ShapeNet-Car(P), and ShapeNet-Car(R).
We use 12-layer MLPs for MGN and ShapeNet-Car(P) and 18-layer MLPs for ShapeNet-Car(R) with 1024 feature maps per layer. They are trained by minimizing the reconstruction loss of Eq.~\ref{eq:loss}. 
The code length is set to 512 and the resolution of the output meshes is $256^3$ in all cases.

We report quantitative comparative results in Table \ref{tab:auto-decoder}. GDF yields significantly more accurate meshes than the other methods in terms of both metrics. Fig. \ref{fig:auto-decoder} features four reconstructed cars. The GDF reconstructions are smooth with accurate details, while the others exhibit some of the same issues as in the single-scene examples of Fig.~\ref{fig:single_mesh}.


\begin{table}[t]
    \caption{\small \textbf{Representing known shapes using an auto-decoder.} Average L2 Chamfer Distance $\times 10^{-4}$ (CD) and normal consistency (NC) for the MGN and ShapeNet-Car datasets. }
    \vspace{-5mm}
    \label{tab:auto-decoder}
    \begin{small}
        \begin{center}
            \setlength{\tabcolsep}{3pt}
            \begin{tabular}{ccc|cc|cc} 
                \multicolumn{1}{l}{} & \multicolumn{2}{c}{\textbf{Garments}}  & \multicolumn{2}{c}{\textbf{SN-Car (P)}} & \multicolumn{2}{c}{\textbf{SN-Car (R)}} \\ \midrule
                \multicolumn{1}{c|}{} & \textit{CD} ($\downarrow$) & \textit{NC}($\uparrow$) & \textit{CD} ($\downarrow$) & \textit{NC}($\uparrow$) & \textit{CD} ($\downarrow$) & \textit{NC}($\uparrow$) \\ \midrule
                \multicolumn{1}{c|}{NDF\cite{Chibane20b}}        & 0.47 & 95.42 & 0.88 & 93.79 &0.76 & 58.73\\
				\multicolumn{1}{c|}{CSP\cite{Venkatesh21}}        & 1.07 & 92.72 & 0.95 & 90.17 &0.85  & 55.37 \\
                \multicolumn{1}{c|}{3PSDF\cite{Chen22e}}      & 0.41  & 91.52 & 0.69 & 91.89&0.54 & 57.35     \\
				\midrule
                \multicolumn{1}{c|}{GDF (Ours)} & \textbf{0.32} & \textbf{97.15} & \textbf{0.67} & \textbf{93.87} &\textbf{0.34} & \textbf{62.66} \\ \midrule
            \end{tabular}
        \end{center}
    \end{small}
\end{table}

\def\subfigsize{0.191\linewidth}
\begin{figure*}[h]
\begin{center}

\includegraphics[width=\subfigsize]{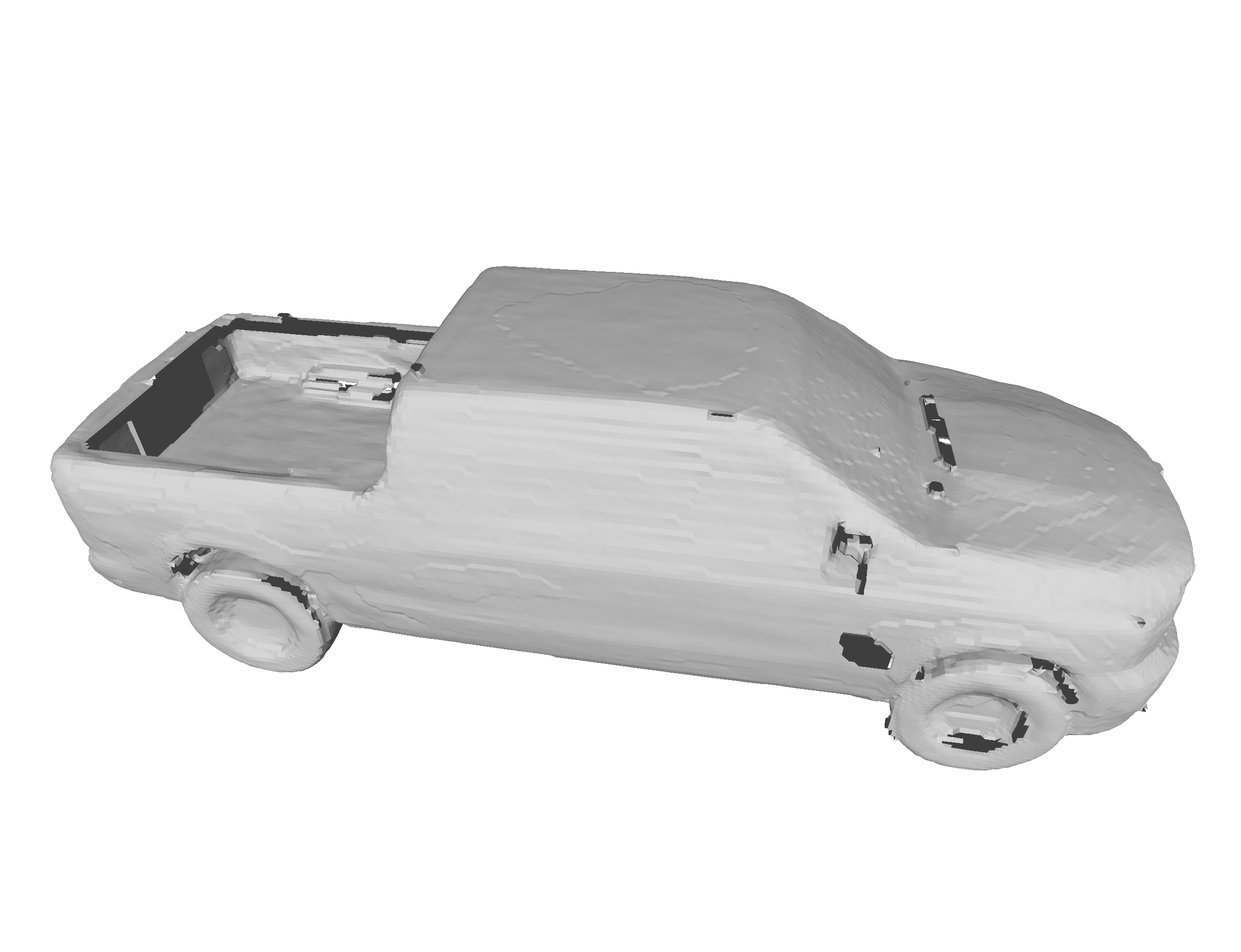}
\includegraphics[width=\subfigsize]{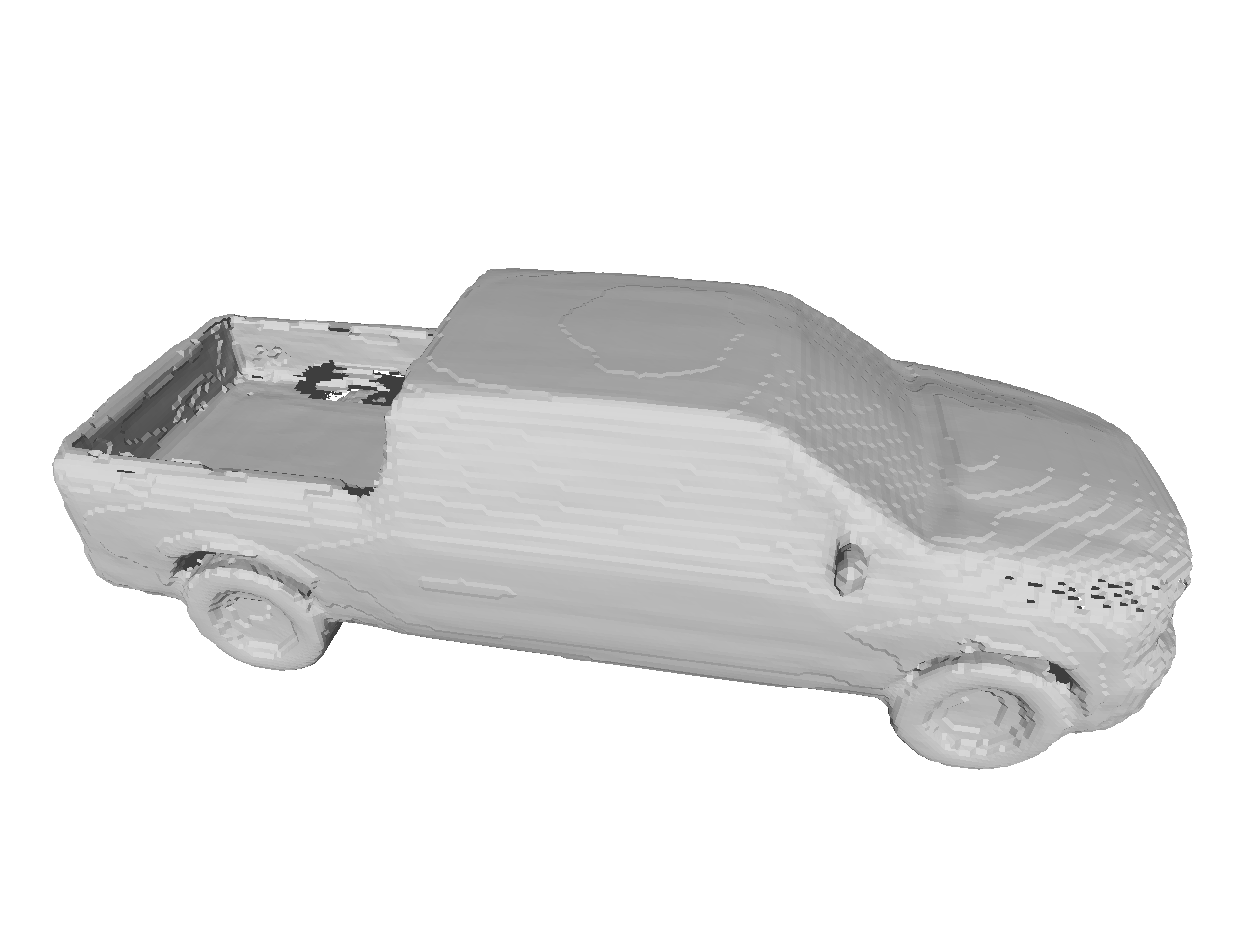}
\includegraphics[width=\subfigsize]{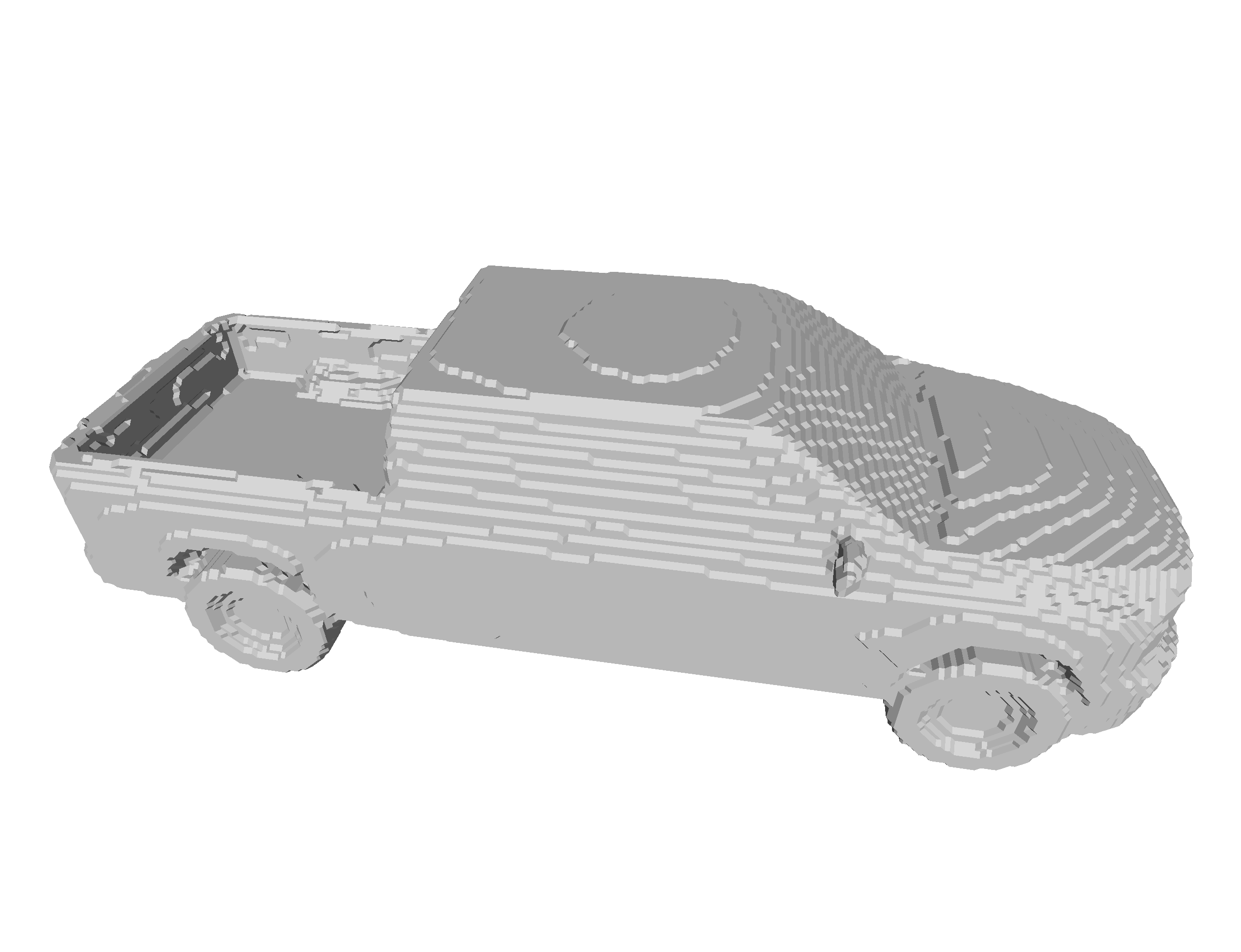}
\includegraphics[width=\subfigsize]{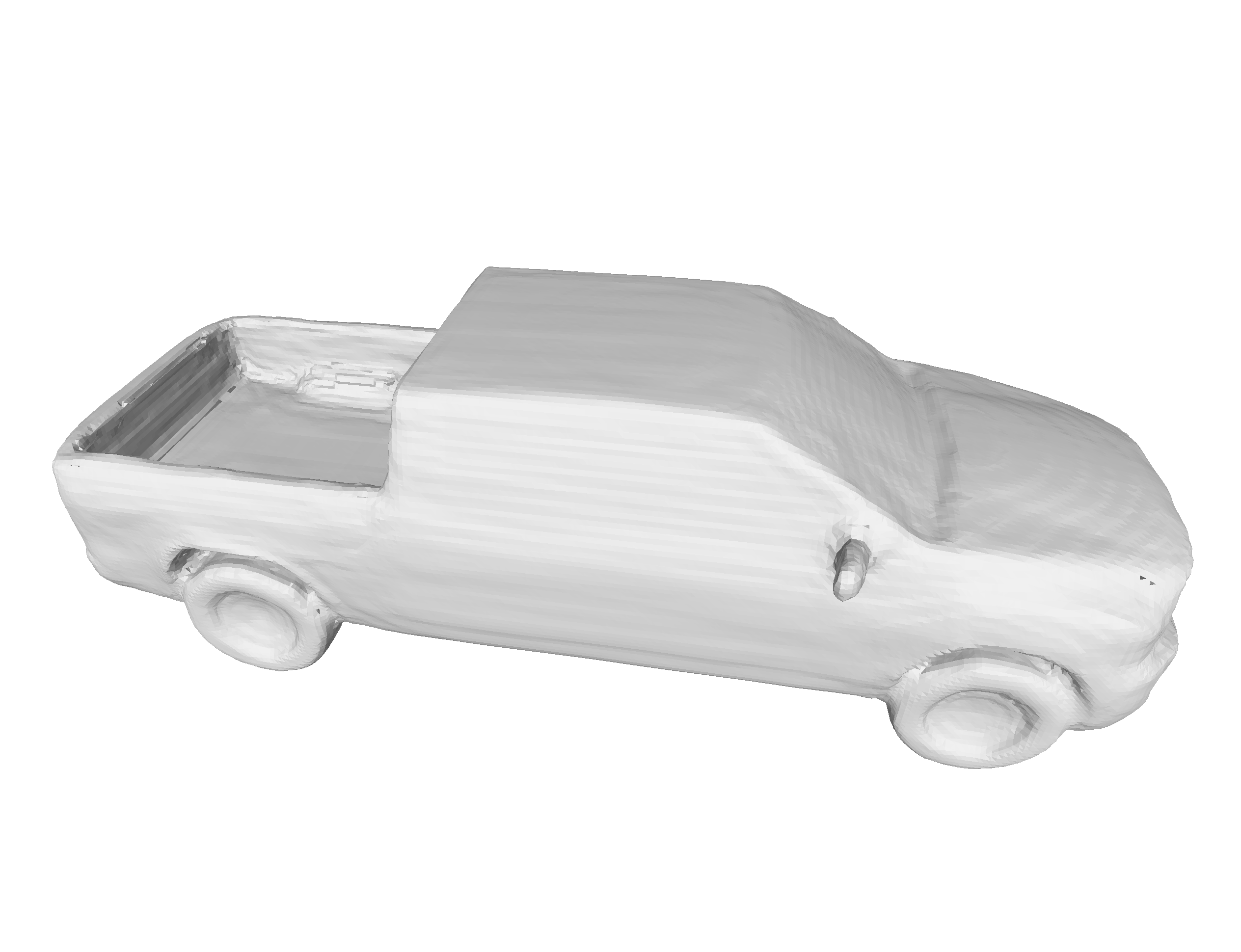}
\includegraphics[width=\subfigsize]{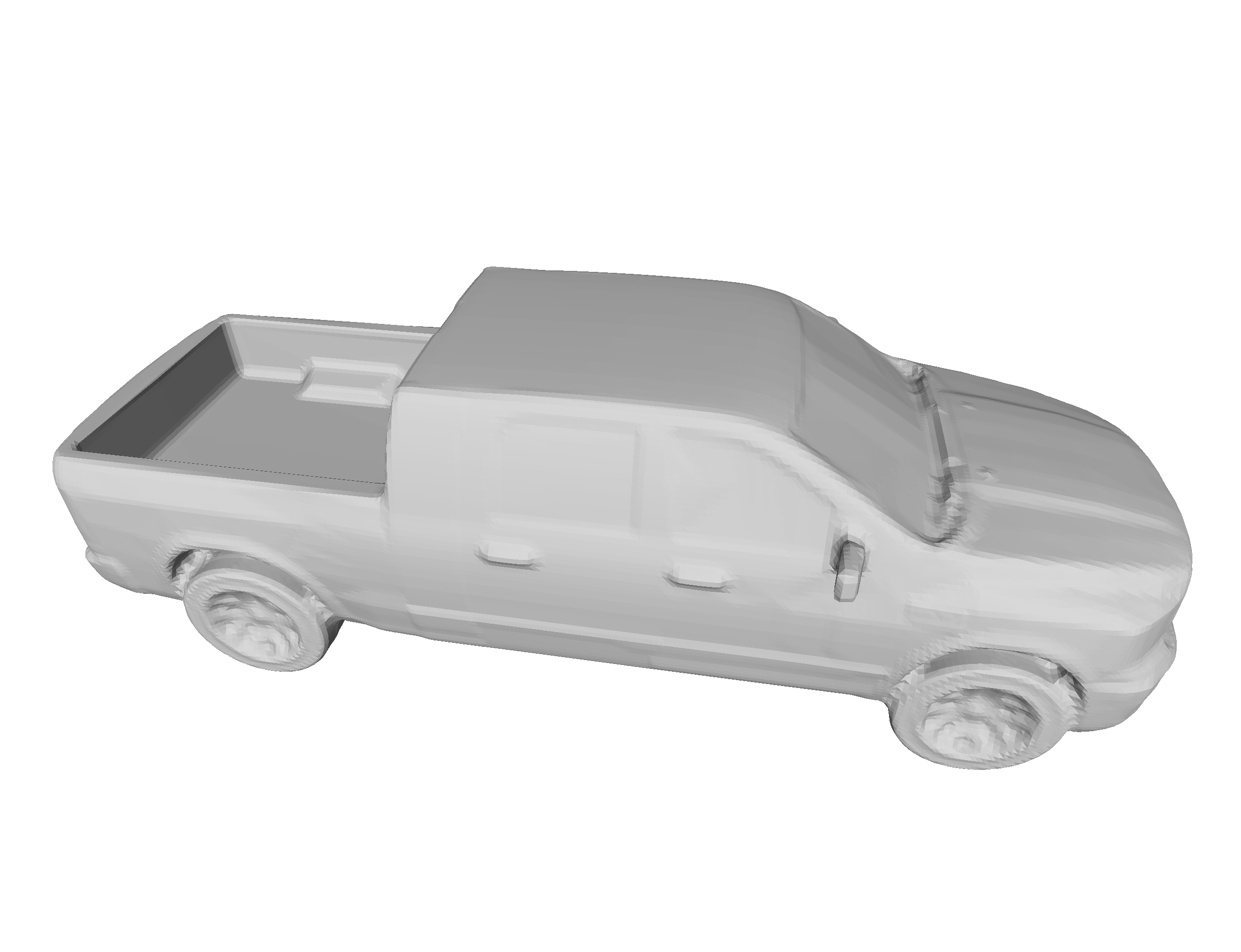}\\
\includegraphics[width=\subfigsize]{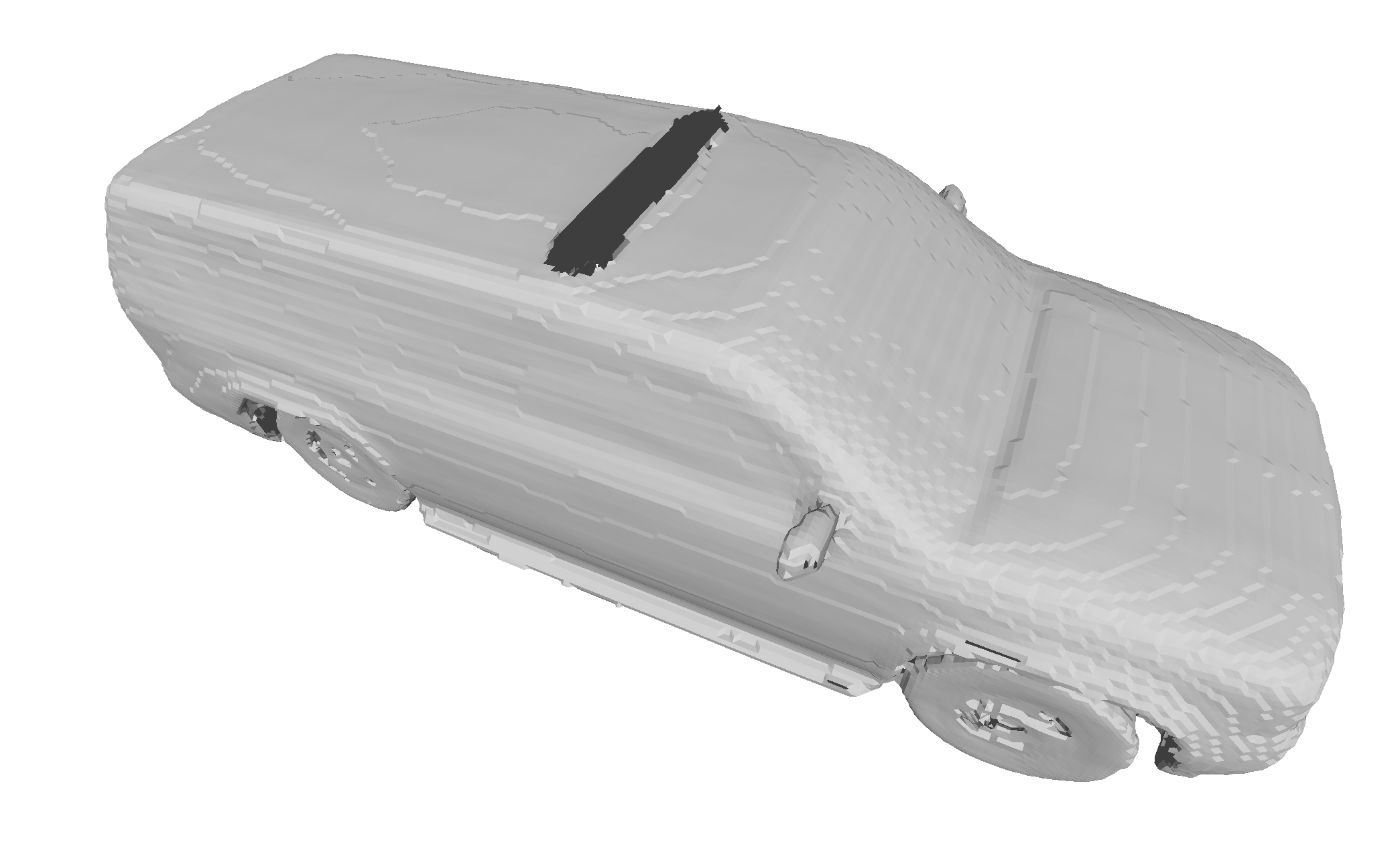}
\includegraphics[width=\subfigsize]{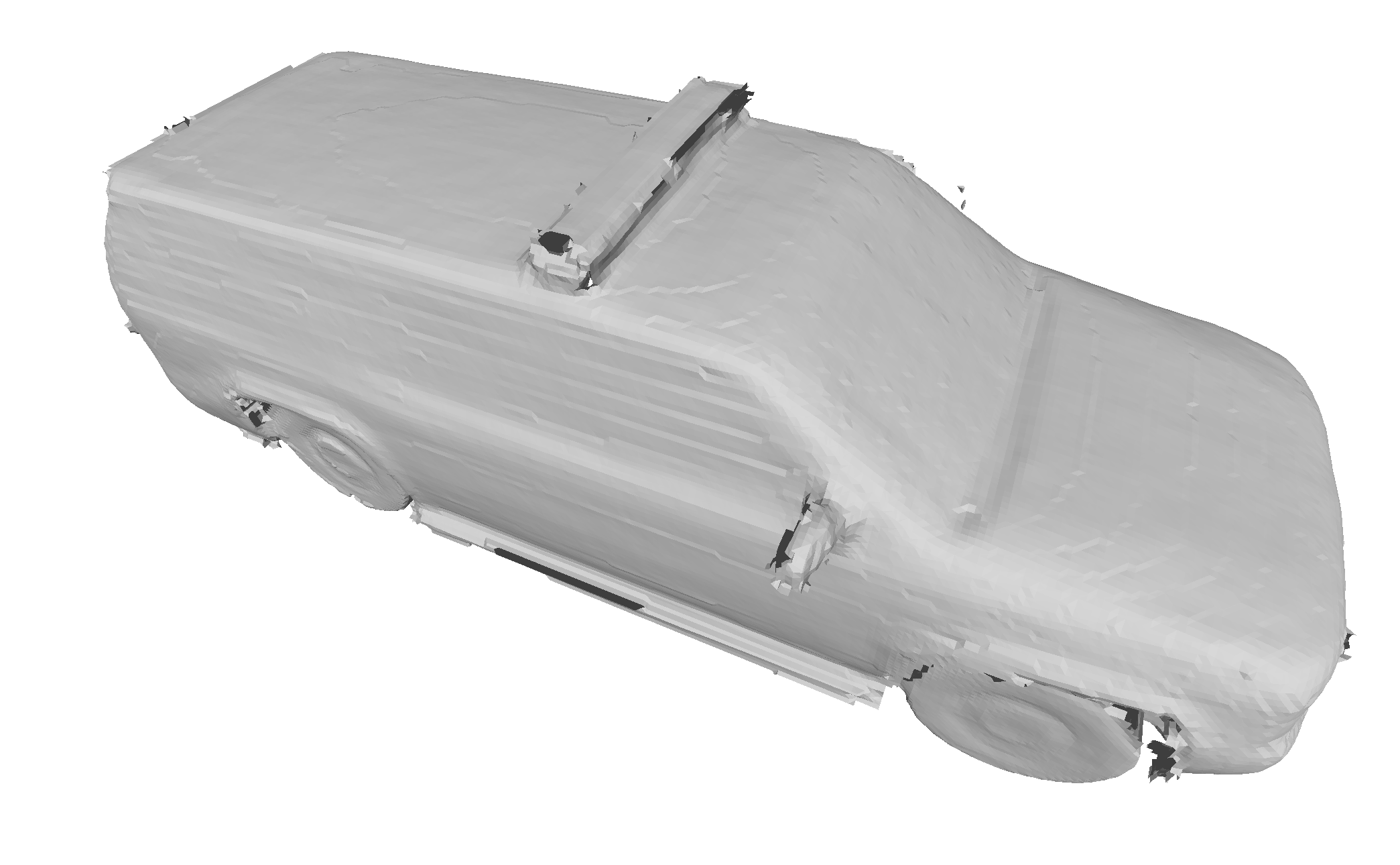}
\includegraphics[width=\subfigsize]{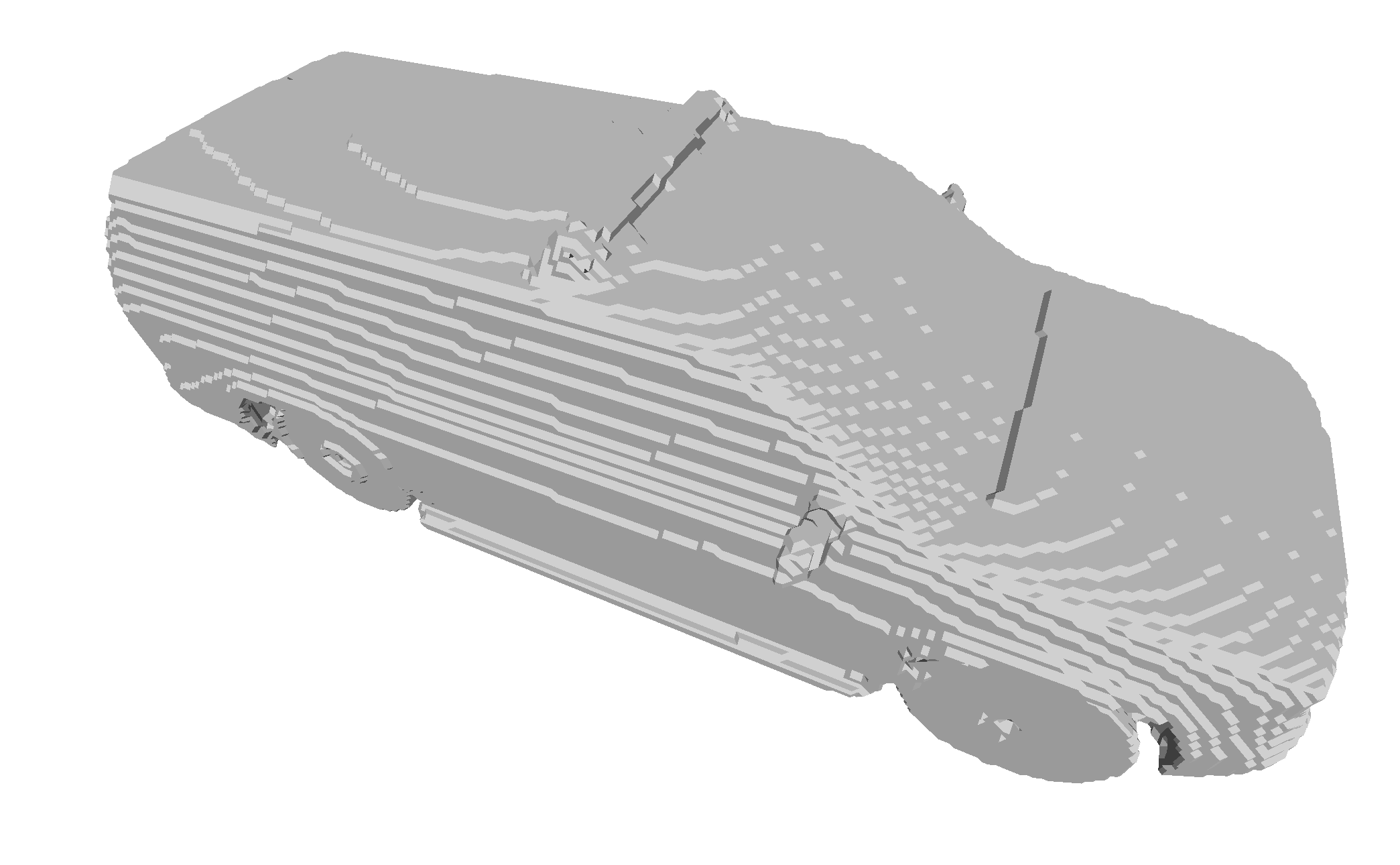}
\includegraphics[width=\subfigsize]{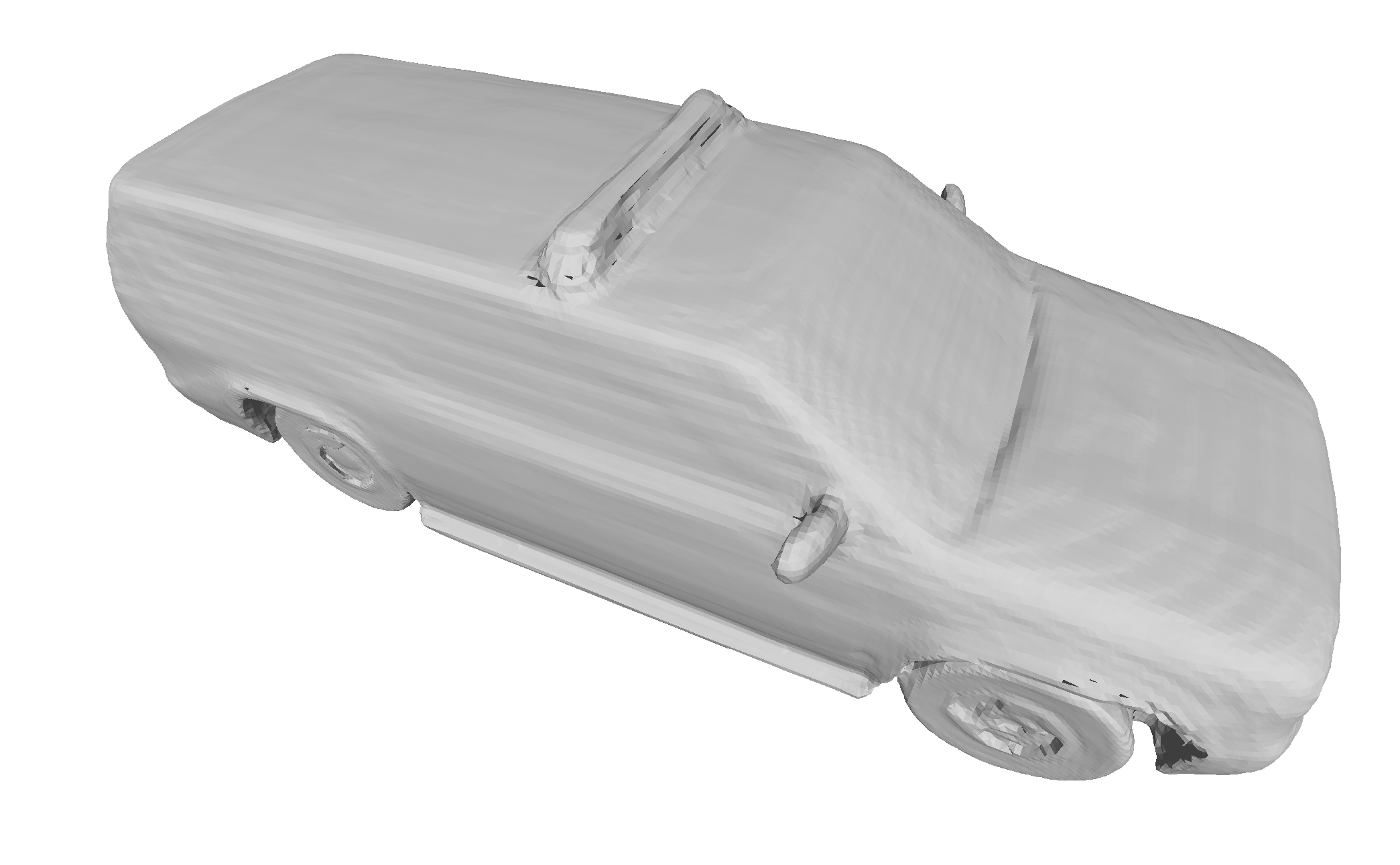}
\includegraphics[width=\subfigsize]{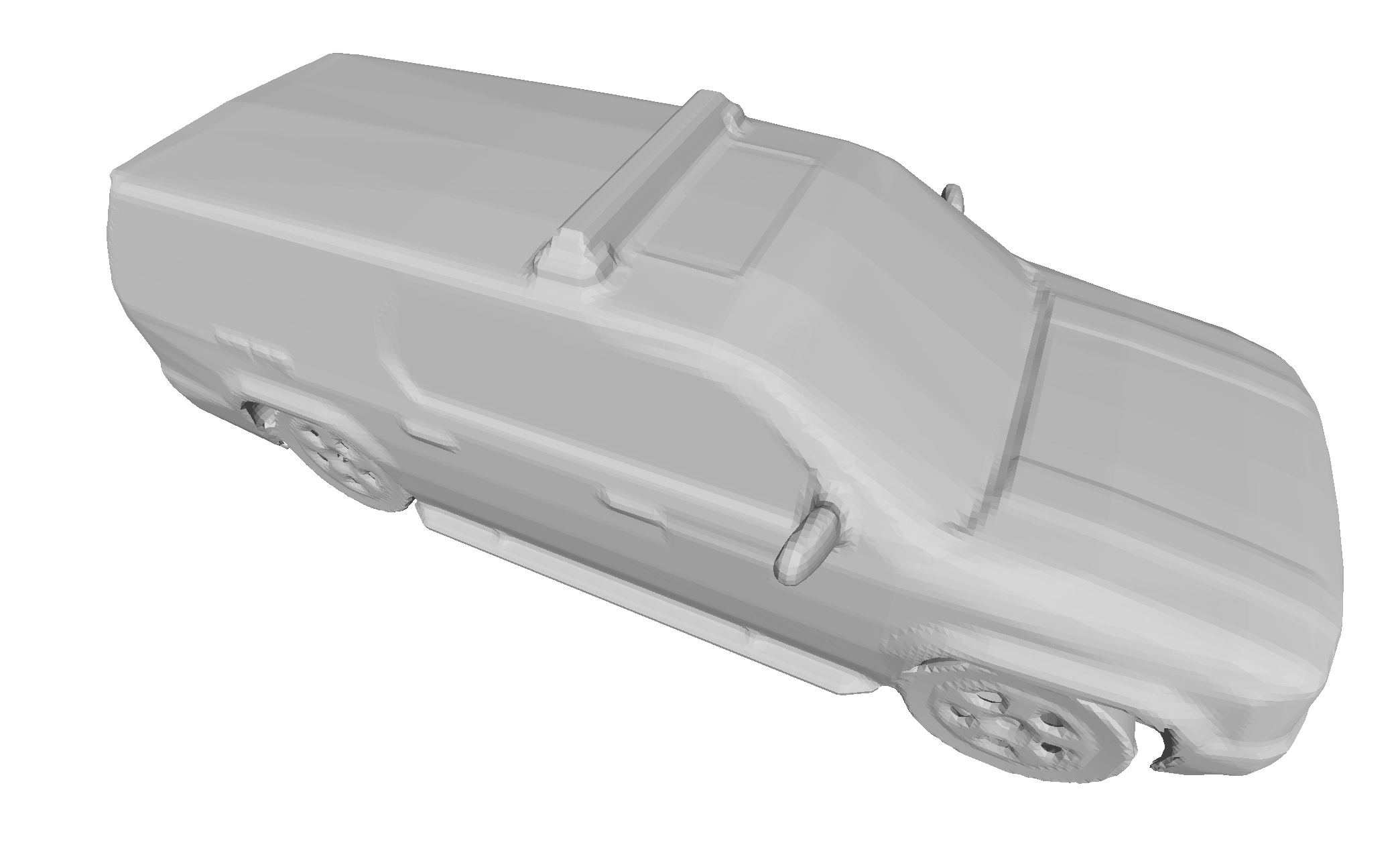}\\
\includegraphics[width=\subfigsize]{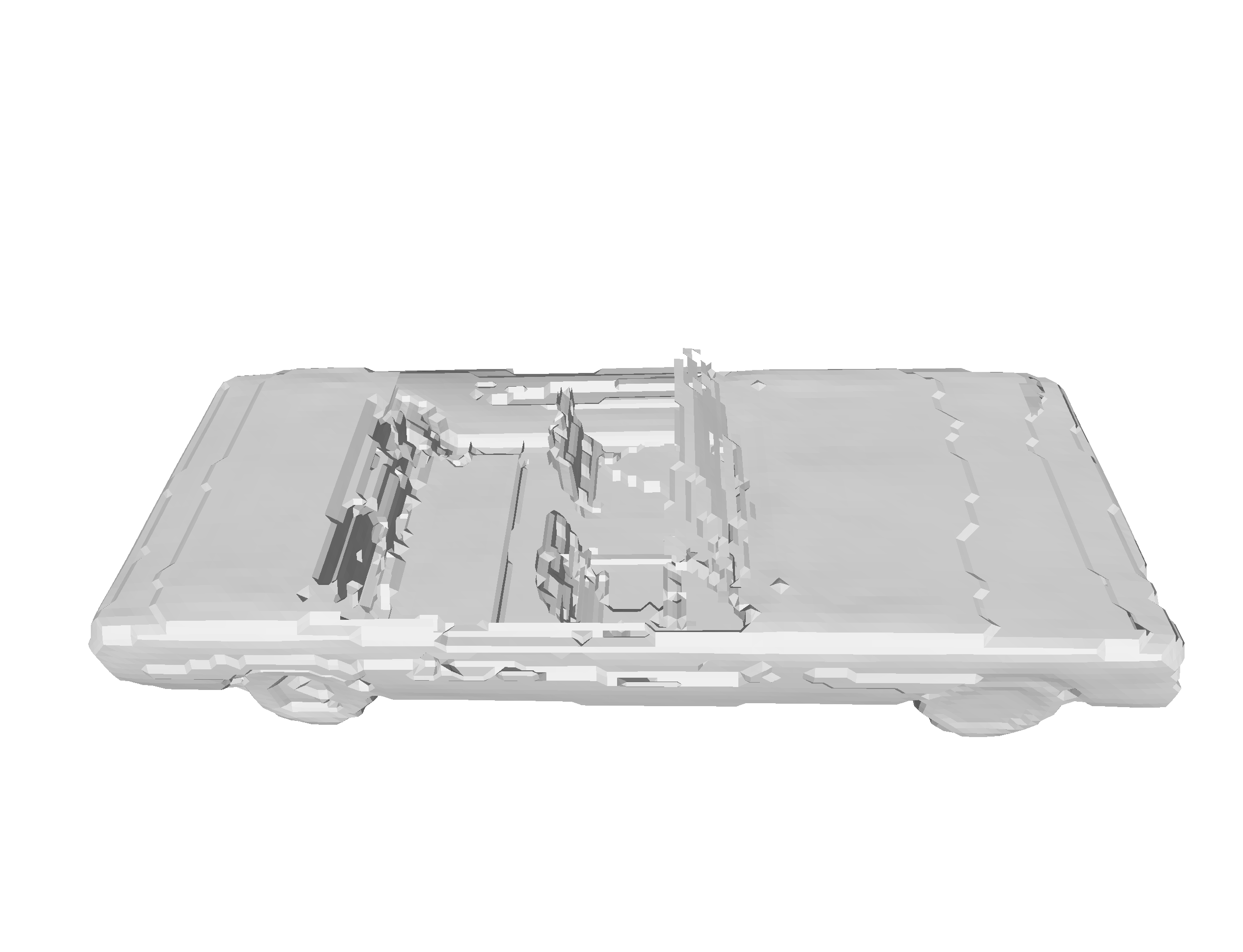}
\includegraphics[width=\subfigsize]{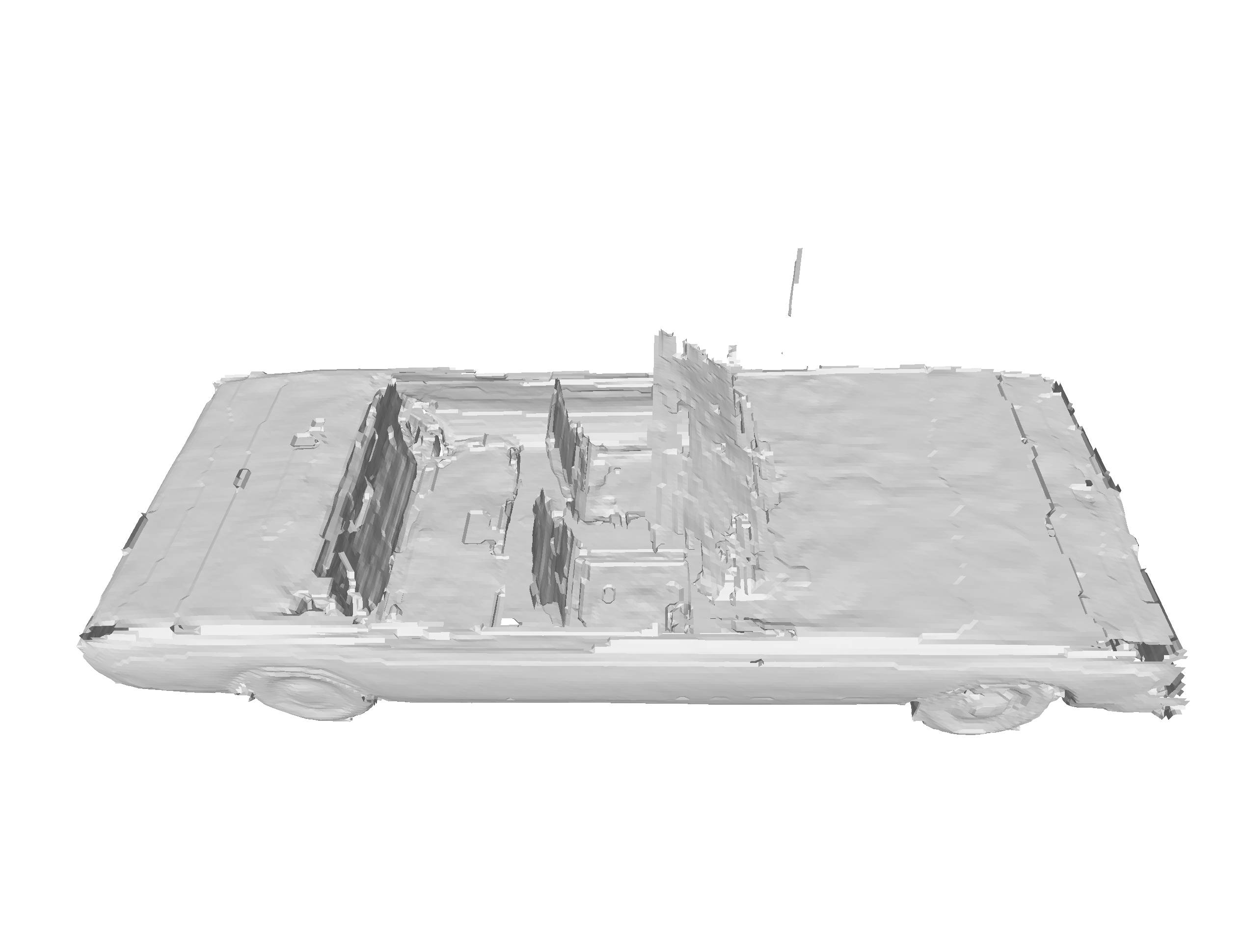}
\includegraphics[width=\subfigsize]{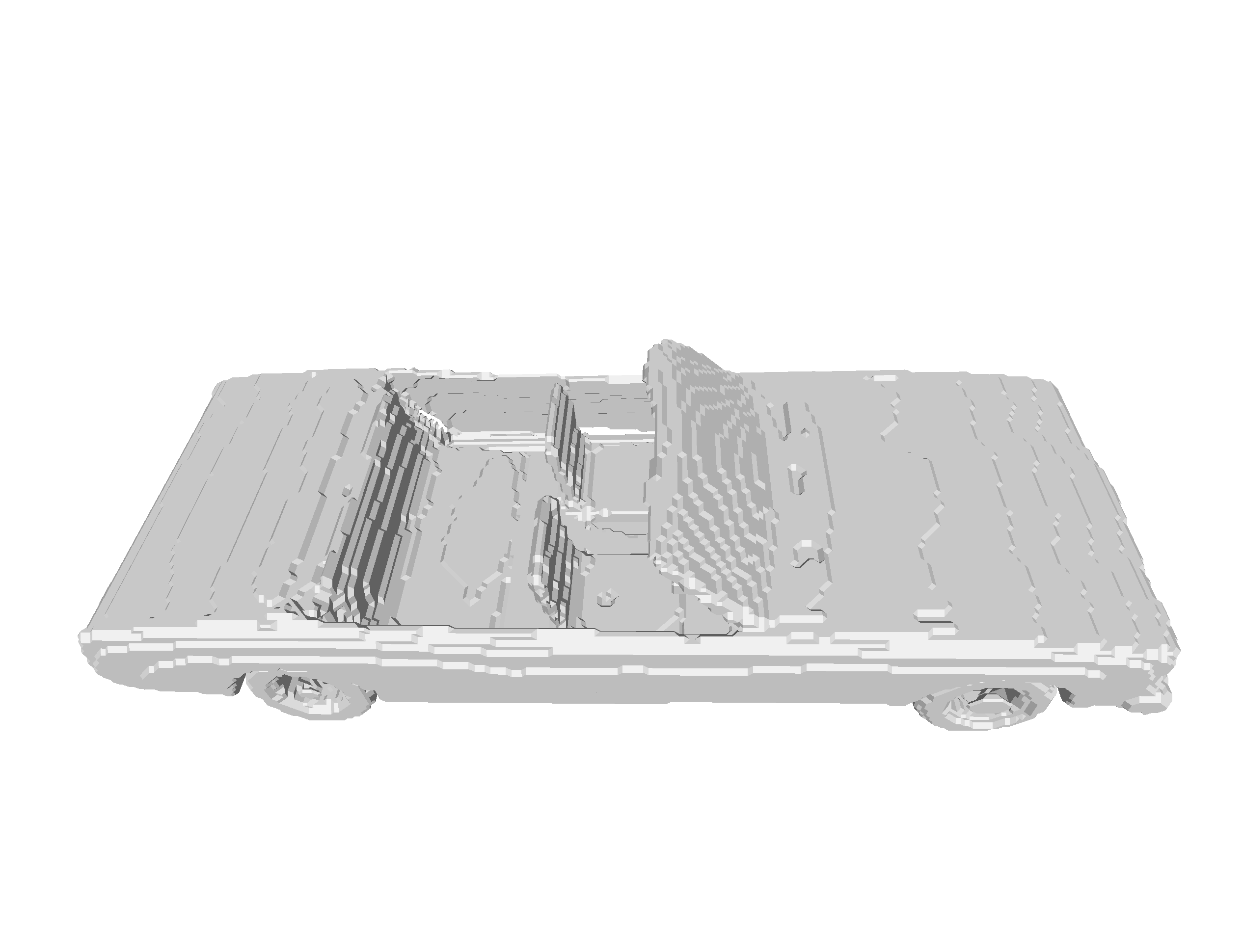}
\includegraphics[width=\subfigsize]{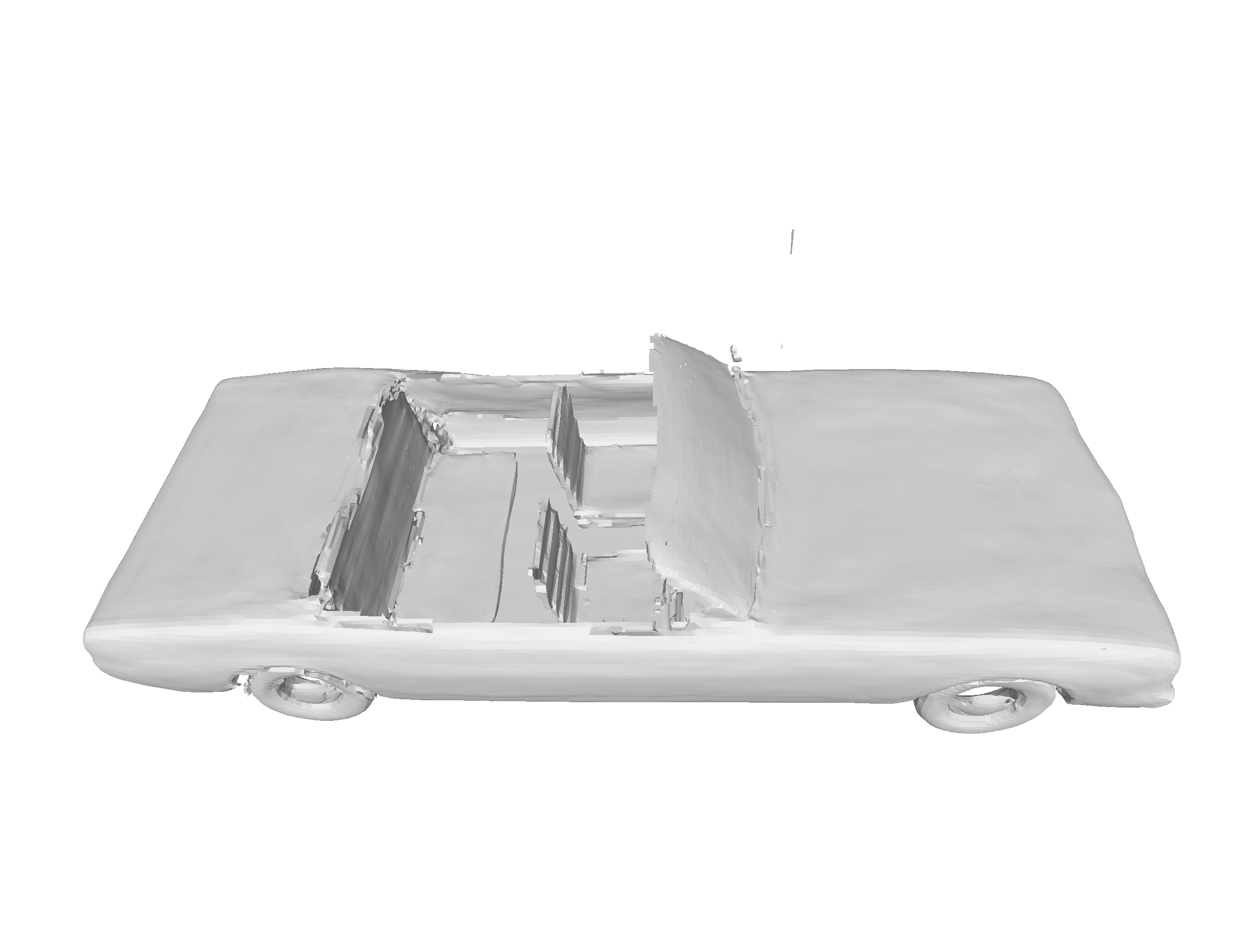}
\includegraphics[width=\subfigsize]{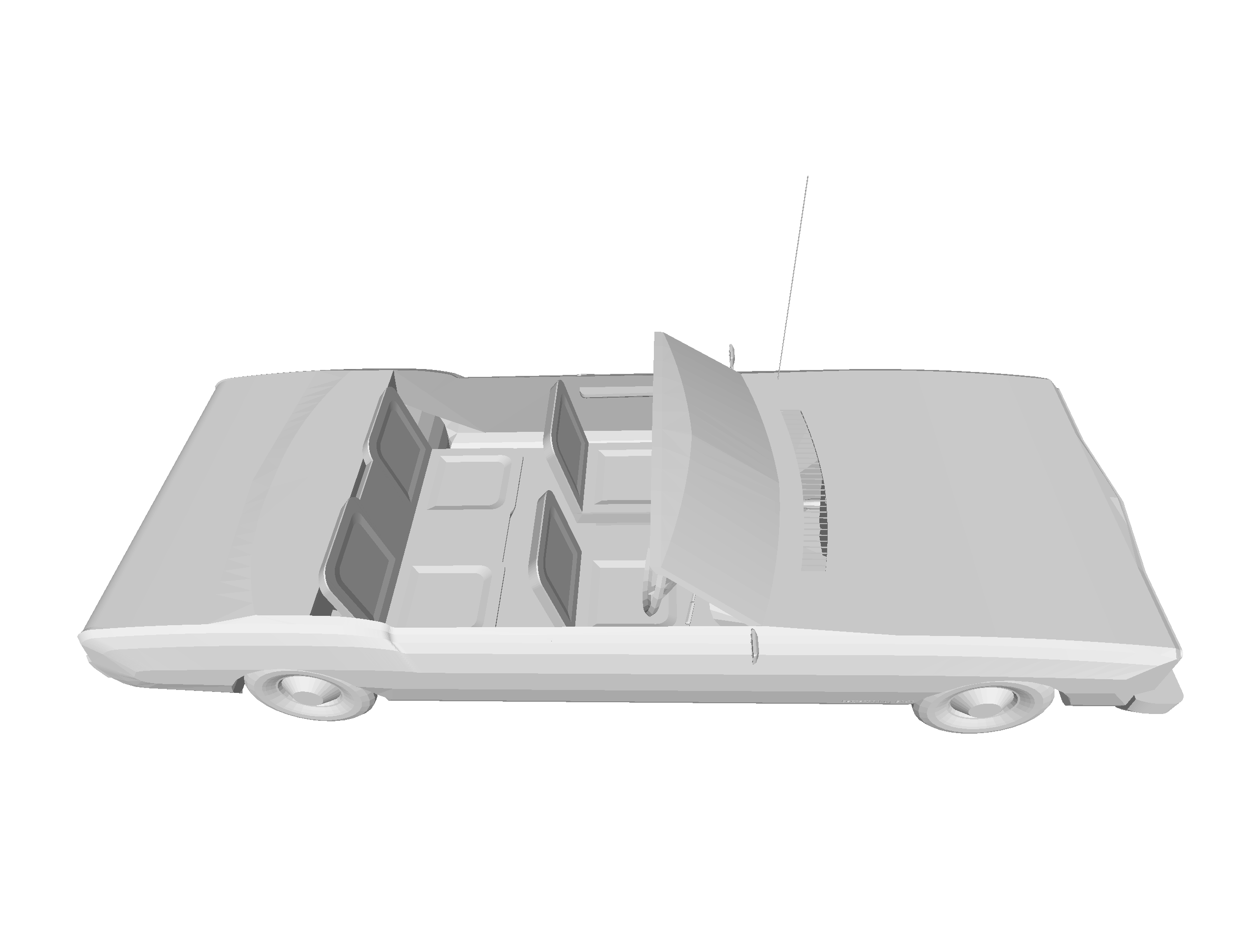}\\
\includegraphics[width=\subfigsize]{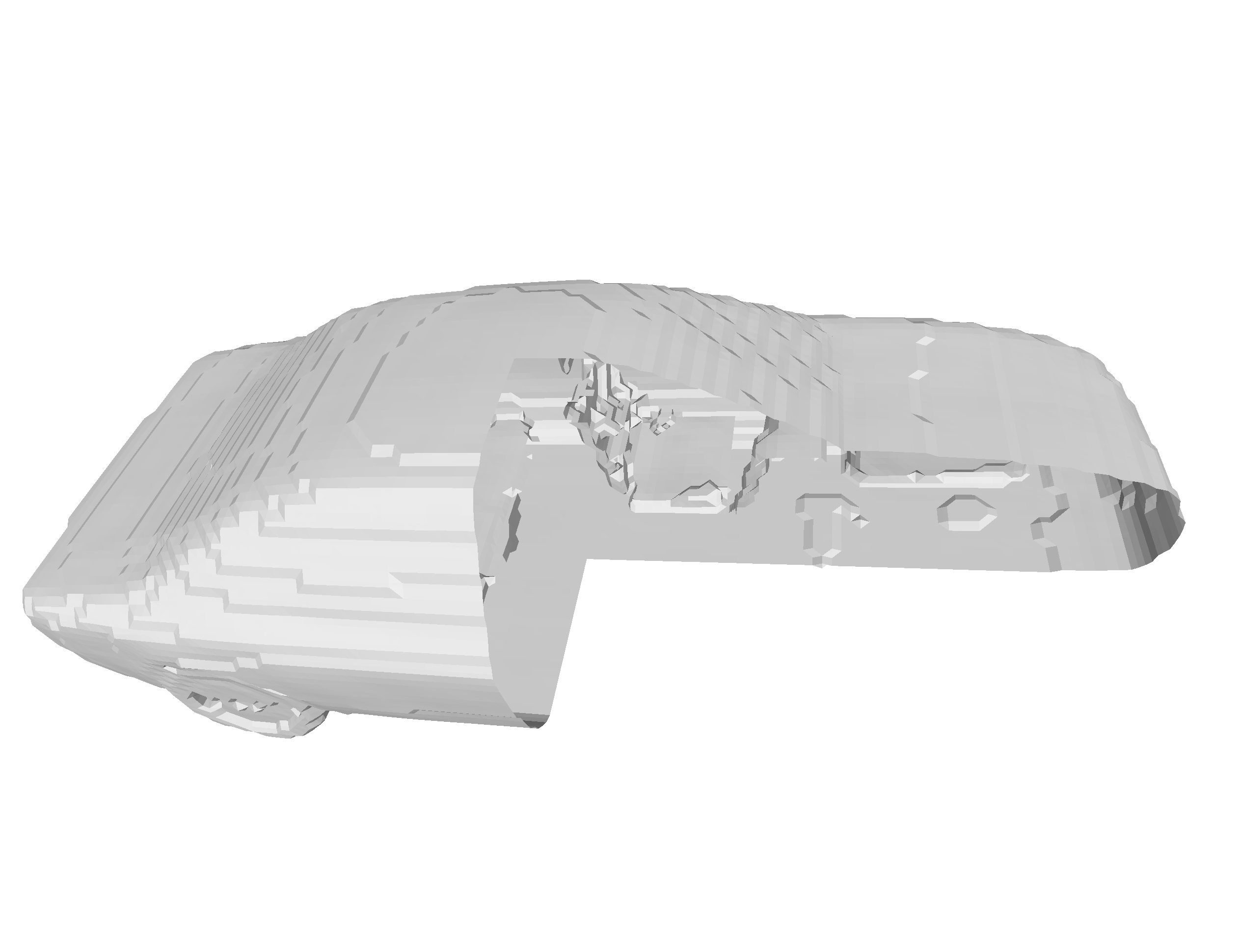}
\includegraphics[width=\subfigsize]{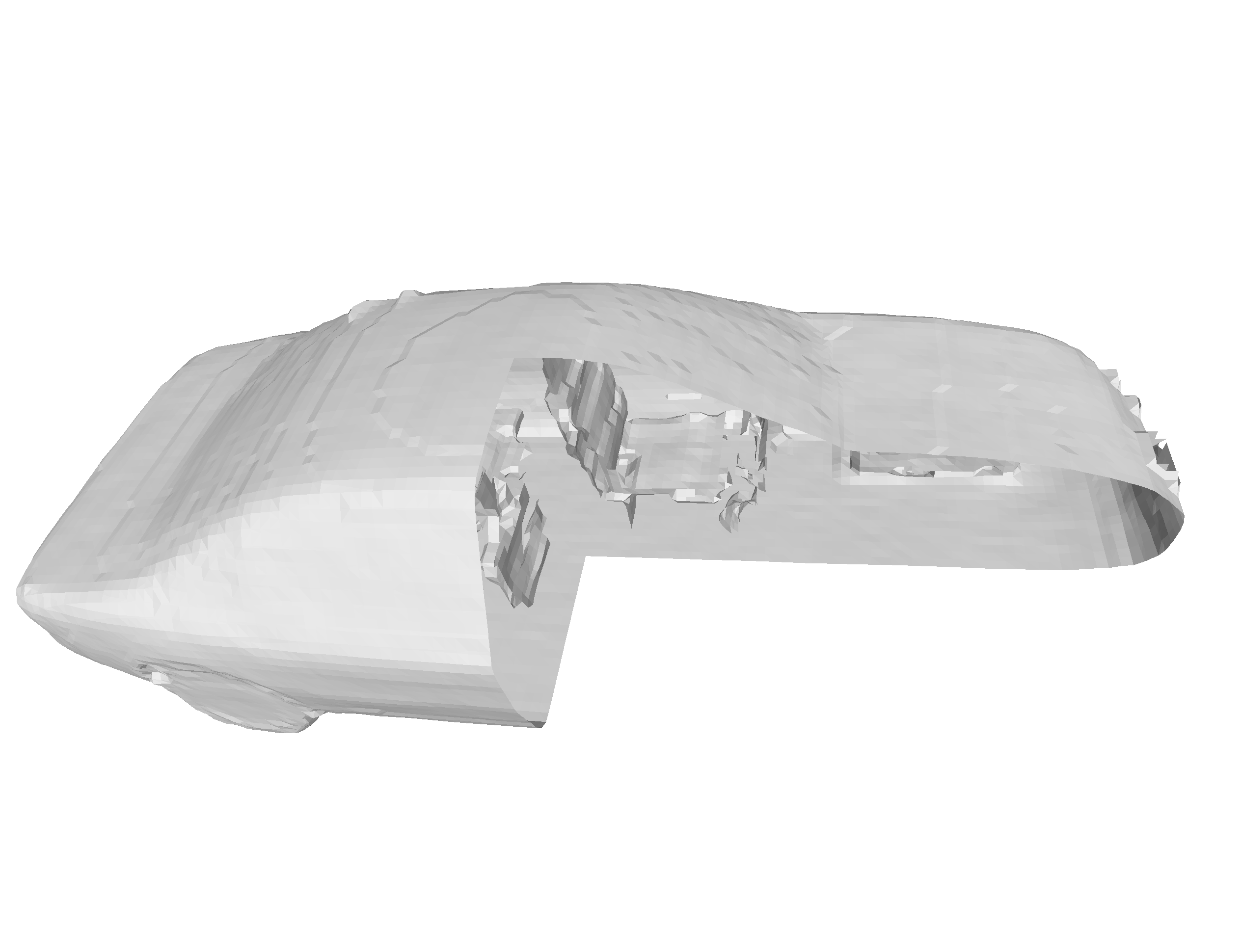}
\includegraphics[width=\subfigsize]{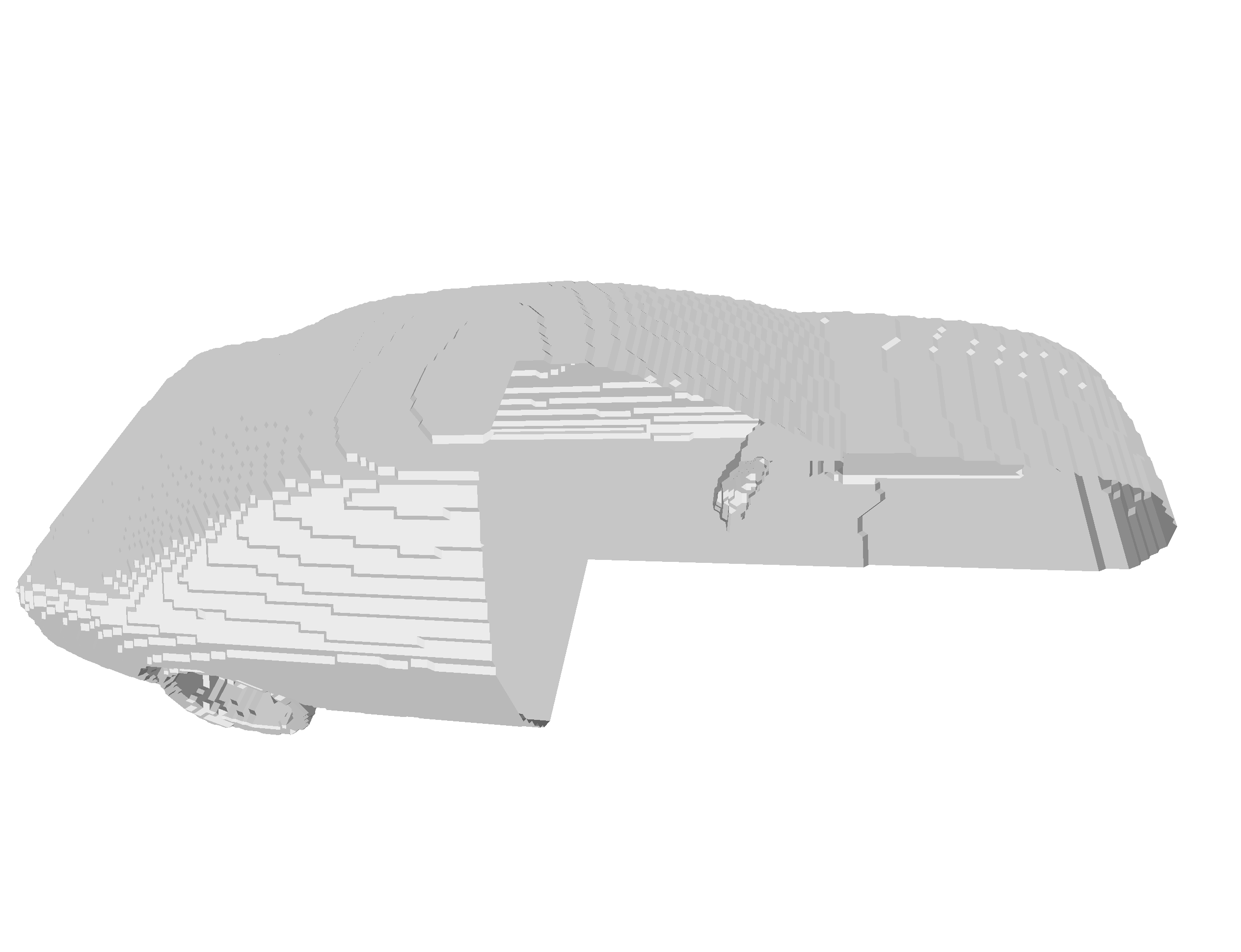}
\includegraphics[width=\subfigsize]{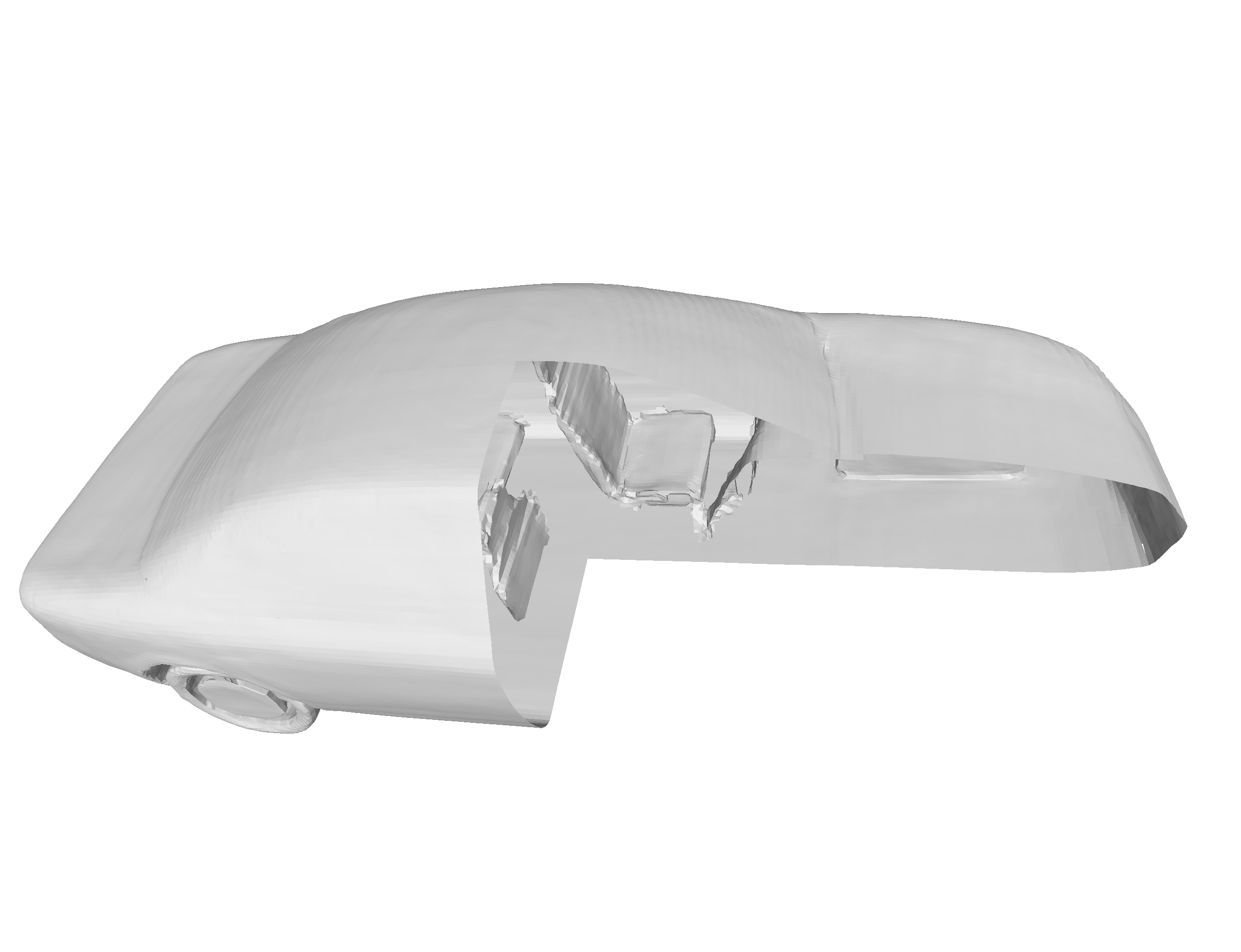}
\includegraphics[width=\subfigsize]{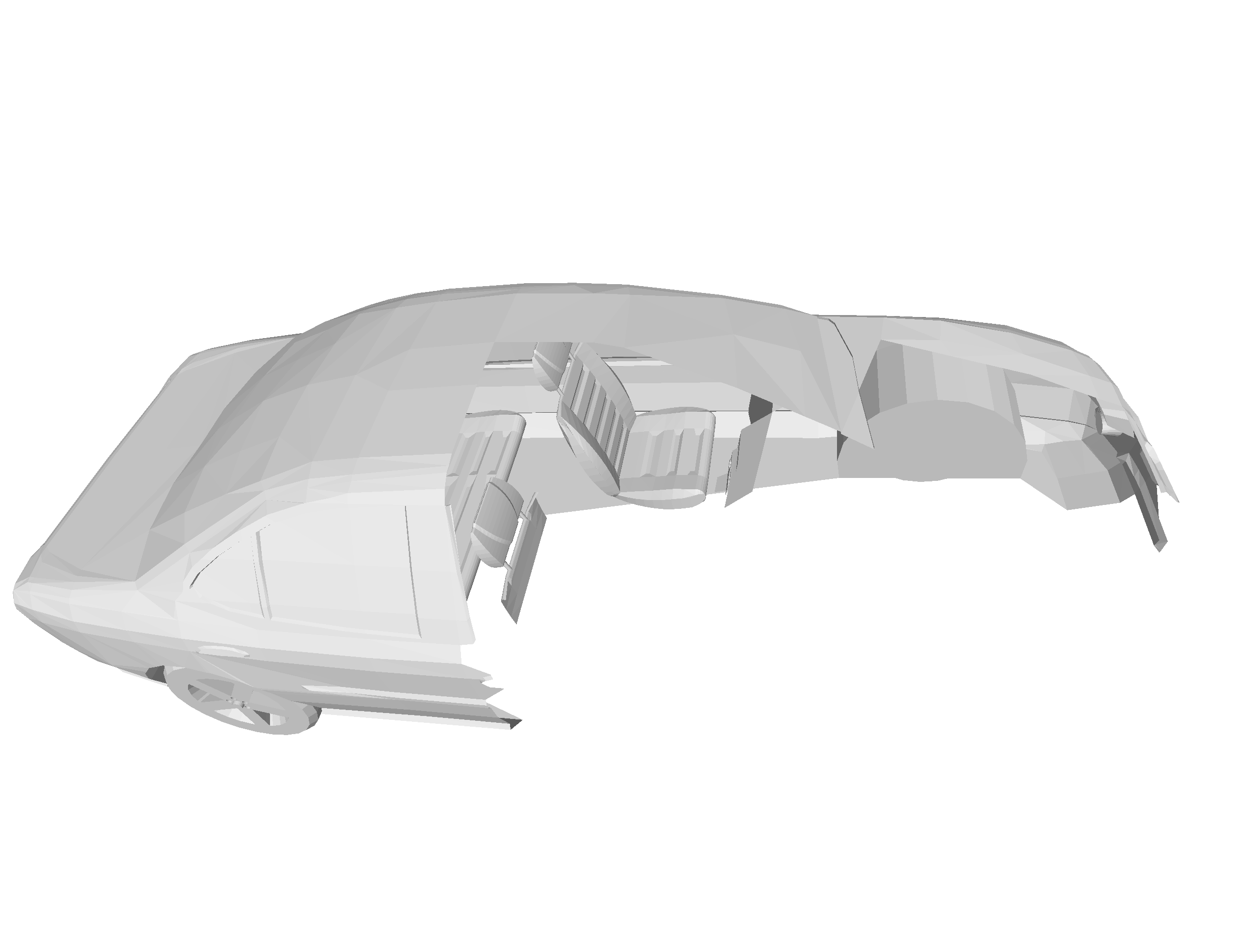}\\

\makebox[\subfigsize]{(a) NDF\cite{Chibane20b} }
\makebox[\subfigsize]{(b) CSP \cite{Venkatesh21} }
\makebox[\subfigsize]{(c) 3PSDF \cite{Chen22e} }
\makebox[\subfigsize]{(d) GDF (Ours) }
\makebox[\subfigsize]{(e) GT }
\end{center} 
\vspace{-5mm}
\caption{
\textbf{Representing known shapes using an auto-decoder.} We use the same MLPs trained using different implicit representations. {\bf Top rows.}  Car from ShapeNet-Car(P).  {\bf Bottom rows.}   Car from the ShapeNet-Car(R), which contains the inside of the car. We remove a quarter of the reconstructed models in the last row to make it visible. 
} 
\label{fig:auto-decoder} 
\end{figure*}

\parag{Generalizing to New Shapes.}
Given a trained auto-decoder, we evaluate its ability to model previously unseen test instances by optimizing a randomly initialized latent code to fit partial observations in the form of a sparse cloud of surface points. 
To this end, given an initial latent code, we sample a set of query points and find the nearest points among the point cloud to compute the gradient distance vectors. These gradient distance vectors serve as pseudo ground-truth for optimizing the latent code. Similarly, we compute the pseudo unsigned distances for optimizing NDF and pseudo closest surface points for CSP. We do not include 3PSDF in this experiment because it requires knowledge of the surface gradients to assign pseudo labels to each query point, which is not trivial to obtain from the point cloud. Note that in the original paper, 3PSDF relies on a separate 3D point encoder for this task. However, the code for this module is not publicly available. We report our comparative results in Table \ref{tab:point-cloud}. Again, our approach delivers significantly better accuracy.  We provide qualitative results in Fig.~\ref{fig:pointcloud}. Unlike those of the other two methods, our results are free of holes in irregular parts of the surfaces.

\begin{table}[t]
    \caption{\small \textbf{Representing previously unseen shapes using an auto-decoder.} Average L2 Chamfer Distance $\times 10^{-4}$ (CD) and normal consistency (NC) for the MGN and closed ShapeNet-Car datasets. }
    \vspace{-5mm}
    \label{tab:point-cloud}
    \begin{small}
        \begin{center}
            \setlength{\tabcolsep}{5pt}
            \begin{tabular}{ccc|cc} 
                \multicolumn{1}{l}{} & \multicolumn{2}{c}{\textbf{SN-Car (P)}} & \multicolumn{2}{c}{\textbf{SN-Car (R)}} \\ \midrule
                \multicolumn{1}{c|}{Input Points} & 3K & 10K & 3K & 10K \\ \midrule
                \multicolumn{1}{c|}{NDF\cite{Chibane20b}}          & 7.54  & 1.53 & 1.65 & 0.78 \\
				\multicolumn{1}{c|}{CSP\cite{Venkatesh21}}          & 5.91  & 1.99 & 1.12 & 0.87 \\
				\midrule
                \multicolumn{1}{c|}{GDF (Ours)}    & \textbf{1.32} & \textbf{0.86} & \textbf{0.98} &\textbf{0.53} \\ \midrule
            \end{tabular}
        \end{center}
    \end{small}
\end{table}


\def\subfigsize{0.192\linewidth}
\begin{figure*}[ht]
\begin{center}

\includegraphics[width=\subfigsize]{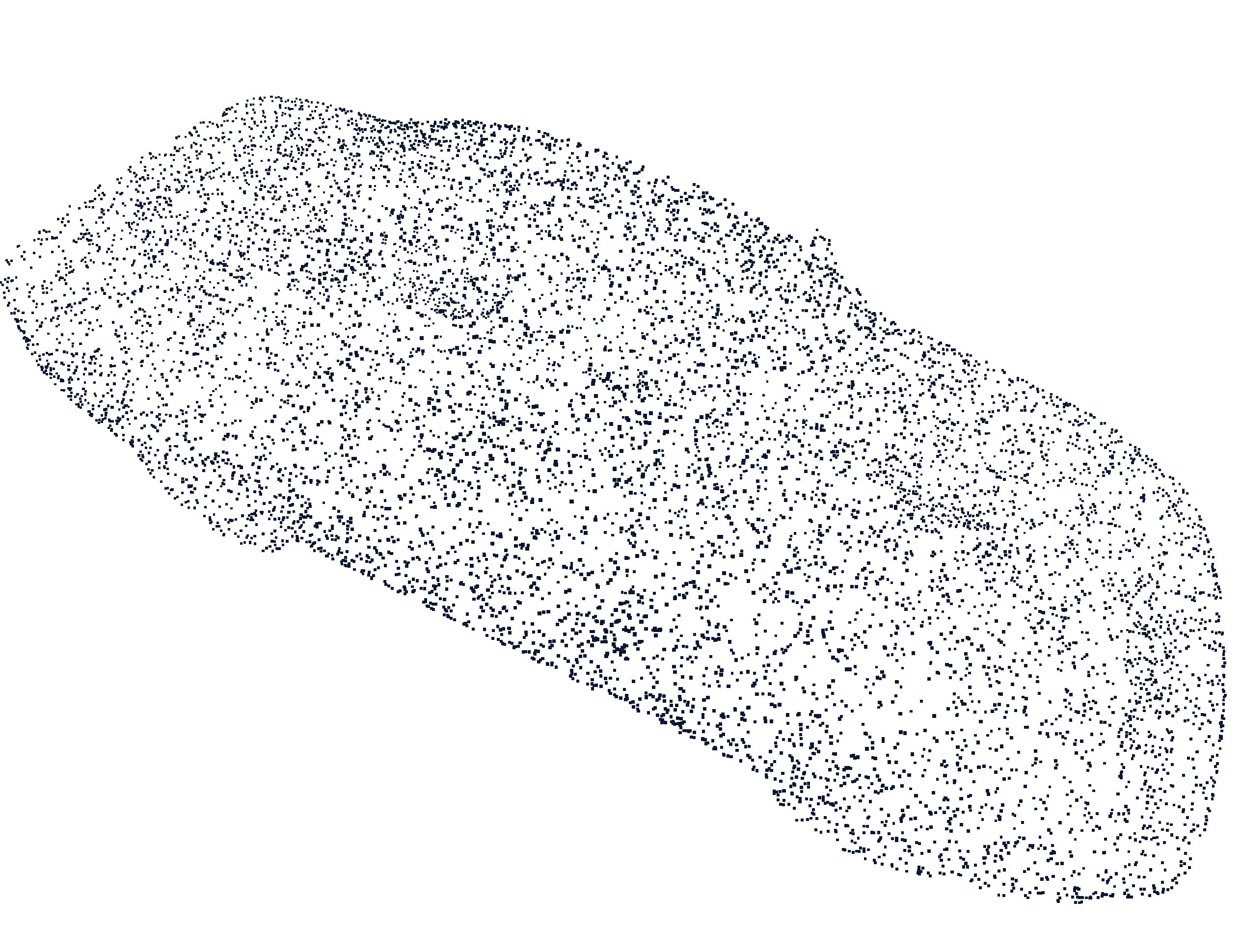}
\includegraphics[width=\subfigsize]{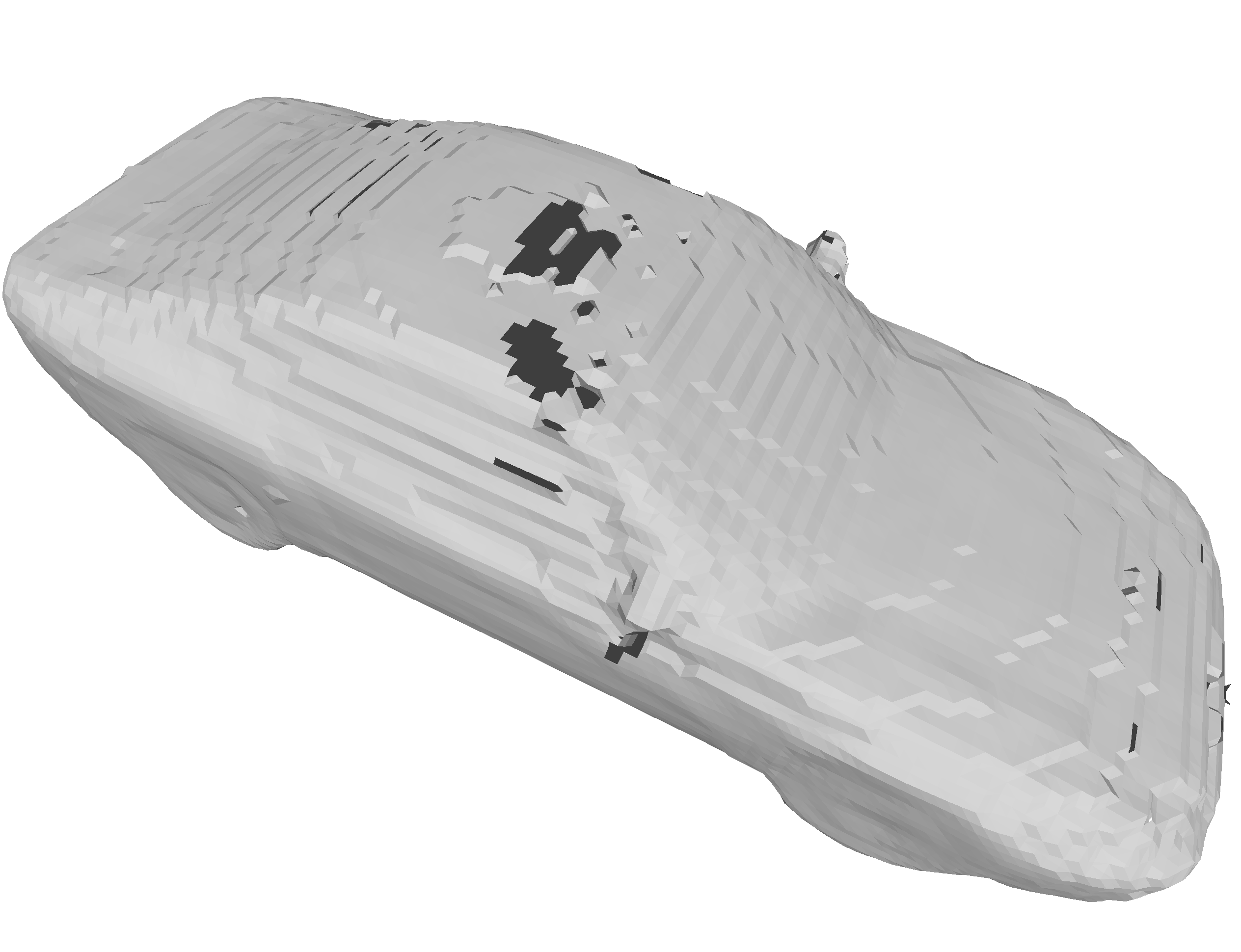}
\includegraphics[width=\subfigsize]{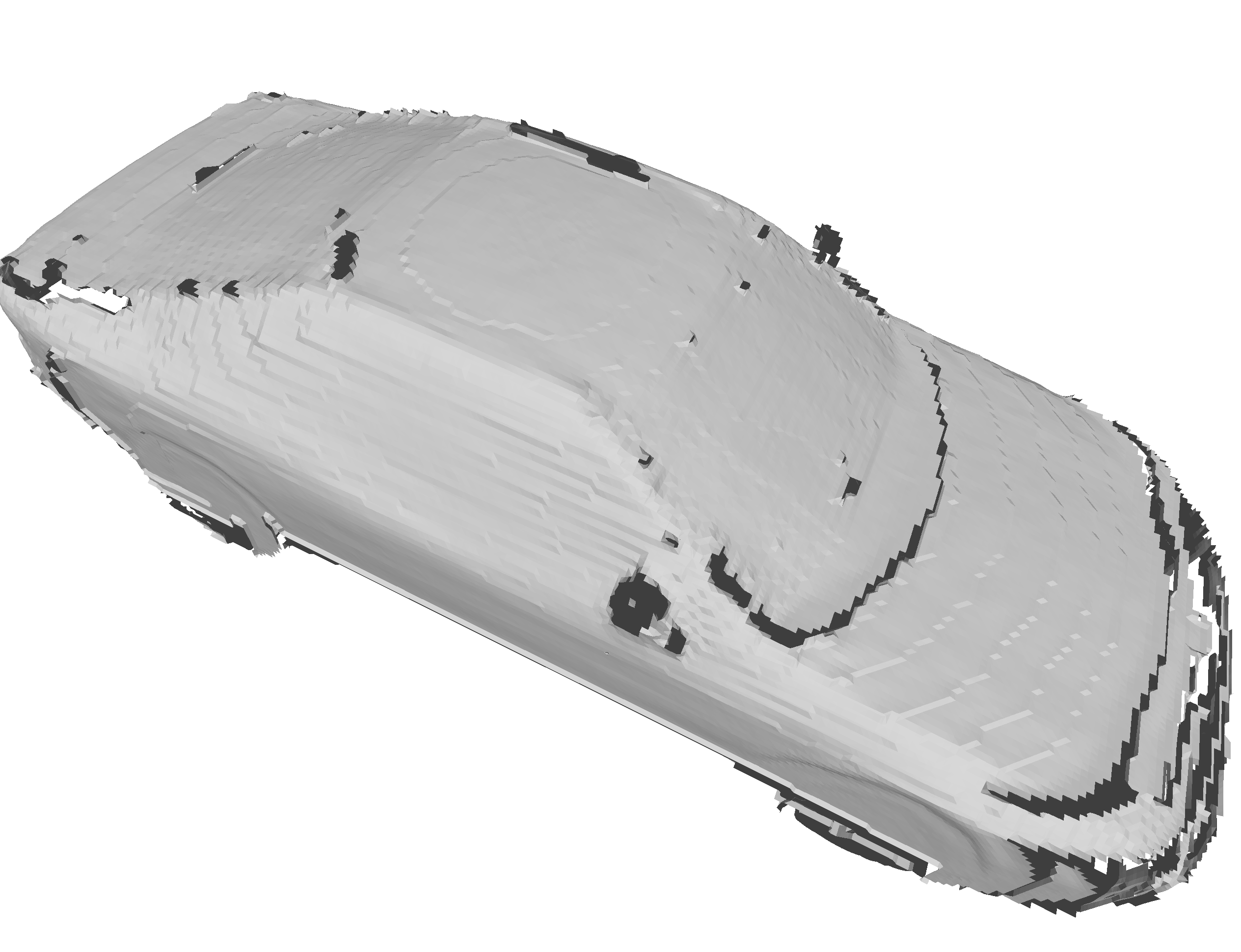}
\includegraphics[width=\subfigsize]{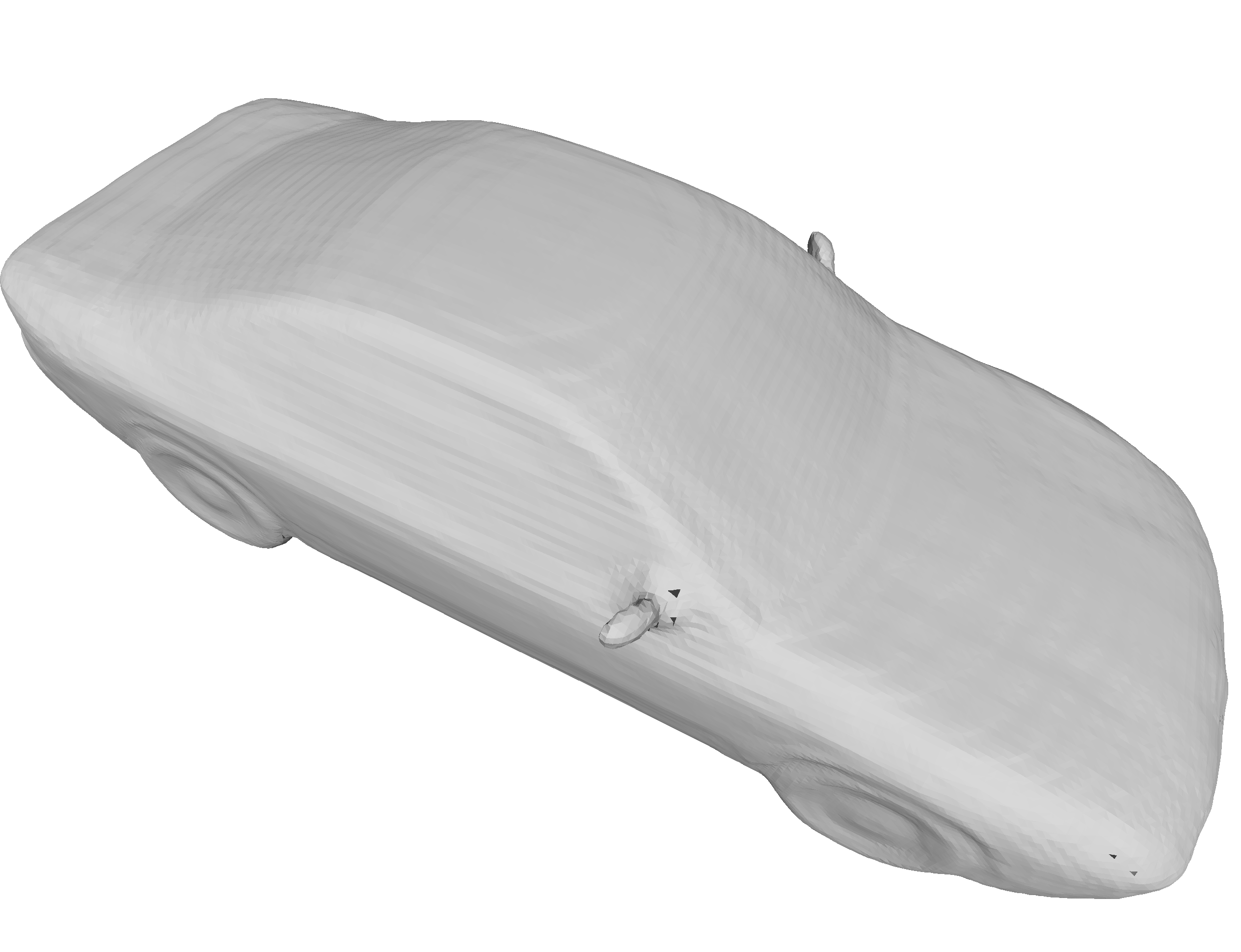}
\includegraphics[width=\subfigsize]{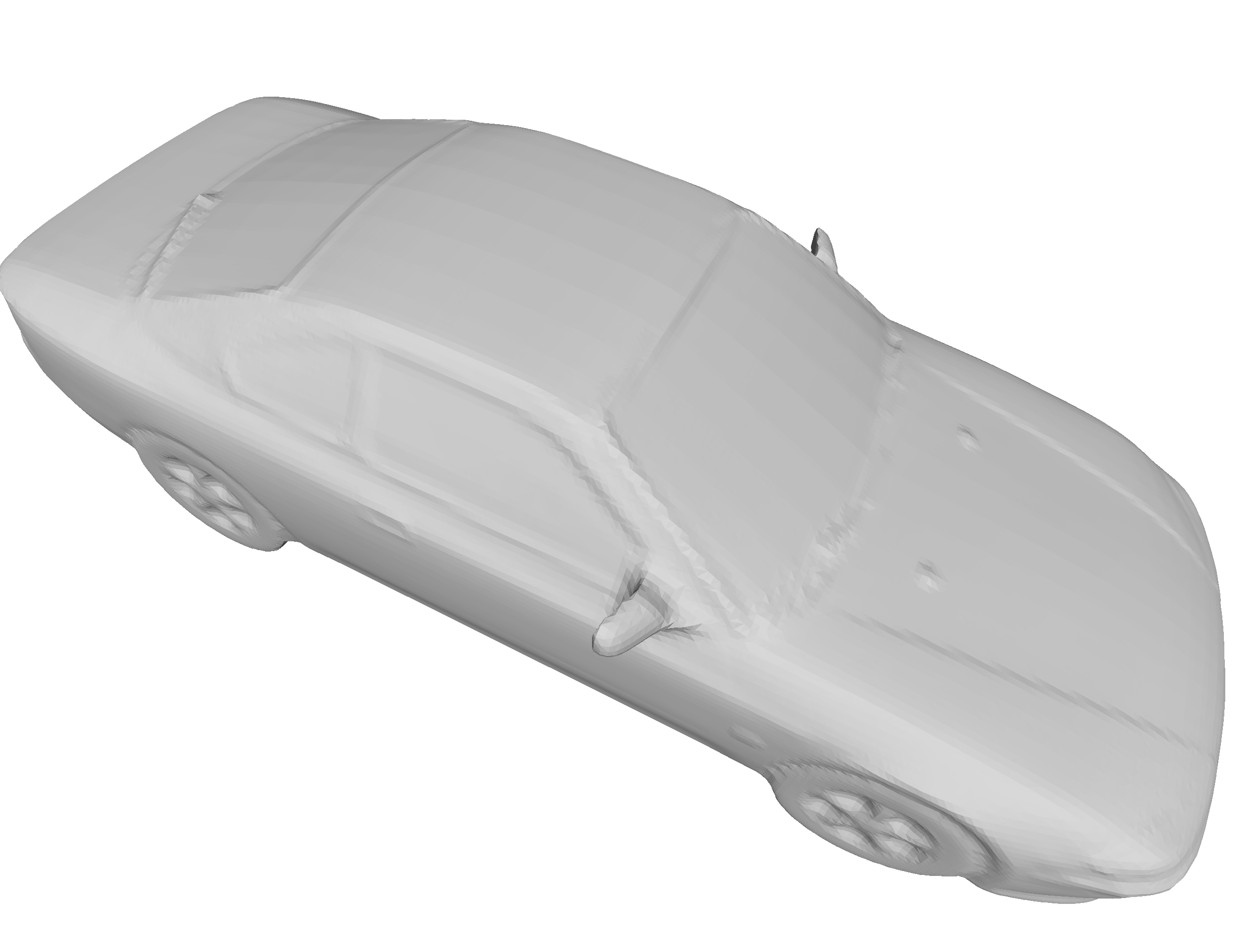}\\
\includegraphics[width=\subfigsize]{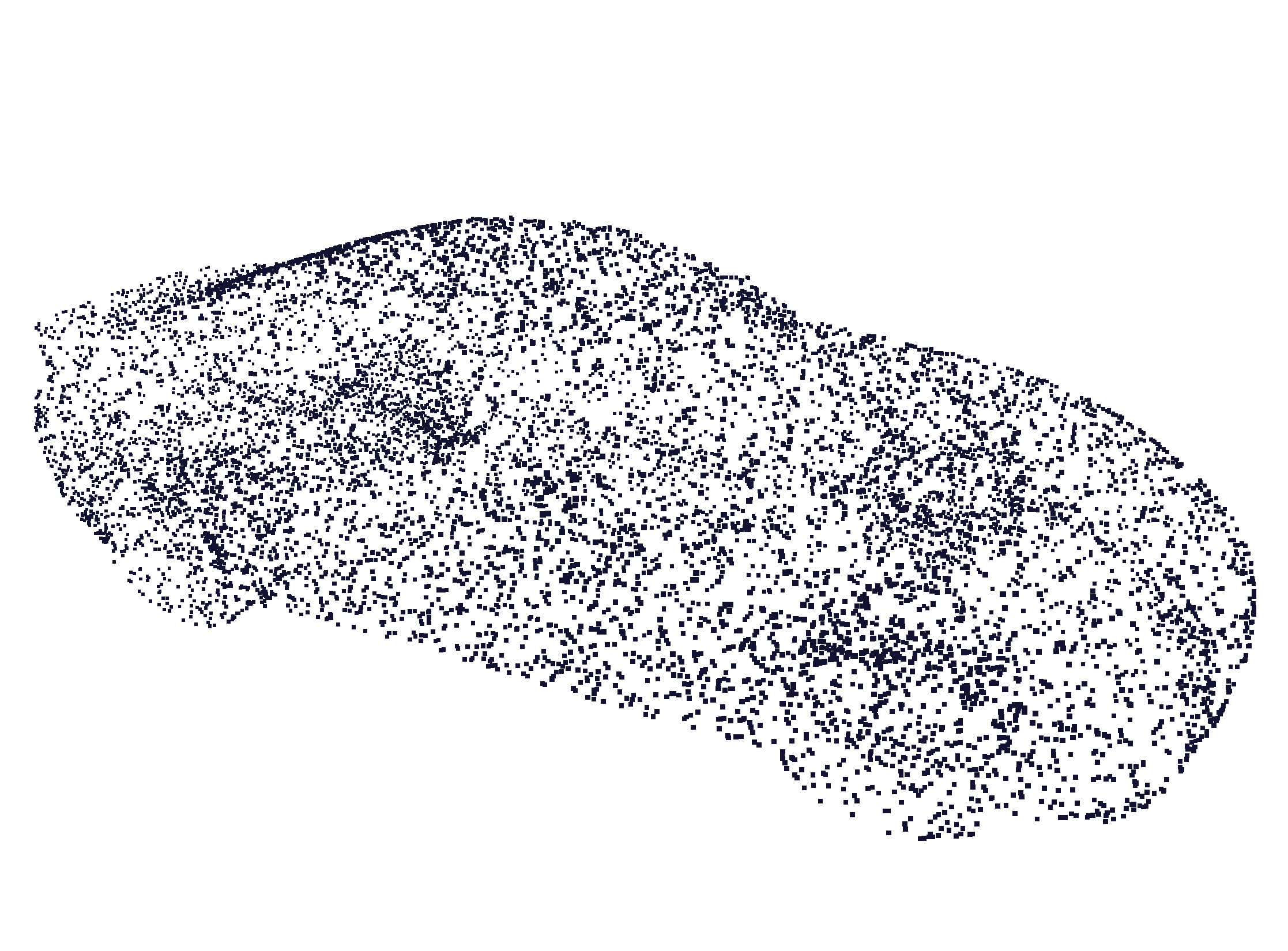}
\includegraphics[width=\subfigsize]{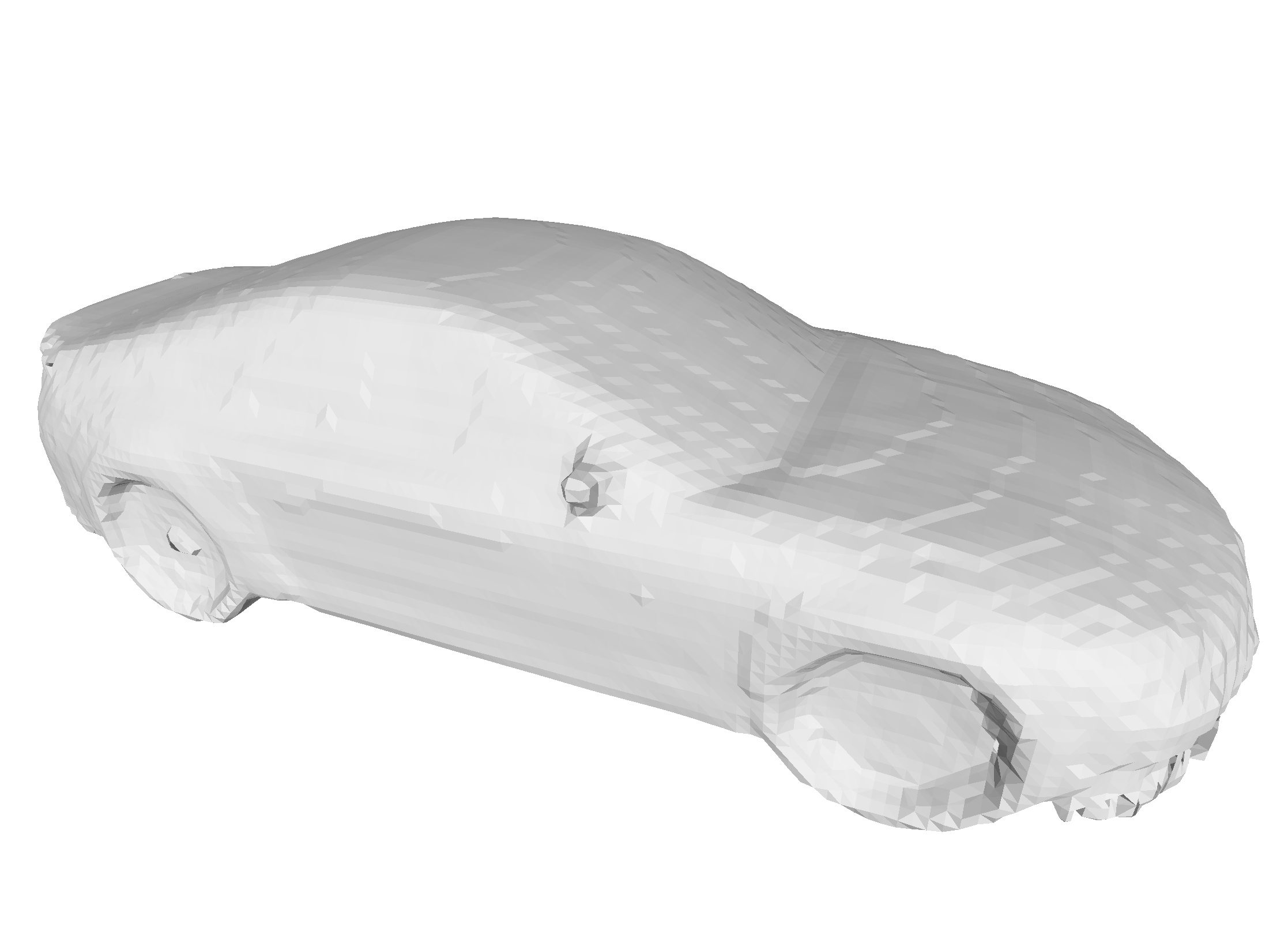}
\includegraphics[width=\subfigsize]{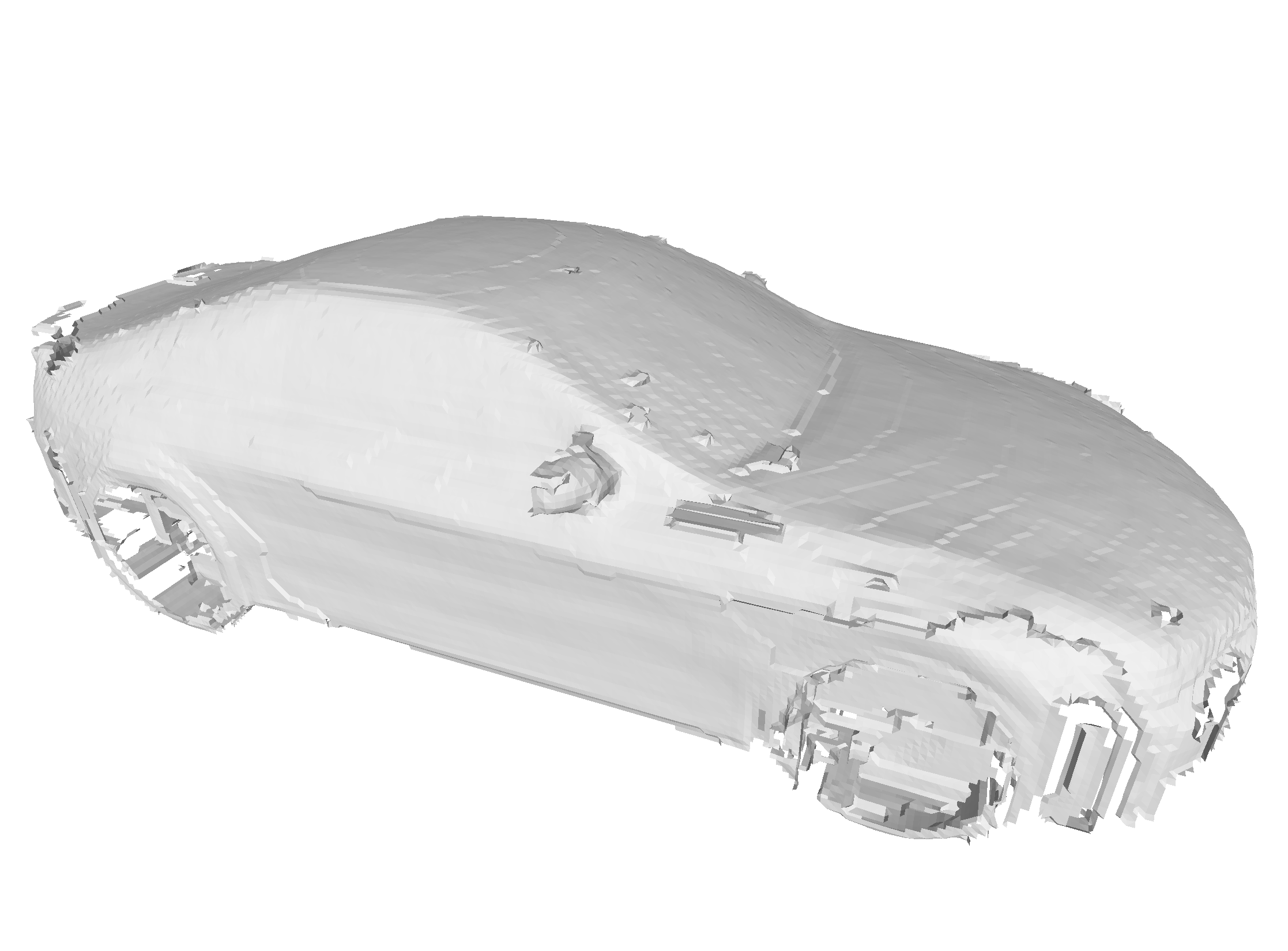}
\includegraphics[width=\subfigsize]{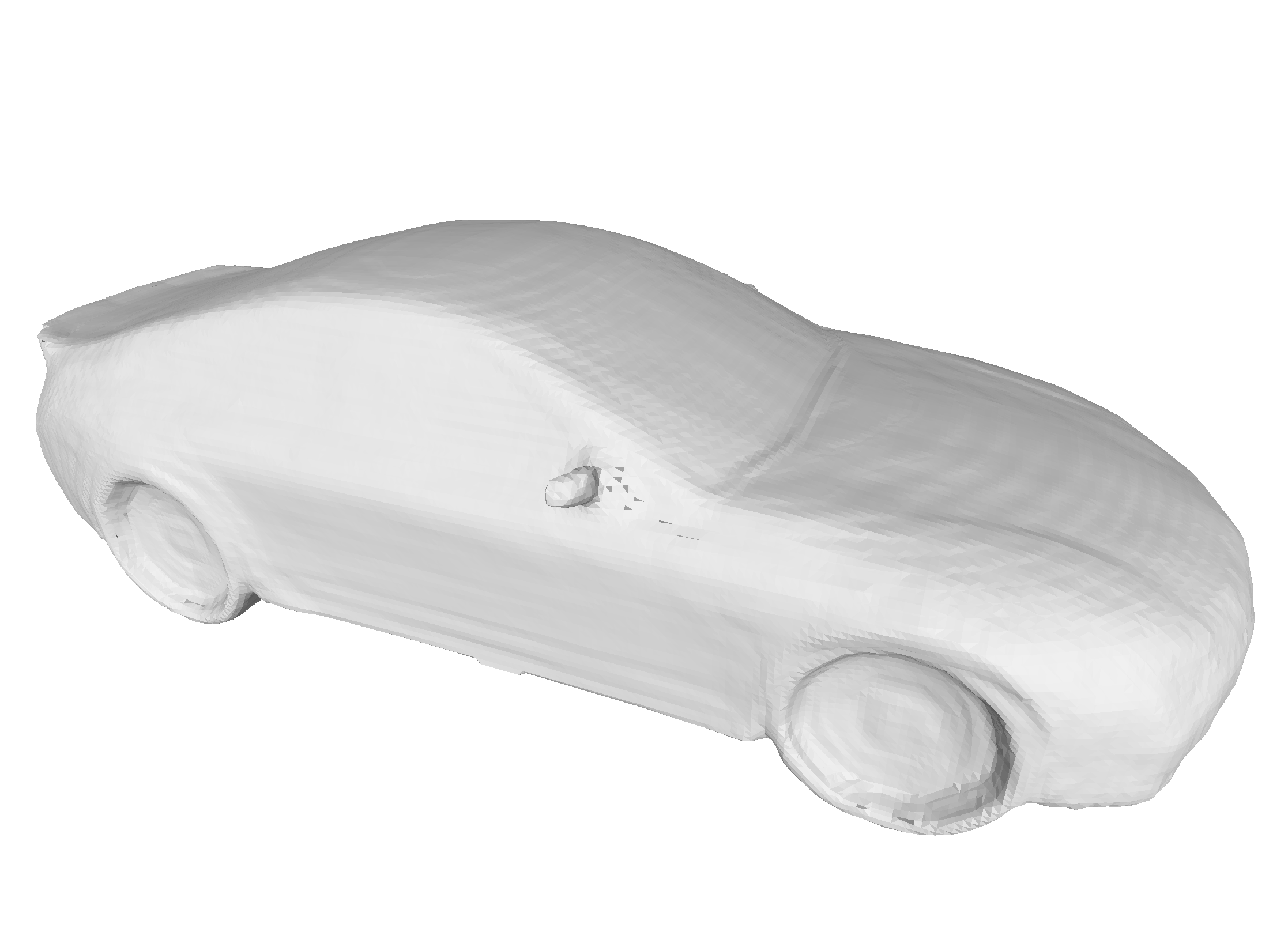}
\includegraphics[width=\subfigsize]{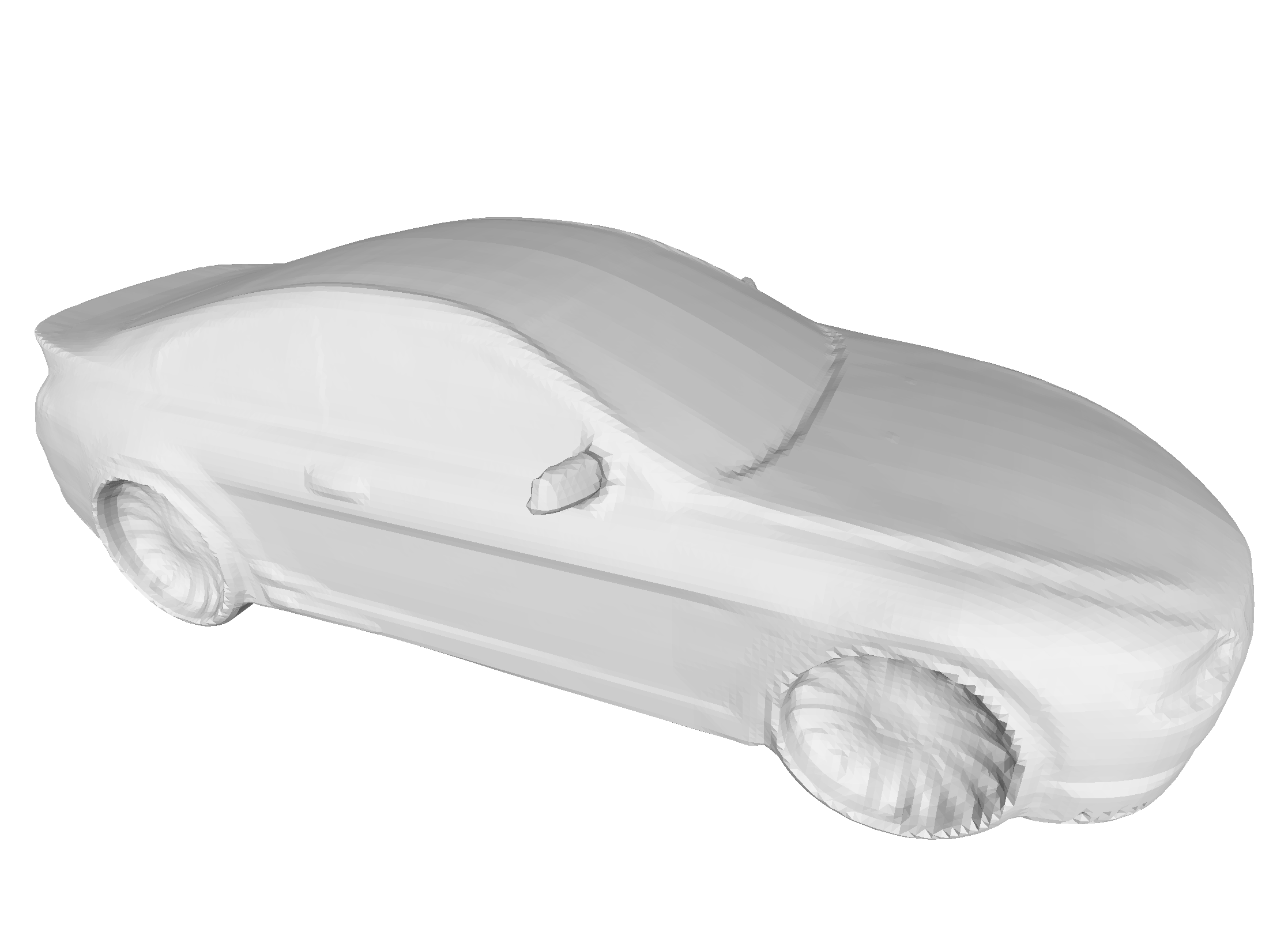}\\
\includegraphics[width=\subfigsize]{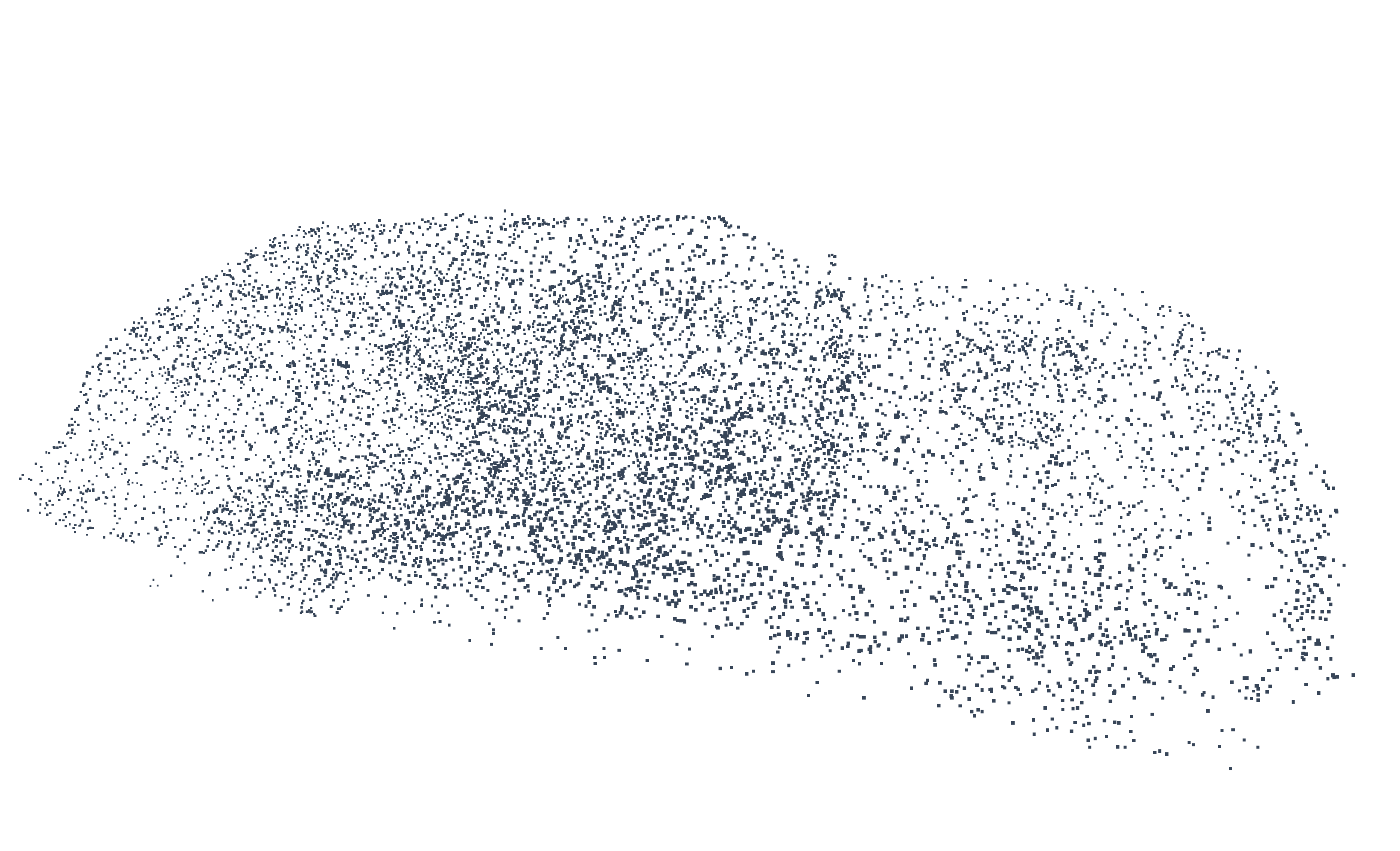}
\includegraphics[width=\subfigsize]{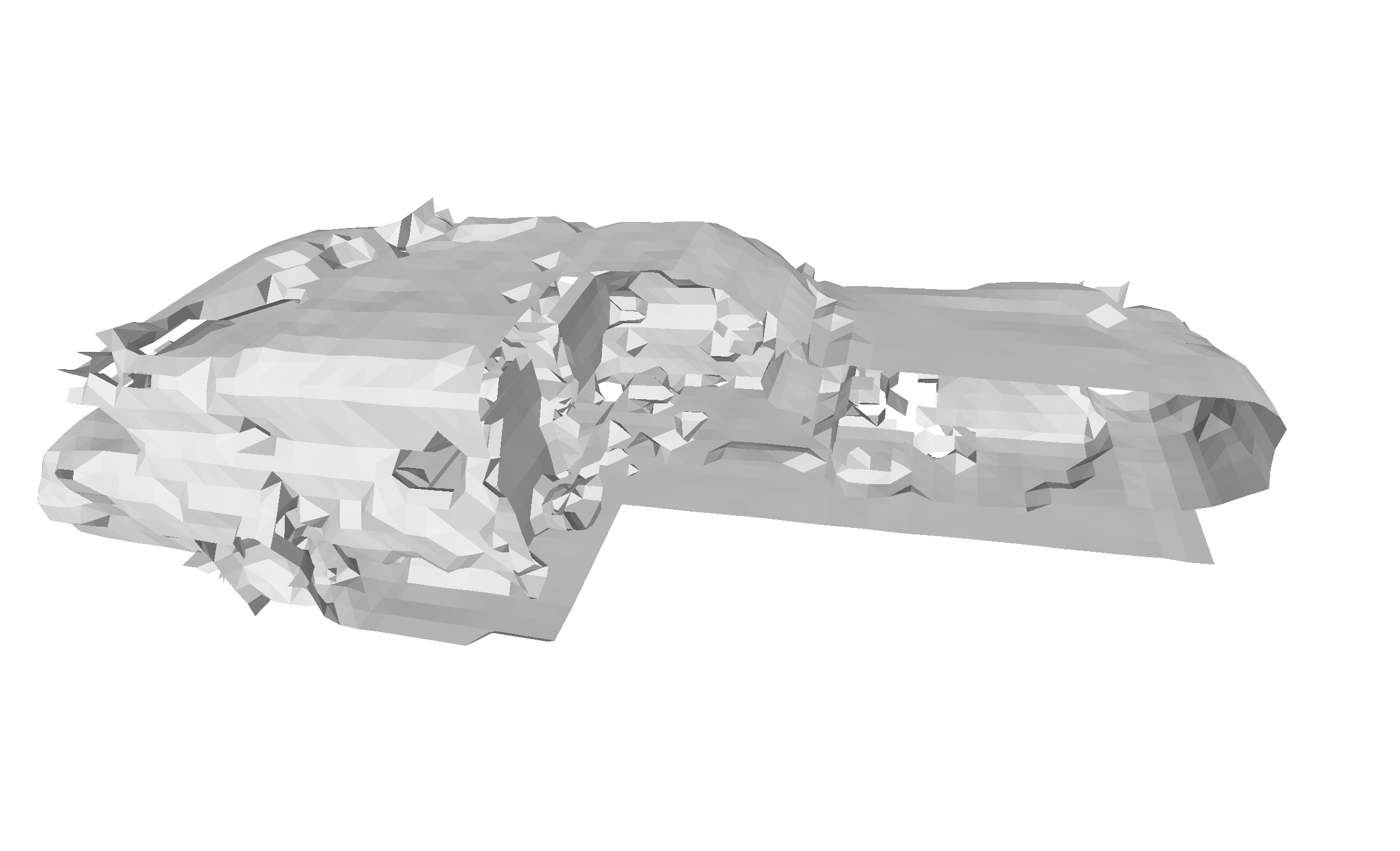}
\includegraphics[width=\subfigsize]{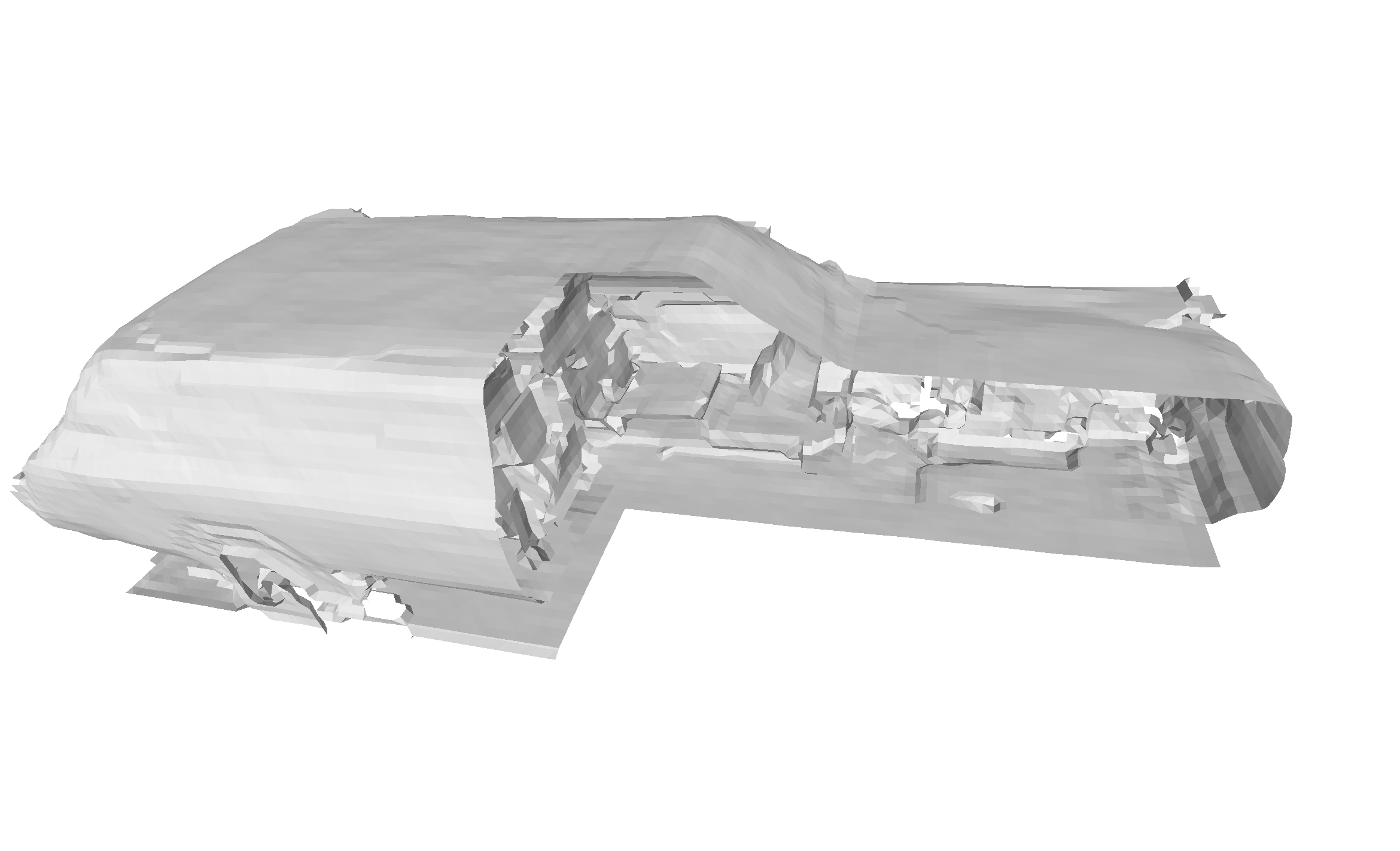}
\includegraphics[width=\subfigsize]{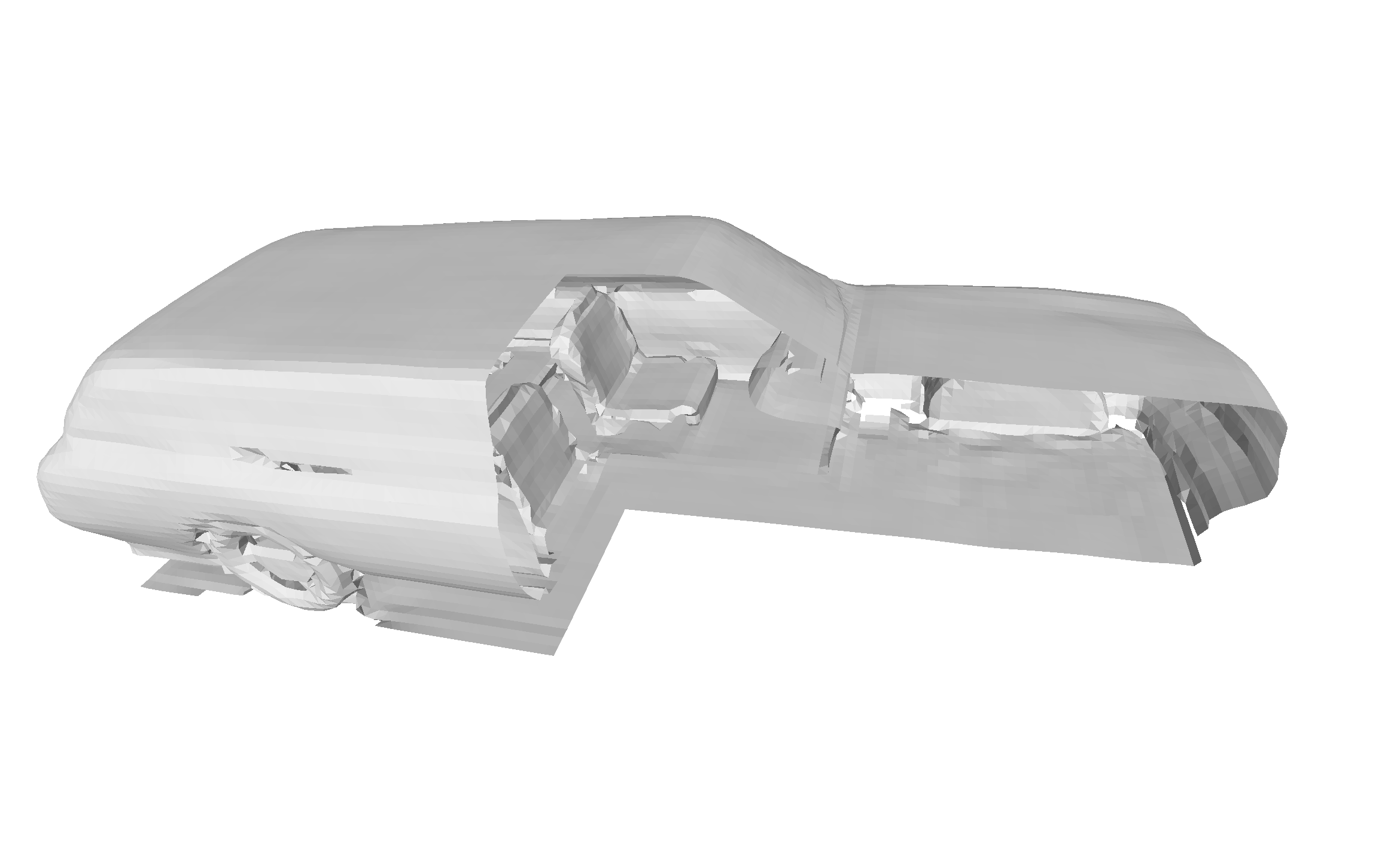}
\includegraphics[width=\subfigsize]{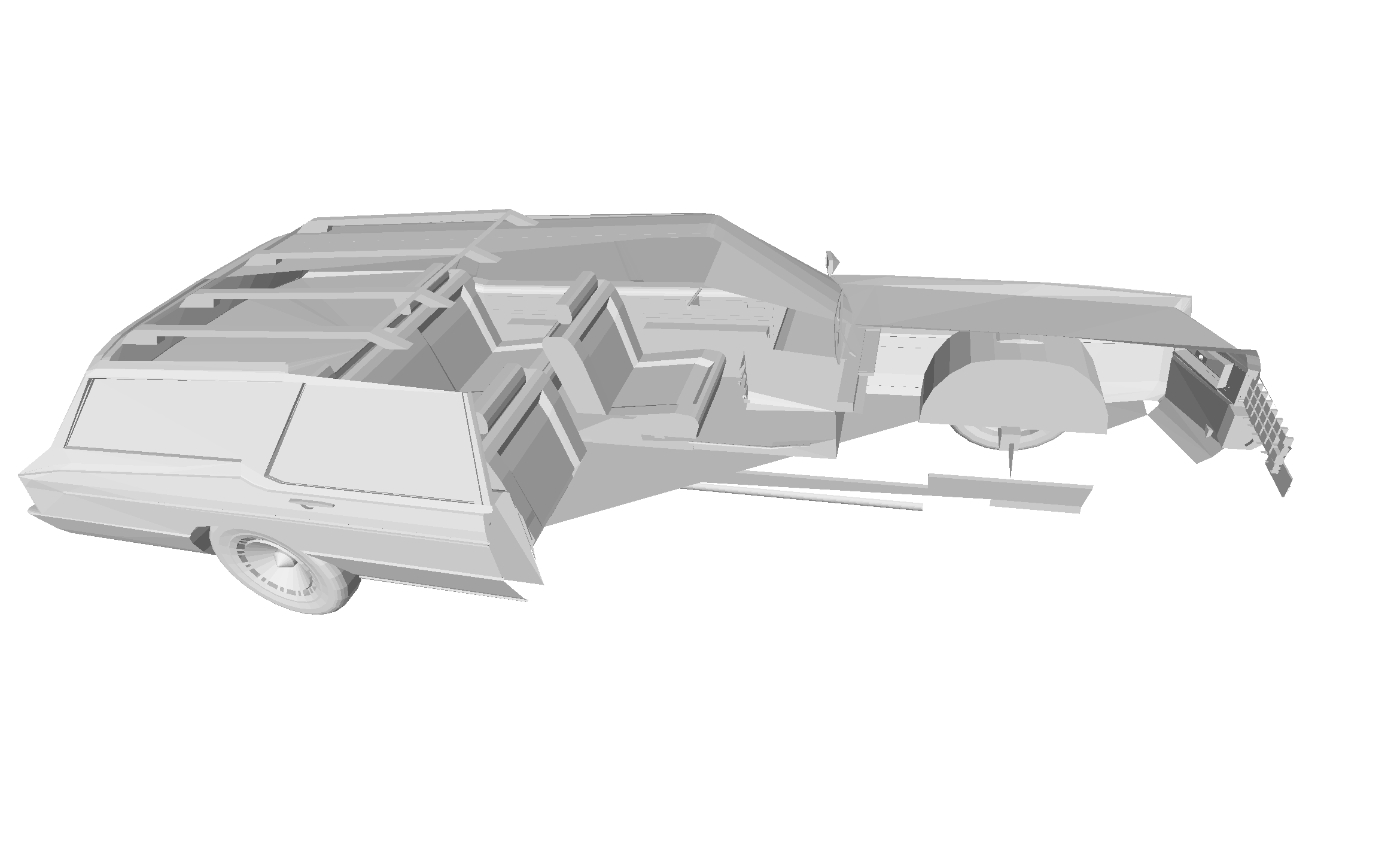}\\
\includegraphics[width=\subfigsize]{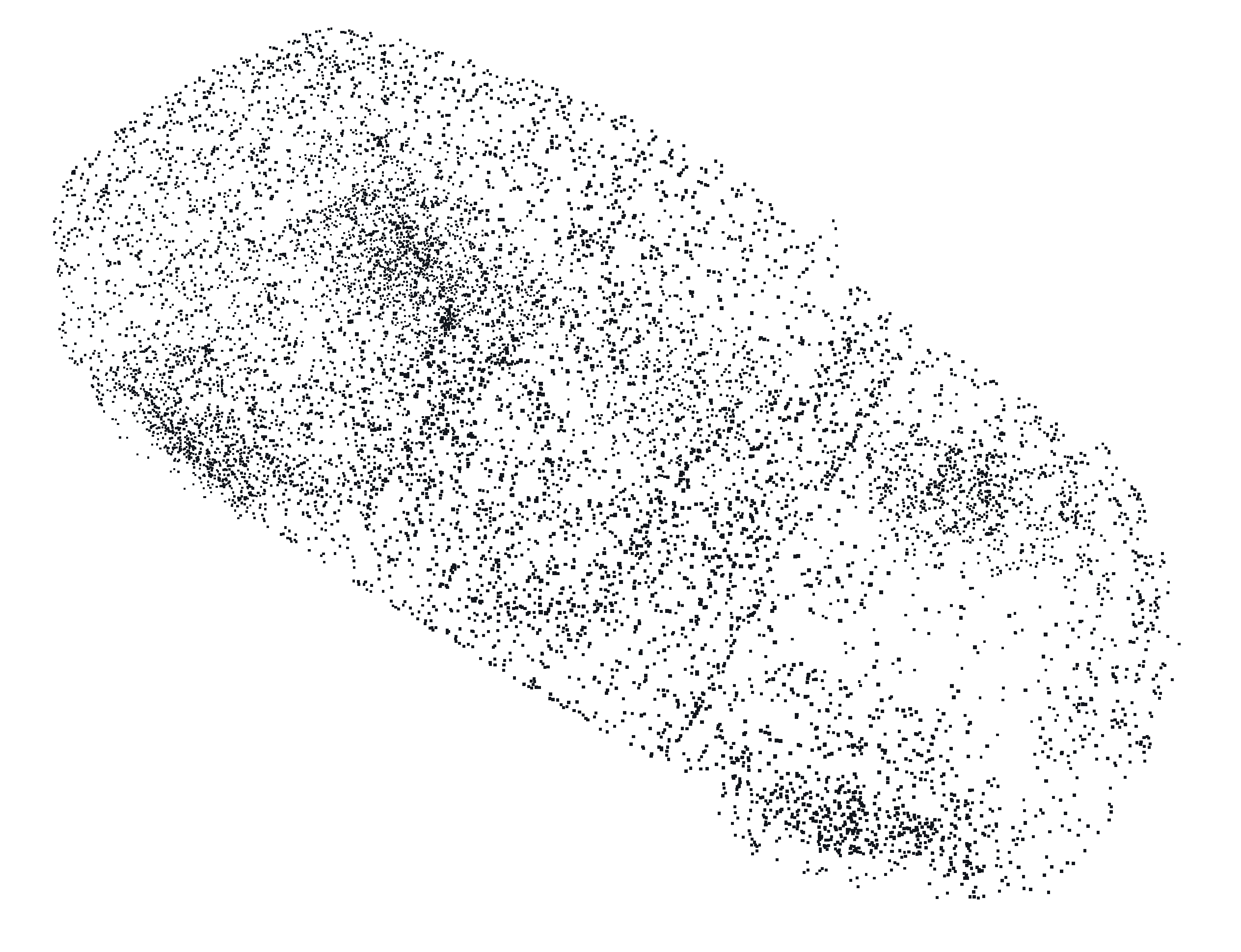}
\includegraphics[width=\subfigsize]{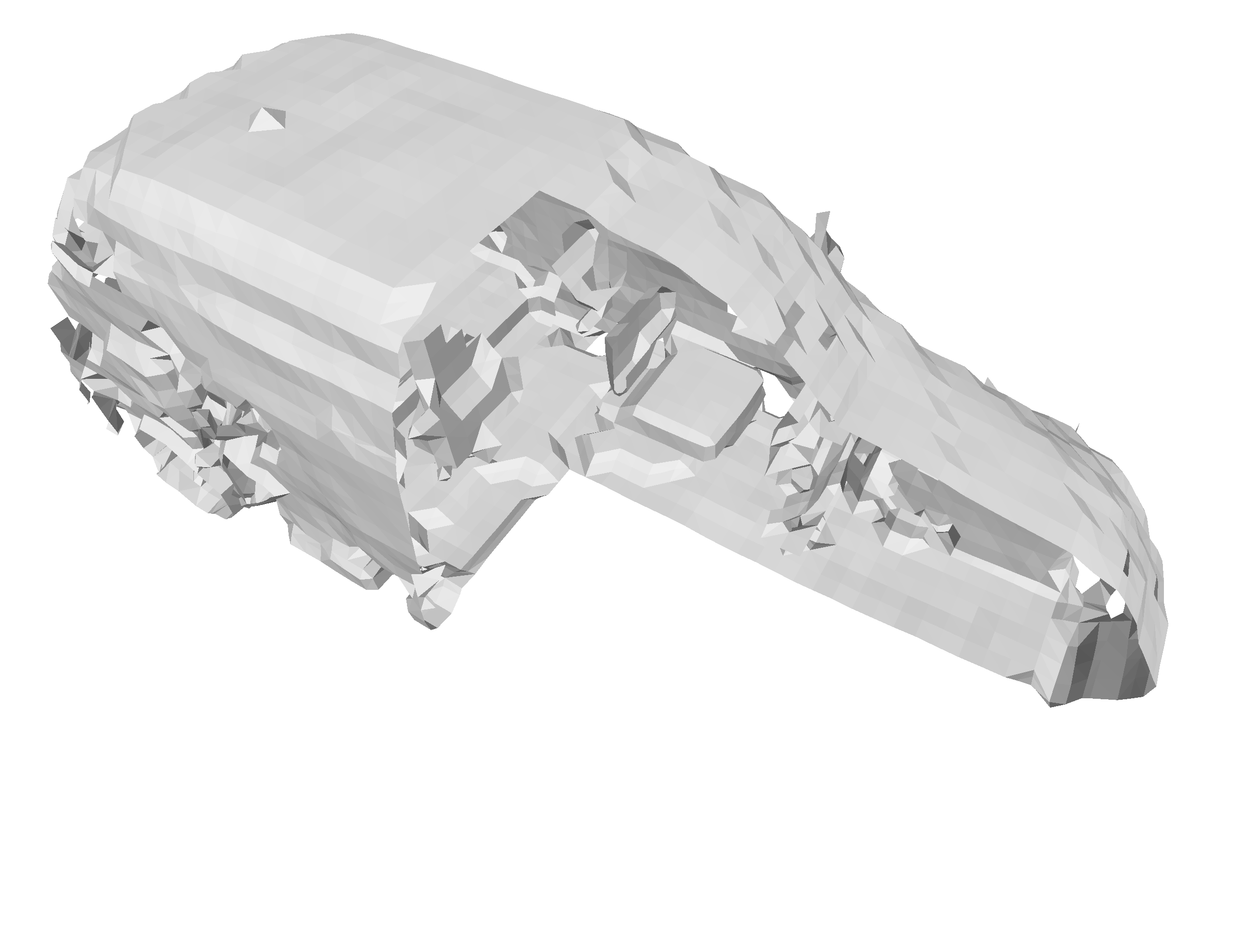}
\includegraphics[width=\subfigsize]{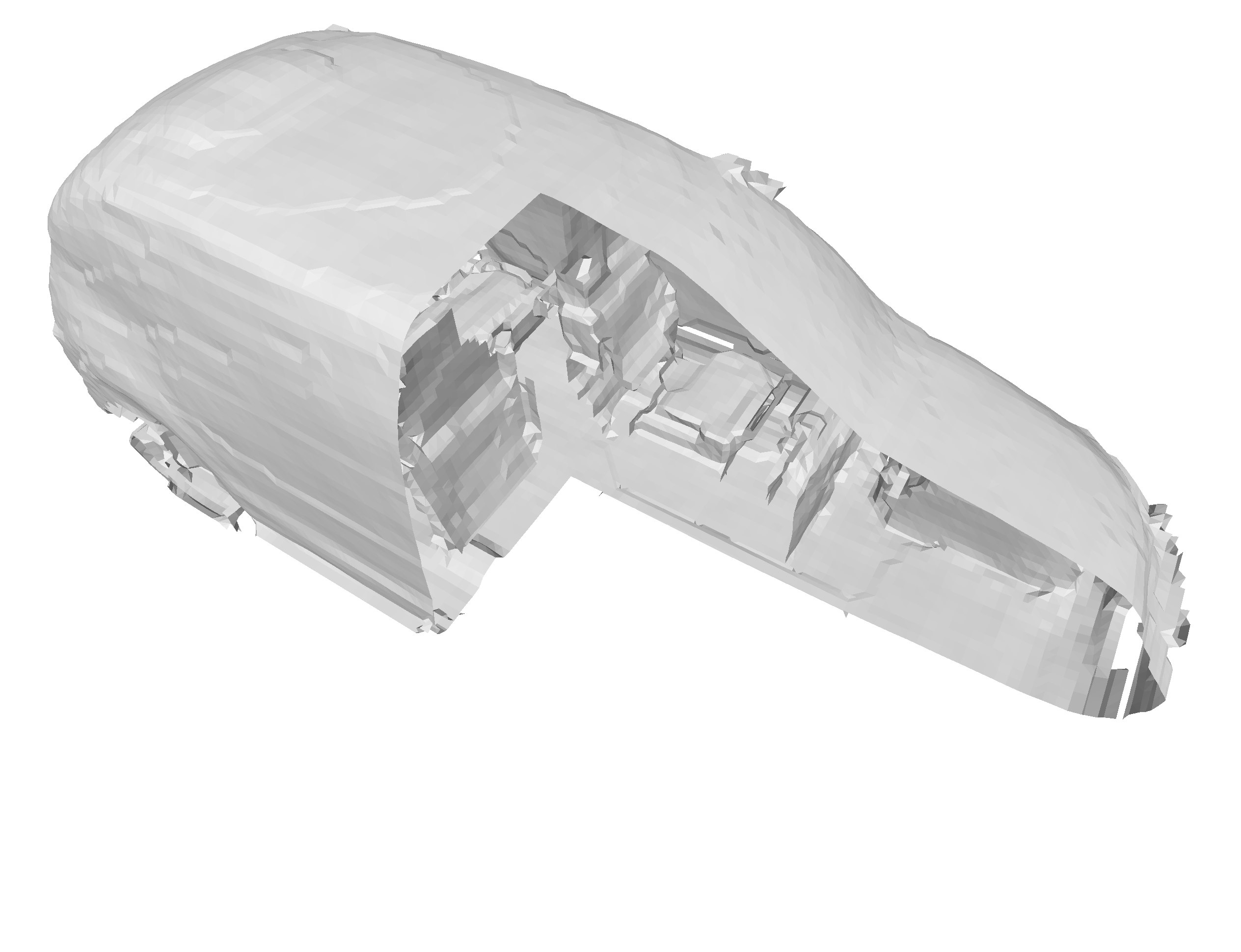}
\includegraphics[width=\subfigsize]{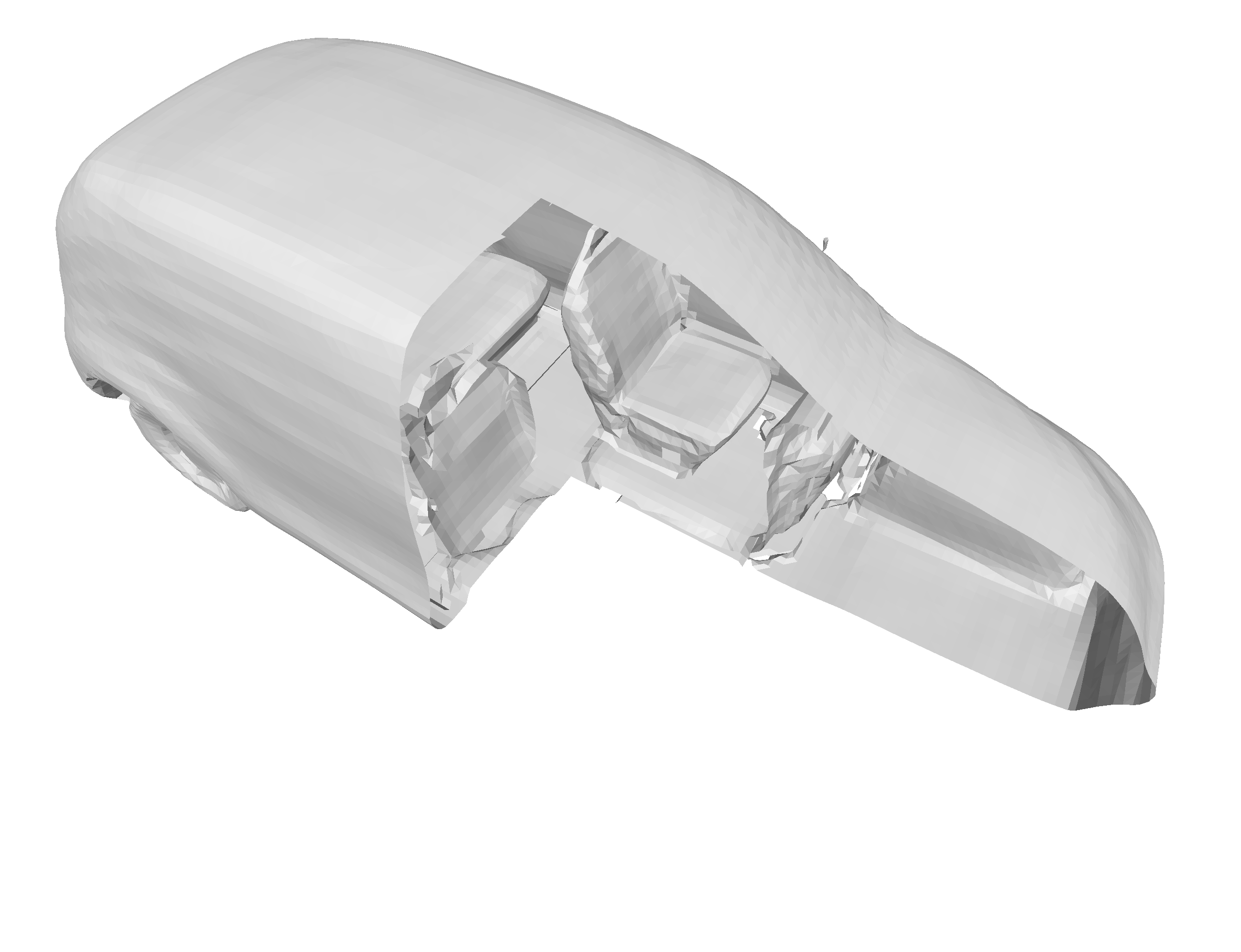}
\includegraphics[width=\subfigsize]{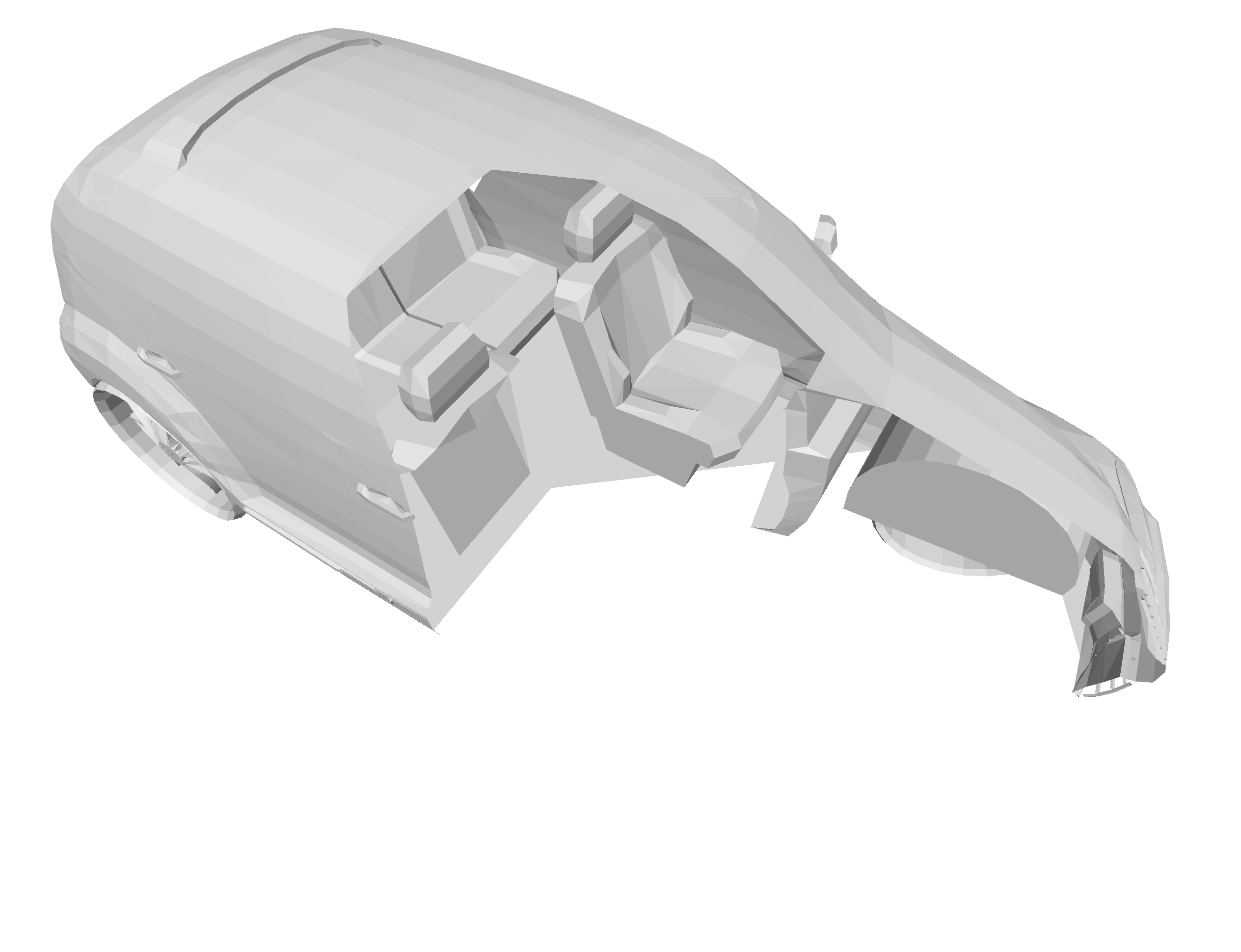}\\
\makebox[\subfigsize]{(a) Input }
\makebox[\subfigsize]{(a) NDF\cite{Chibane20b} }
\makebox[\subfigsize]{(b) CSP\cite{Venkatesh21} }
\makebox[\subfigsize]{(d) GDF (Ours) }
\makebox[\subfigsize]{(e) GT }
\end{center}
\caption{
\textbf{Representing previously unseen shapes using an auto-decoder.} Meshes reconstructed by fitting a latent variable model to 10000 surface points.  {\bf Top rows.}  Car from ShapeNet-Car(P).  {\bf Bottom rows}.   Cars from the ShapeNet-Car(R), which contains the inside of the car. We remove a quarter of the reconstructed models to make it visible. 
}
\label{fig:pointcloud} 
\end{figure*}



\section{Analysis and Ablations}
\subsection{Convergence Rate}

\def\subfigsize{0.48\linewidth}
\begin{figure}[t]
\begin{center}
\includegraphics[width=0.48\linewidth]{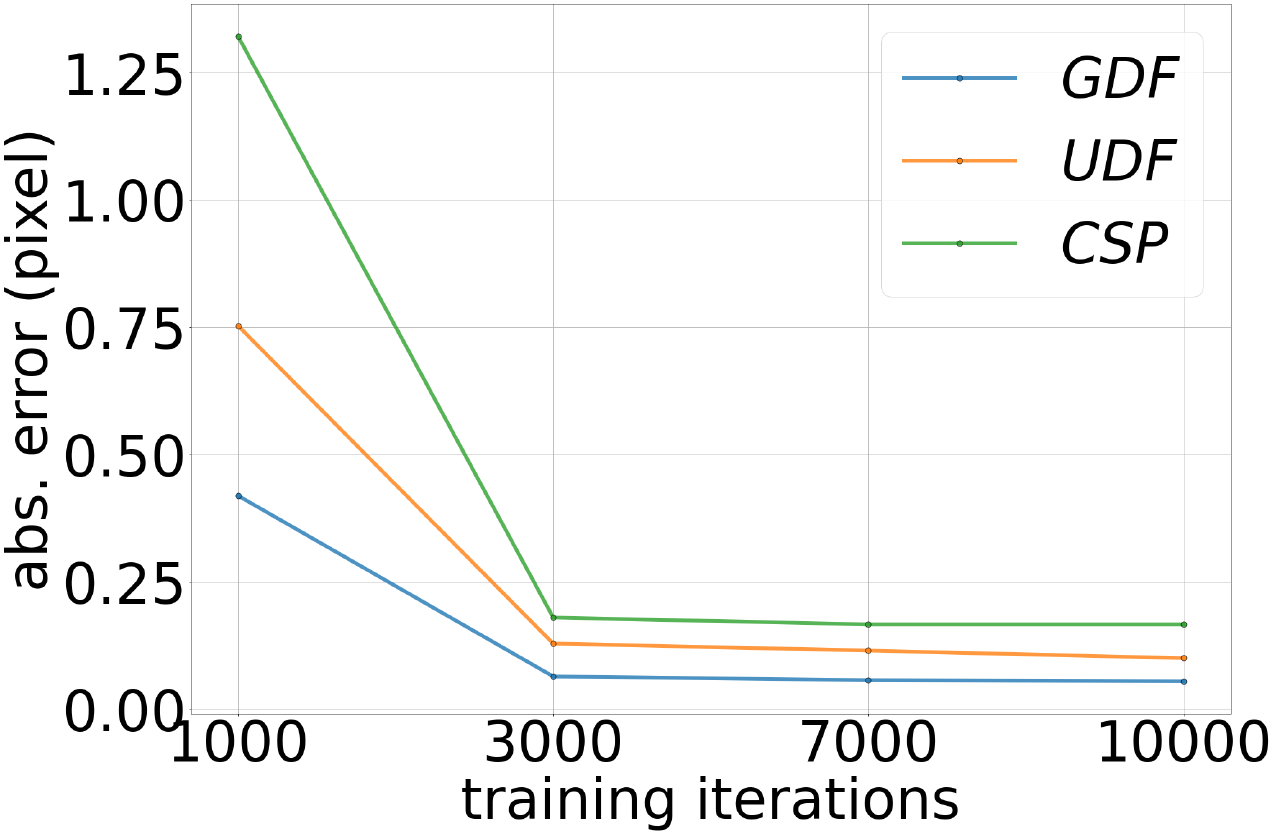}
\includegraphics[width=0.48\linewidth]{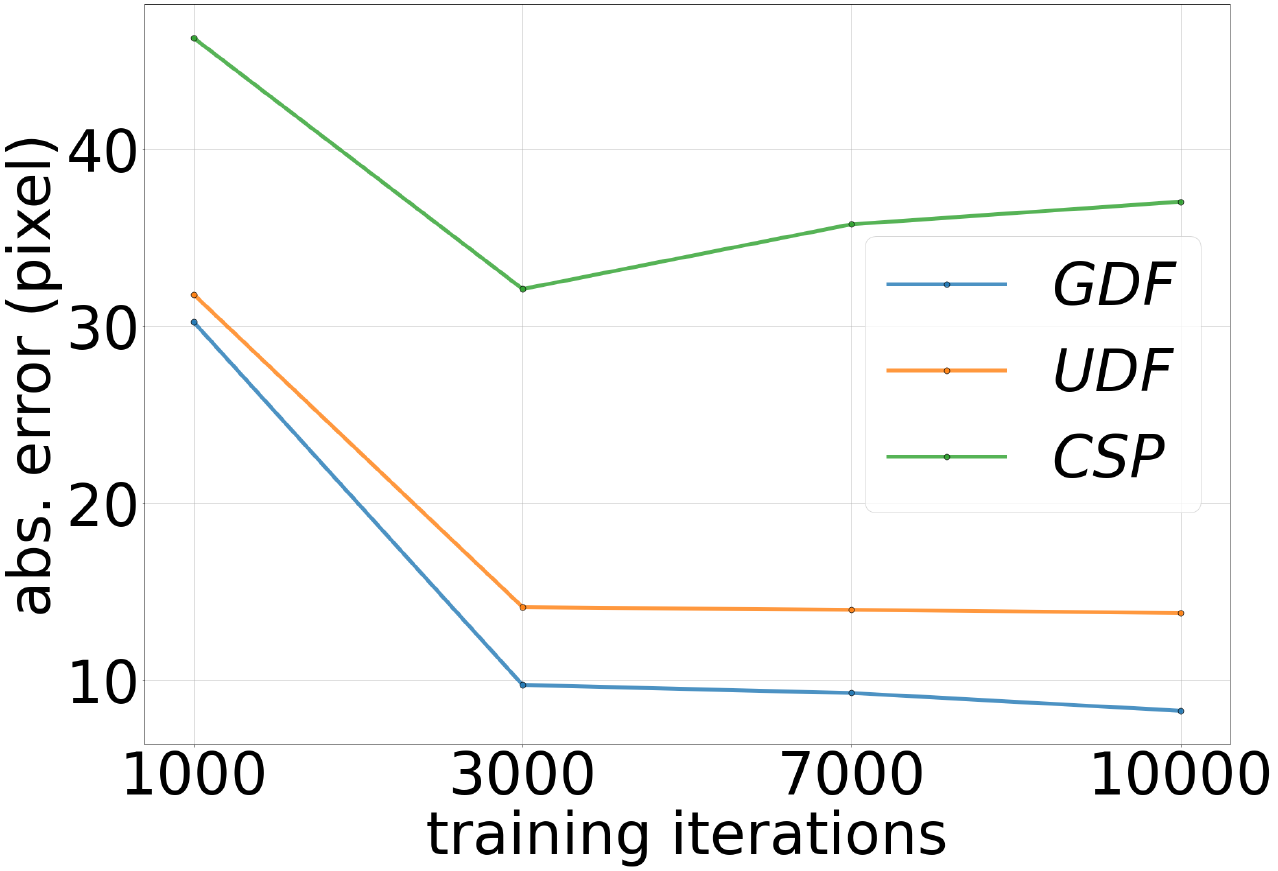}
\end{center} 
\vspace{-5mm}
\caption{
\textbf{Distance and gradient reconstruction errors during fitting.} We train identical networks to fit a single mesh using different implicit representations including GDF (Ours), CSP\cite{Venkatesh21} and NDF\cite{Chibane20b}. At different iterations, we measure the L1 distance between the estimated unsigned distances and gradients via the deep networks and the ones computed from the ground-truth mesh. Our GDF can minimize the errors much faster and more accurately.
} 
\label{fig:converge} 
\end{figure}

\def\subfigsize{0.22\linewidth}
\begin{figure}[t]
\begin{center}
\includegraphics[width=\linewidth]{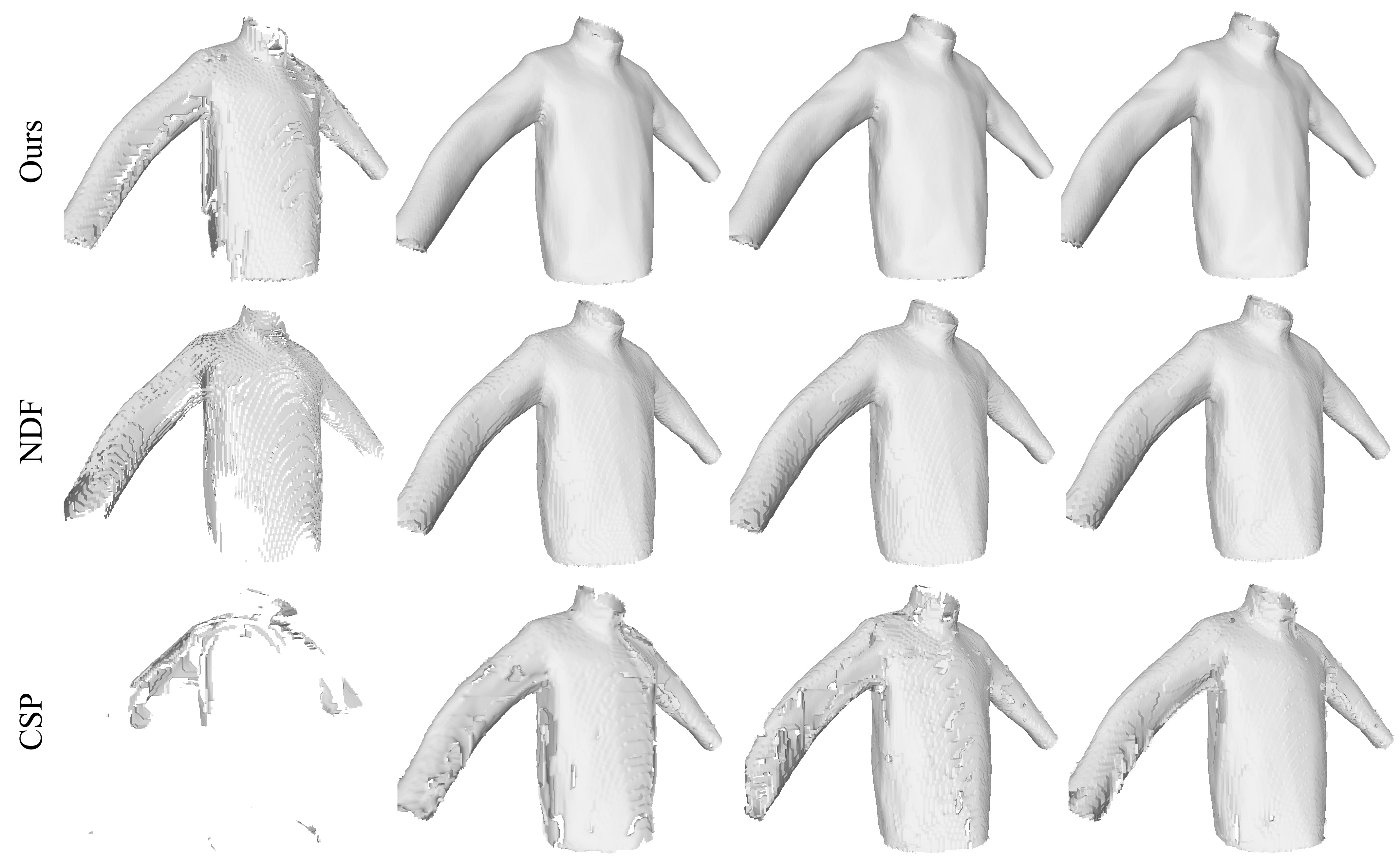}
\makebox[0.5cm]{ Iter.}
\makebox[\subfigsize]{1k }
\makebox[\subfigsize]{3k }
\makebox[\subfigsize]{7k }
\makebox[\subfigsize]{10k }
\end{center} 
\caption{
\textbf{The resulting surfaces at the fitting progresses.} We compare the output surfaces at different fitting iterations when using a single network to represent a single mesh via different representations including GDF (Ours), CSP\cite{Venkatesh21} and NDF\cite{Chibane20b}. After only 3000 iterations, our proposed GDF is able to output a smooth surface with almost no unwanted holes.  
} 
\label{fig:converge_vis} 
\end{figure}

Using the same metrics as above, we now look at the convergence rates while training the same network architecture but using different surface representations: UDF, GDF and CSP. As can be seen in Fig.~\ref{fig:converge}, GDF not only yields better final results but also converges much faster. The corresponding meshes are shown Fig.~\ref{fig:converge_vis}. After only 3000 iterations, the output from our GDF is already smooth and almost without holes, which is not true for either CSP or NDF. 



\subsection{Additional Loss Terms }
\label{sec:seperate-sup}

Unlike for SDFs, meshing a UDF often requires reasoning about its gradient~\cite{Guillard22b,Zhang23b}. Our proposed GDF can be easily decomposed into the unsigned distance and its gradient via Eqs.~\ref{eq:esdf} and~\ref{eq:esgrad}. This gives us direct access to the distance and its gradient for meshing.  We show that we can further supervise the training for each term separately. This additional supervision helps preserve the underlying structural integrity of the gradient field and results in more robust estimation of the UDF and its gradient.
In this scenario, we minimize the composite loss function:
\begin{align}
    \mathcal{L}(\Phi,\bC_1,\ldots,\bC_N)=  \sum_{i,j}  \bigl[  &   \lambda_{adf}  |\hat{\bv}_j^i  - \bv_j^i| \label{eq:lossAugm} \\
    + &\lambda_{grad}  \left | \frac{\hat{\bv}_j^i }{\| \hat{\bv}_j^i \|} - \bg_j^i \right|  \nonumber  \\
    + &\lambda_{udf} \left| \| \hat{\bv}_j^i  \| - \bu\right| \bigr]\; , \nonumber 
    \end{align}
where $\lambda_{adf},\lambda_{grad},\lambda_{udf}$ are scalar weights chosen to balance the loss terms. 


To demonstrate the effectiveness of this, we trained a network to reconstruct garments from the MGN dataset with and without the separate loss terms. The values of $\lambda_{adf},\lambda_{grad},\lambda_{udf}$ are empirically set to $(100,4,50)$. We then measured the distances between the reconstructed meshes and the ground-truth meshes for both unsigned distances and gradients. To this end, we sampled evenly points in the volume to obtain a 3D tensor of $512^3$ query points and then select from these only the points with close proximity (<1 pixel) to the ground-truth surface. We measured the L1 differences between the unsigned distances and gradients at those points between the ground-truth and the learned GDFs. We report the results in Table \ref{tab:ablation}.  The additional loss terms result in more accurate results. Nevertheless, even without them, GDF still outperforms the other methods. 

In general, GDF allows the application of direct supervision or, potentially,  regularization terms on both the unsigned distance and its gradient. Thus, previous methods \cite{Zhou22,Zhao21a} designed for UDFs can be directly applicable to GDFs. 

 

\begin{table}[t]
    \caption{\small \textbf{Influence of the Additional Loss Terms.} The first two columns indicate which components of the loss function of Eq.~\ref{eq:lossAugm} when training the network the reconstruct MGN garments. The L2 Chamfer Distances are multiplied by $ 10^{-5}$. The first rows is the same as the results in Table \ref{tab:single-mesh}.}
    \vspace{-3mm}
    \label{tab:ablation}
    \begin{small}
        \begin{center}
            \setlength{\tabcolsep}{3pt}
            \begin{tabular}{cc|cccc} 
                Dist. & Grad. & \textit{CD} ($\downarrow$) & \textit{NC}($\uparrow$) & \textit{Dist.E.} ($\downarrow$) & \textit{Grad. E.}($\downarrow$) \\ \midrule
                - &  -                   & 2.13 & 99.16 & $5\times10^{-4}$ & 0.0438 \\
				\checkmark  & -          & 2.09 & 99.53 & $3\times10^{-4}$ & 0.0367 \\
                - & \checkmark           & 2.09 & 99.67 & $5\times10^{-4}$ & 0.0277 \\
                \checkmark & \checkmark  & 2.07  & 99.67 & $3\times10^{-4}$ & 0.0262 \\ \midrule
            \end{tabular}
            \vspace{-18pt}
        \end{center}
    \end{small}
\end{table}

\subsection{Meshing at High Resolution}

\begin{figure}[ht!]
    \begin{center}
    \includegraphics[width=0.22\linewidth]{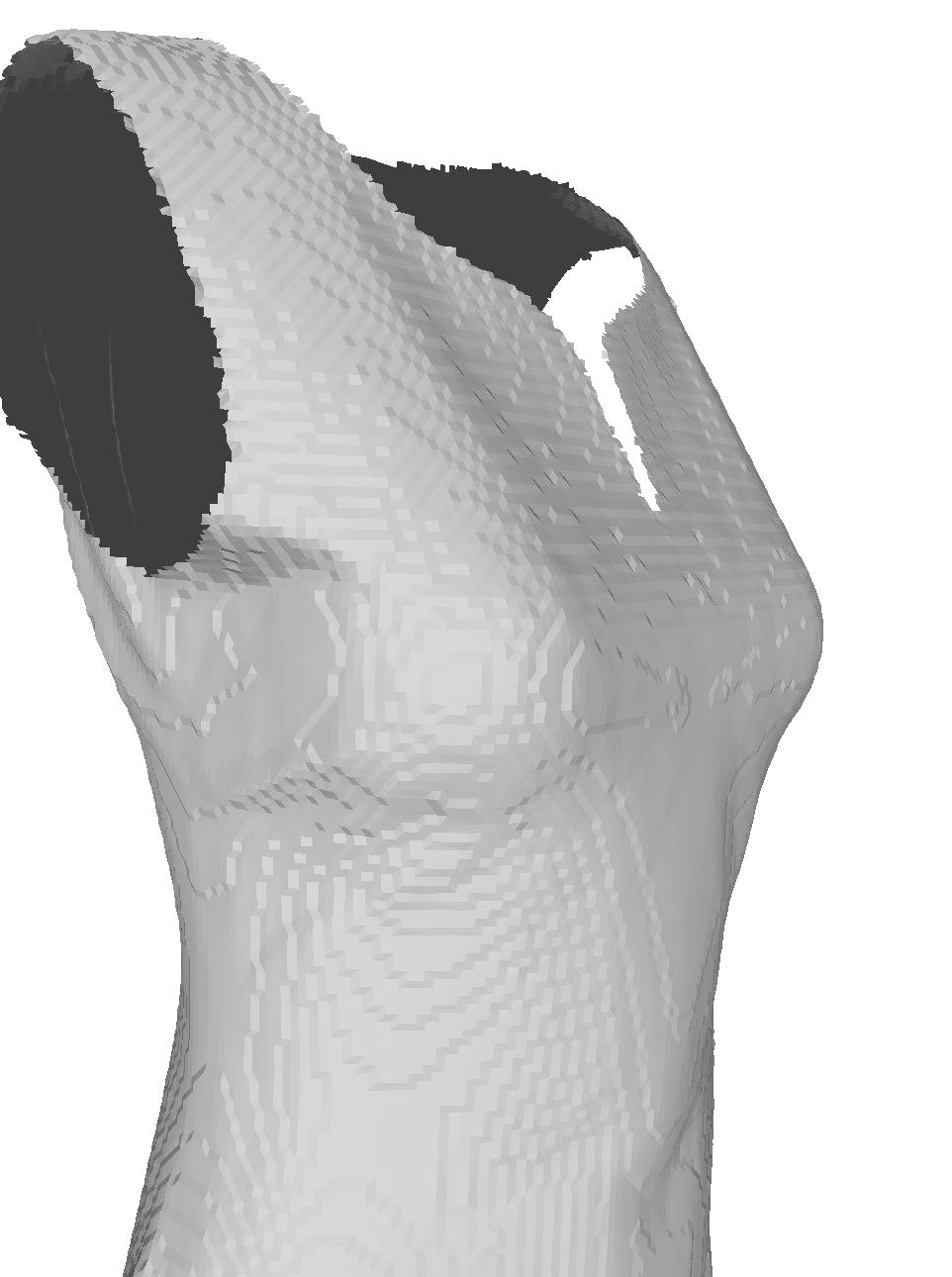}
    \includegraphics[width=0.22\linewidth]{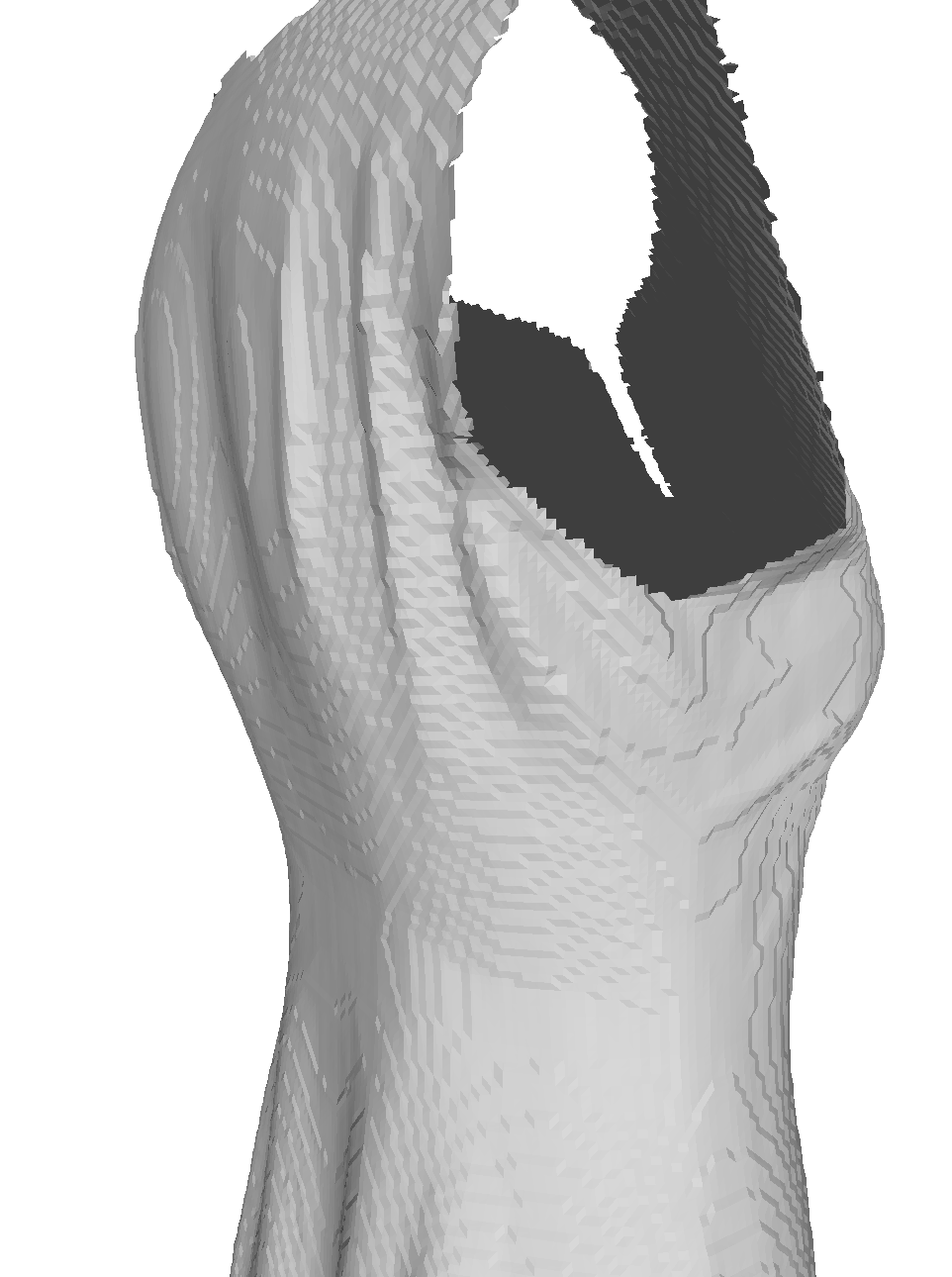}
    \includegraphics[width=0.22\linewidth]{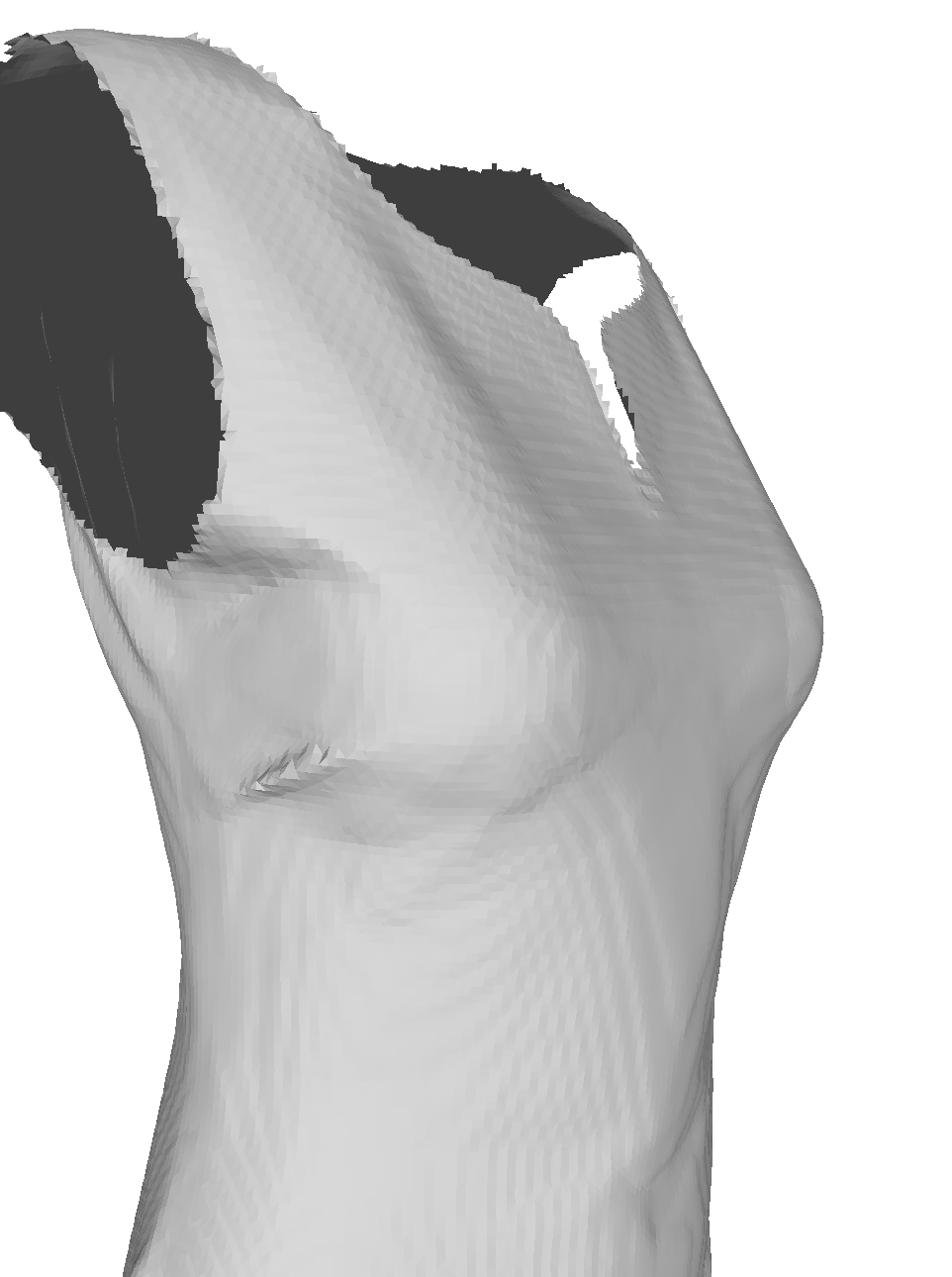}
    \includegraphics[width=0.22\linewidth]{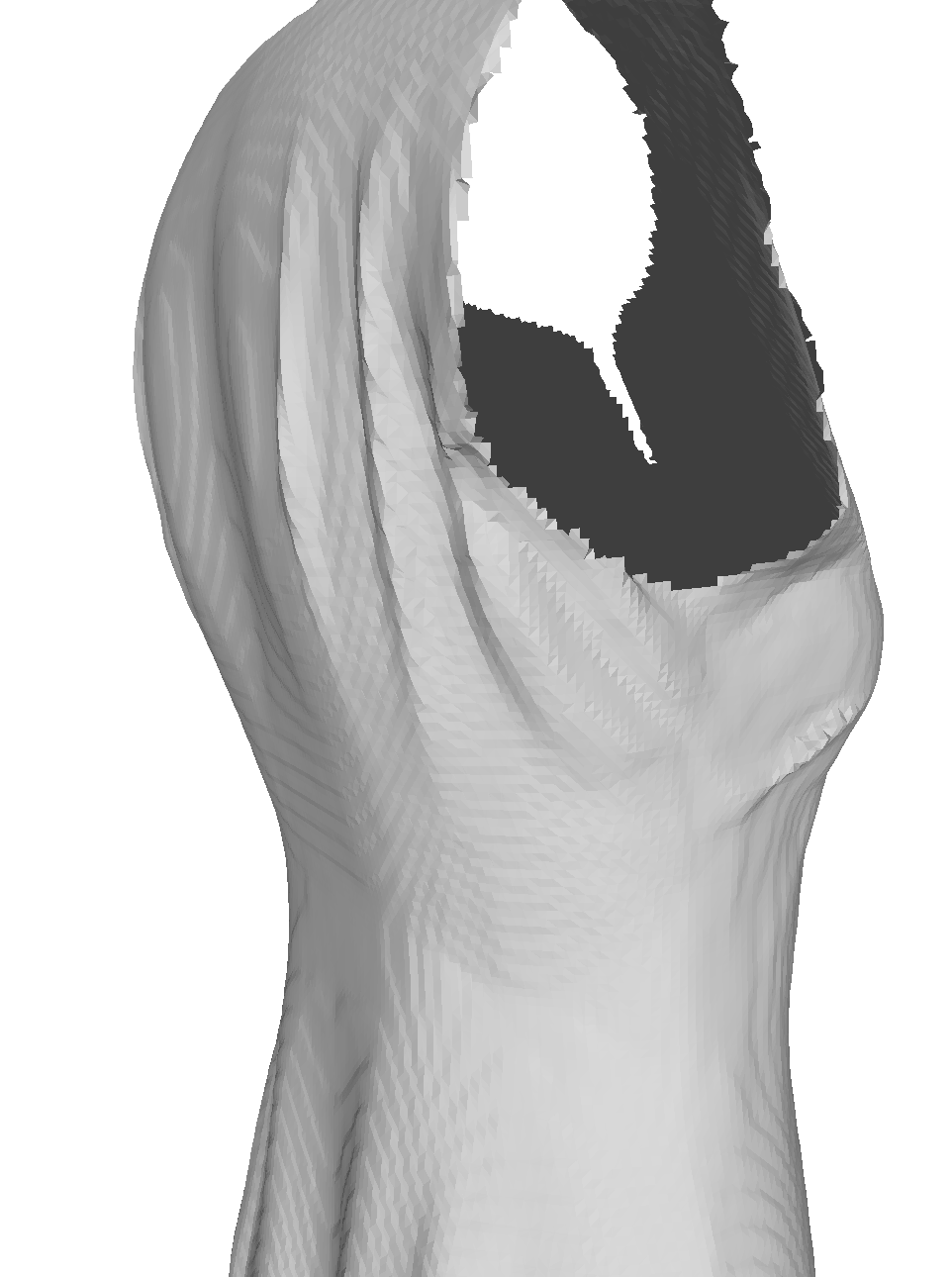}\\
    \includegraphics[width=0.22\linewidth]{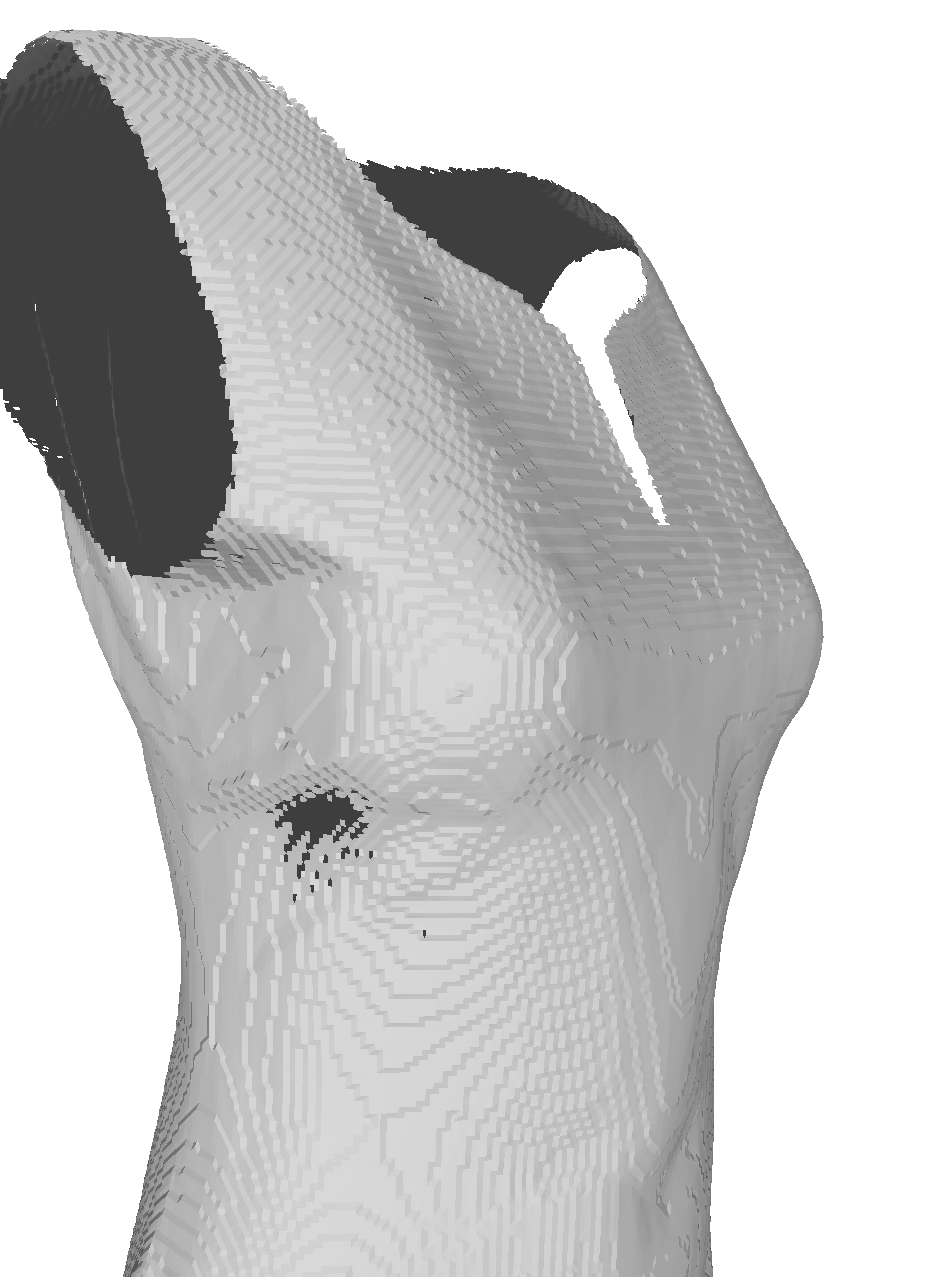}
    \includegraphics[width=0.22\linewidth]{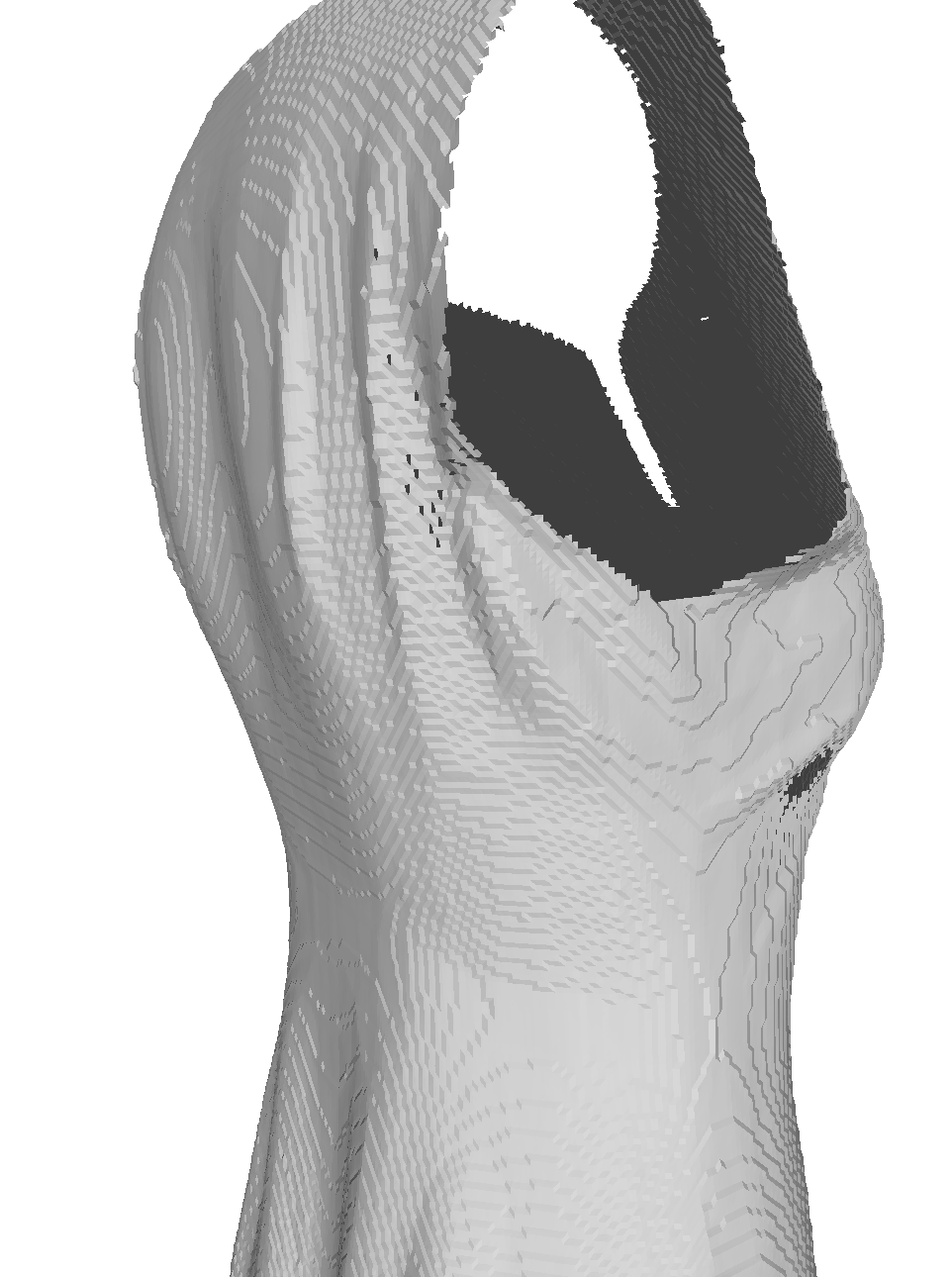}
    \includegraphics[width=0.22\linewidth]{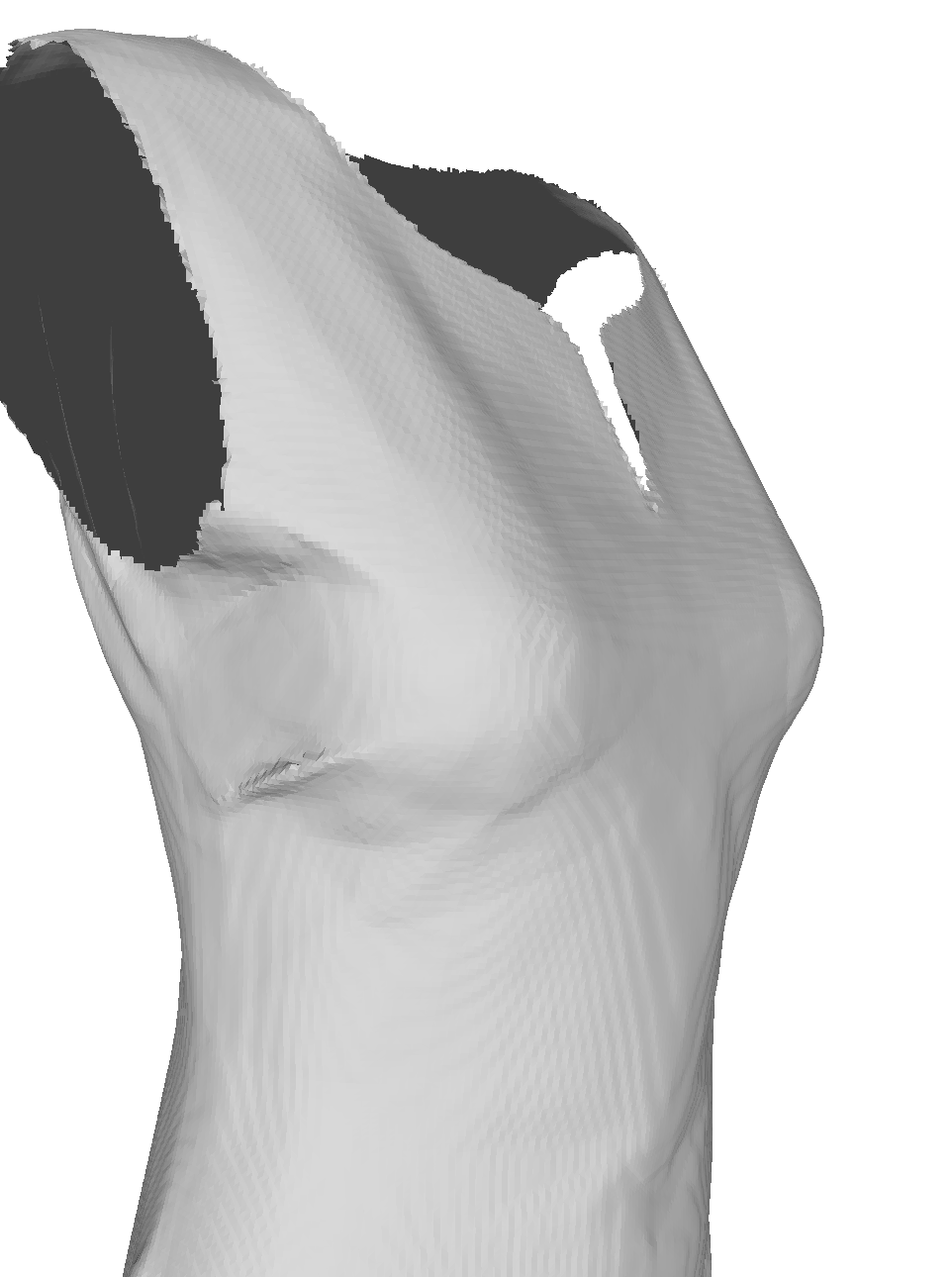}
    \includegraphics[width=0.22\linewidth]{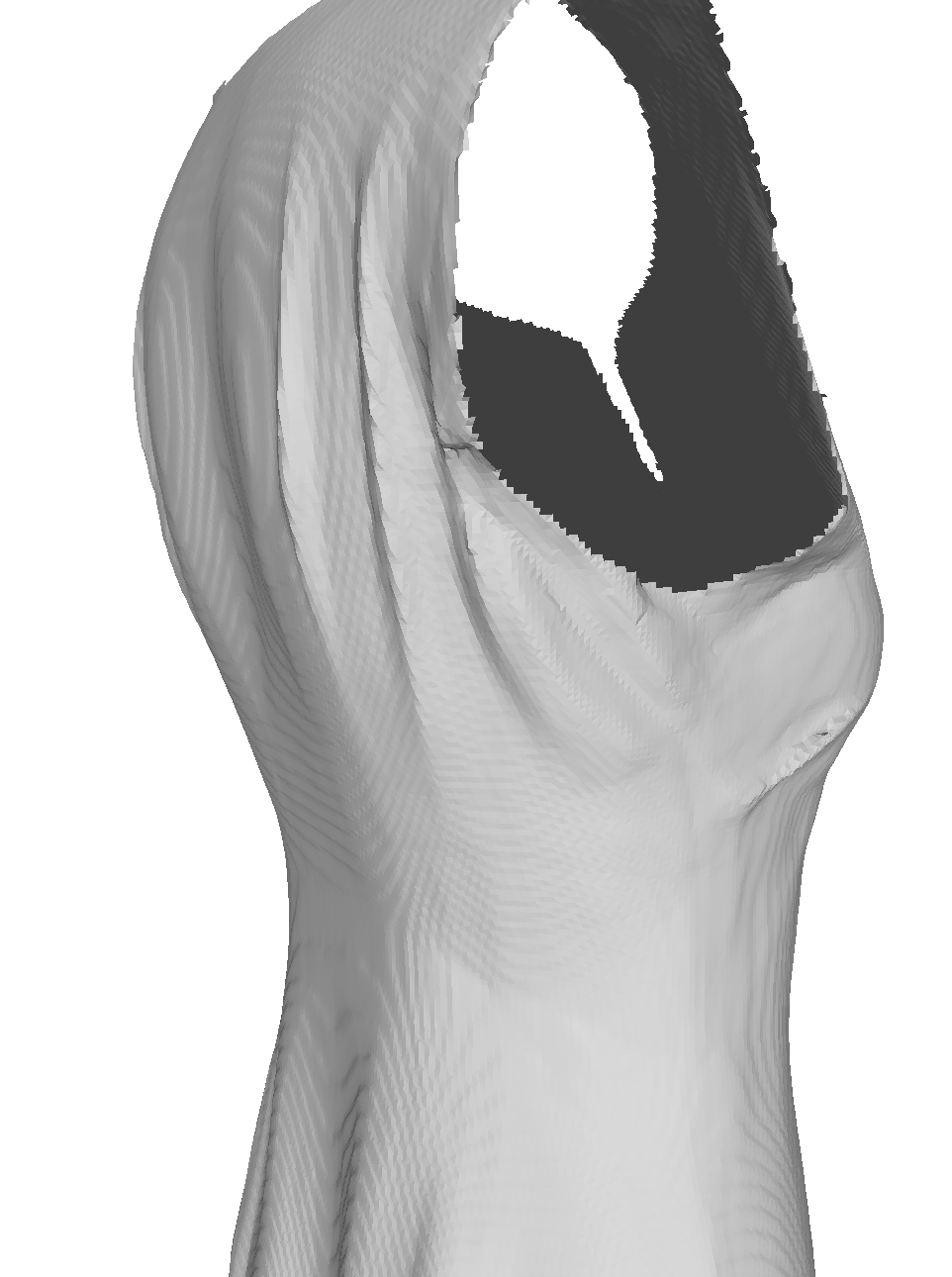}
    \makebox[0.45\linewidth]{(a) NDF\cite{Chibane20b}}
    \makebox[0.45\linewidth]{(b) Ours}
    \end{center} 
    \vspace{-5mm}
    \caption{
    \textbf{Reconstruction results of a shape at $512^3$ (top) and $768^3$ (bottom) resolution.} Higher resolution mesh reconstruction requires more precise distance and gradient estimation to prevent the appearance of holes. GDF representation results in smooth surface without holes when meshing at $768^3$ resolution. 
    }  
    \label{fig:highres} 
    \end{figure}

In Fig.~\ref{fig:highres}, we show the reconstruction results of a shape at $512^3$ and $768^3$ resolution. Higher resolution mesh reconstruction requires more precise distance and gradient estimation to prevent holes on surfaces. Training a network to output the GDF representation results in smooth surface without holes. It is not the case when regressing unsigned distance values, as in NDF\cite{Chibane20b}. Here we train the two networks to represent a single shape for 50000 iterations with the batch size of 32000 query points. 


\subsection{Gradient Distance Field Visualization}

\begin{figure}[ht!]
    \begin{center}
    \includegraphics[width=0.95\linewidth]{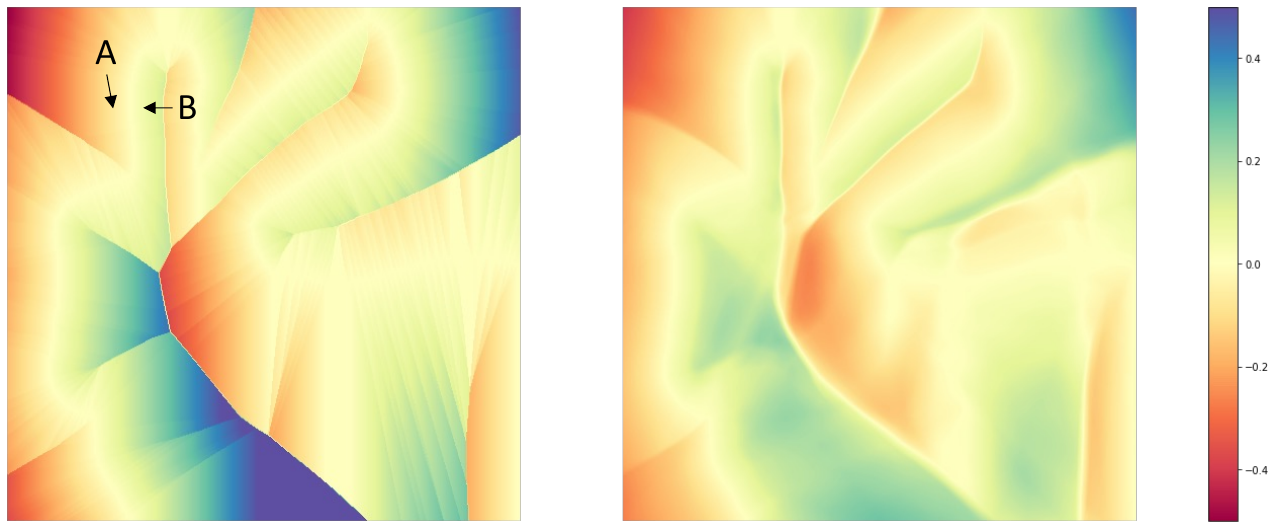}
\makebox[0.44\linewidth]{(a) GT distance}
\makebox[0.44\linewidth]{(b) Learned distance}
    \includegraphics[width=0.95\linewidth]{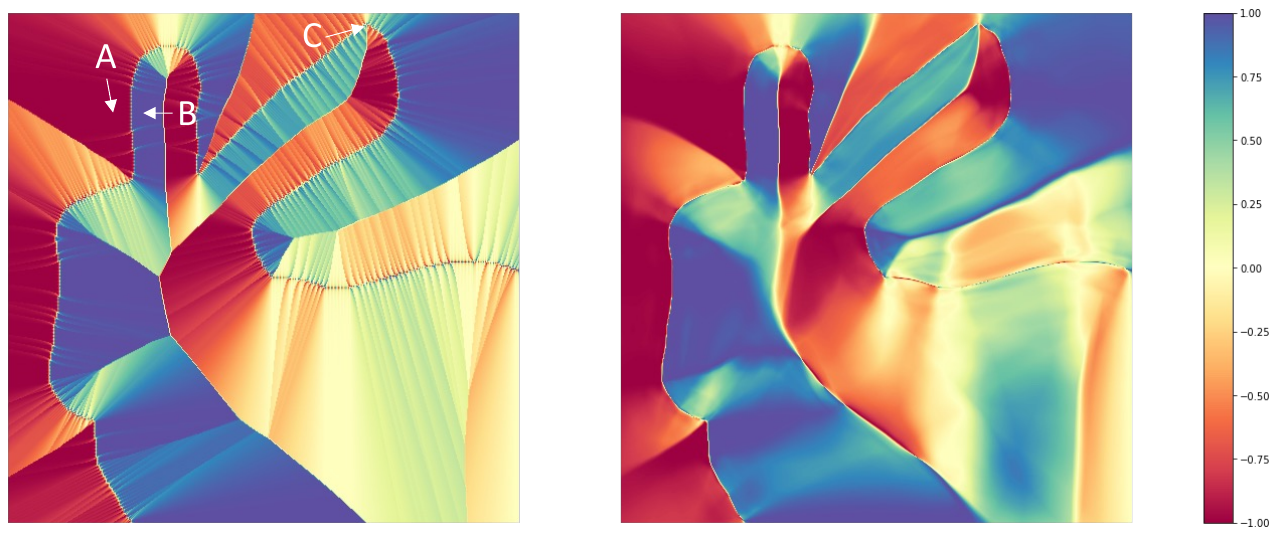}
\makebox[0.44\linewidth]{(c) GT x-grad}
\makebox[0.44\linewidth]{(d) Learned x-grad}

    \end{center} 
    \caption{
    \textbf{Ground-Truth vs Learned Gradient Distance Field.} We visualize GDF and the gradient component extracted from it for an open 2D contour for a portion of the Stanford Bunny (see Fig.\ref{fig:teaser}). The first two columns show the ground-truth and learned distance values. The last two columns show the corresponding gradient component for the x-dimension. 
    }  
    \label{fig:vis} 
    \end{figure}

We visualize the values of the gradient distance field extracted directly from the ground-truth data and those estimated via a deep network in Fig.~\ref{fig:vis} for a 2D version of GDF. In this experiment, we train a network to output a GDF representation of an open contour for a portion of the Stanford Bunny (as shown in Fig.\ref{fig:teaser}a).  The ground-truth GDF value at each query point is computed by finding the nearest point on the contour. We only show the values of the x-dimension. The top row visualizes the vectors while the bottom row visualizes the normalized gradient components extracted from them (via Eq.\ref{eq:esgrad}). As can be seen, the contour is placed at smooth, continuous, and differentiable areas of the GDF scalar field. For a pair of example points, $A$ and $B$, on the two sides of the contour, depicted as the end-point of the arrows in the first column, they have opposite gradient values: $\bg_{A_x} = -1$ and $\bg_{B_x} = 1$. The values of A and B in the first row have opposite signs: $\bv_{A_x}<0$ while $\bv_{B_x}>0$ and the contour in between is at exact zero-level set. In the second column, the deep network can effectively capture this continuity accurately. 


Further, it can be seen that GDF is inherently not differentiable everywhere. Discontinuities occur at locations where there are multiple closest surface points, \textit{i.e.}, medial axis of the shape. However, there points are often far away from the surface and do not affect the reconstruction accuracy. In rare instances where it happens near the surface, it typically affects a small number of points, such as the point $C$ illustrated in the bottom-left image. In practice, we did not observe substantial issues arising from the discontinuity caused by the medial axis. In fact, it is a common practice to only use near surface points for training the network. Thus, the majority of the medial axis would not be used for training. In this particular case shown in the figure, it can be seen that the network mainly models the near surface values. Note that even though the ground-truth gradients can be noisy at some surface points, the outputs of the deep network are not.  




\section{Conclusion}

We have shown that the Gradient Distance Function is an effective representation for 3D open surfaces. Each point is associated to a vector pointing to the closest point on the object surface. The norm of this vector is the unsigned distance while the orientation is the negative gradient of the unsigned distance. This representation is differentiable at the surface and can easily be learned by a deep network. We have demonstrated that, compared to UDF and other open surface representations, GDF yields more accurate distances and gradients, resulting in better object reconstruction. Extensive evaluations on three datasets demonstrates the expressive capacity and generalizability of GDF in capturing complex surface geometries. There are several directions for future work. Checking the zero-crossing points across different dimensions of the vector fields can avoid the misalignment between them. Further, a sampling scheme prioritizing potentially errornous areas such as near-surface medial axis would be beneficial when modelling complex surfaces.

\bibliographystyle
{splncs04}
\bibliography{string,vision,learning,cfd,graphics,optim,biomed,misc,main}
\end{document}